\documentclass{IEEE-Con-Sys-mag}

\usepackage{adjustbox}
\usepackage{multirow}
\usepackage{textcomp}
\usepackage{amssymb} 
\usepackage{amsthm}
\usepackage{physics}
\usepackage{bm}
\usepackage{pifont}
\usepackage{comment}
\usepackage{siunitx} 
\usepackage{xspace}

\usepackage{algorithm}
\usepackage{algpseudocode}

\usepackage{tcolorbox}
\usepackage{caption}

\usepackage{multicol} 
\usepackage{stfloats}

\usepackage{ifthen}

\newboolean{arxiv}
\setboolean{arxiv}{true} 

\jvol{XX}
\jnum{XX}
\paper{8}
\ifthenelse{\boolean{arxiv}}
{
  \jname{Journal}
  \jmonth{February}
  \pubyear{2025}
  \identifier{XXXXXX}
  \cover{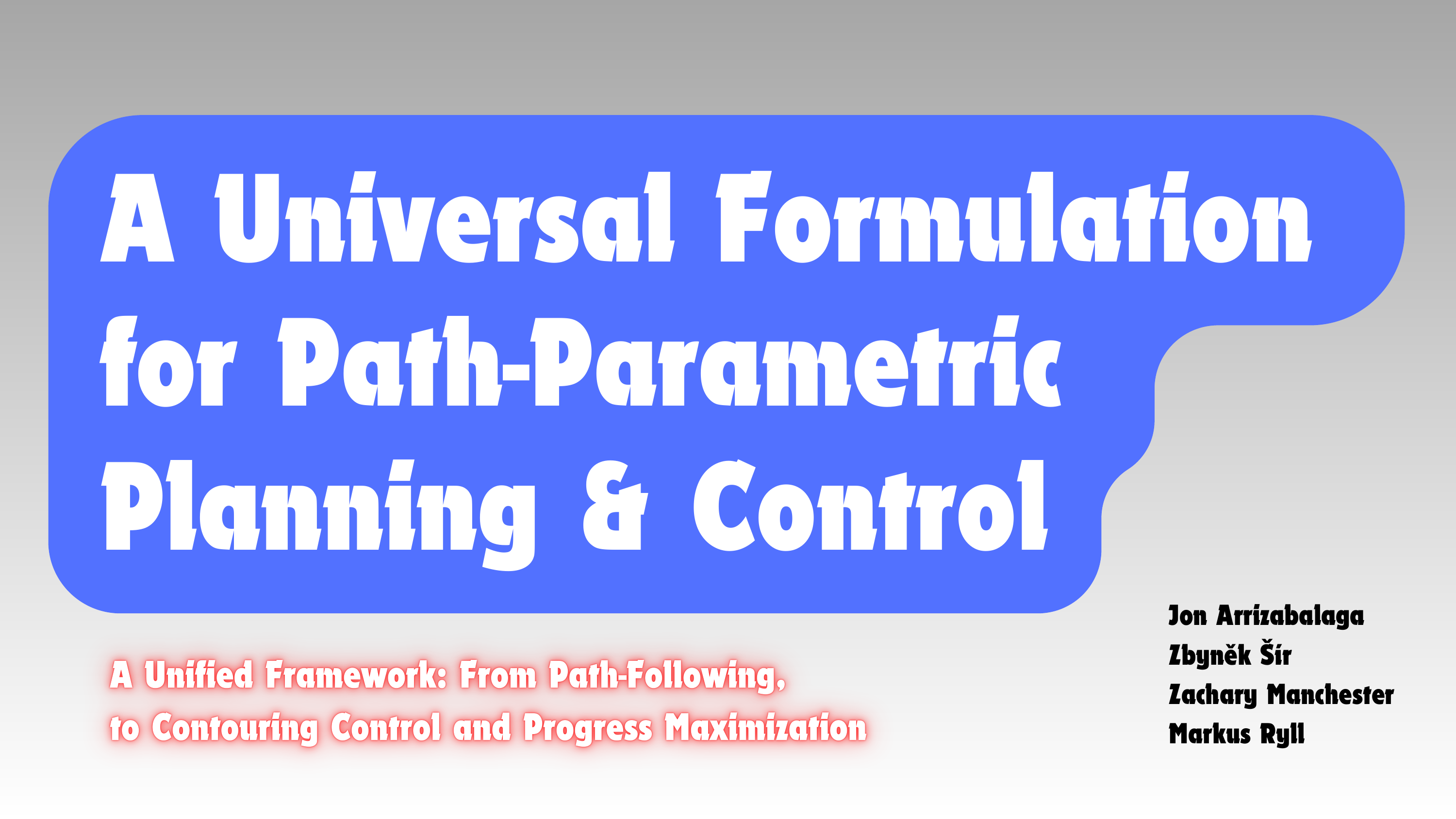}
}
{
\jname{IEEE CONTROL SYSTEMS}
\jmonth{March}
\pubyear{2025}
\identifier{10.1109/MCS.2025.000000}
\cover{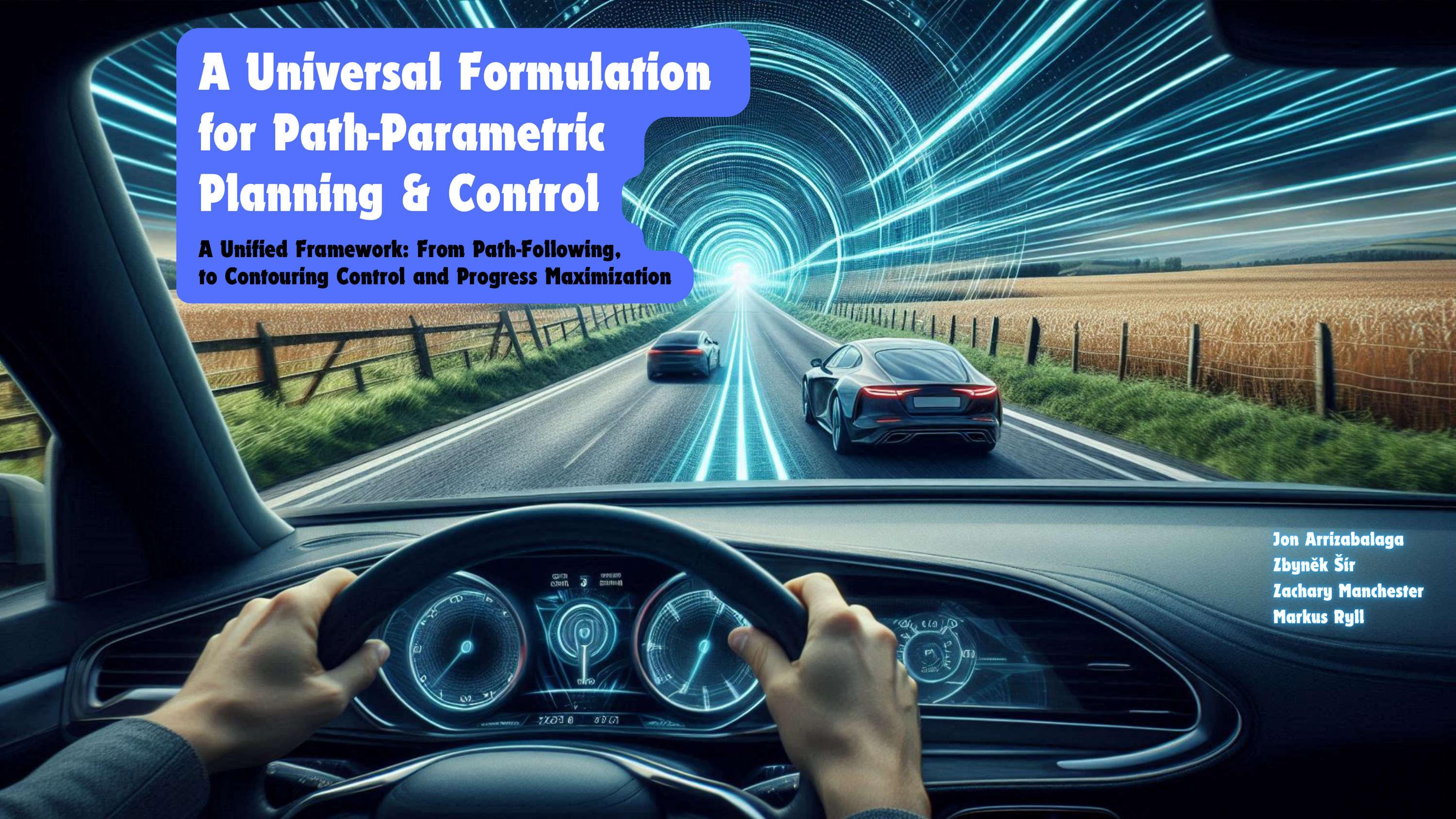}
}

\definecolor{ArriGreen}{rgb}{0, 0.5, 0}
\definecolor{progressColor}{RGB}{220, 255, 220}
\definecolor{progressColorStrong}{RGB}{0, 204, 0}
\definecolor{timeColor}{RGB}{255, 220, 255}
\definecolor{timeColorStrong}{RGB}{204,51,204}

\newcommand{\cmark}{\color{ArriGreen}{\ding{51}}}%
\newcommand{\xmark}{\color{red}\ding{55}}%

\newcommand{\txi}{\xi(t)}
\newcommand\ringring[1]{%
  {
   \mathop{\kern0pt #1}\limits^{
     \vbox to-1.85ex{
       \kern-2ex 
       \hbox to 0pt{\hss\normalfont\kern.1em \r{}\kern-.45em \r{}\hss}%
       \vss 
     }
   }
  }
}

\DeclareRobustCommand{\bi}{\textbf{i}}

\newcommand{\tm}{\textcolor{timeColorStrong}{\textit{\textbf{time-minimization}}}\xspace}
\newcommand{\prm}{\textcolor{progressColorStrong}{\textit{\textbf{progress-maximization}}}\xspace}

\newcommand{\tmp}{\textcolor{timeColorStrong}{\textit{\textbf{temporal}}}\xspace}
\newcommand{\spa}{\textcolor{progressColorStrong}{\textit{\textbf{spatial}}}\xspace}

\newcommand{\bluefigure}{
\def\figurename{\textcolor{strongerBlue}{\textbf{FIGURE}}}
\renewcommand{\thefigure}{\textcolor{strongerBlue}{\textbf{\arabic{figure}}}}
}
\newtheoremstyle{colon}%
{}
{}
{\itshape}
{}
{\bfseries}
{:}
{ }
{}
\theoremstyle{colon}
\theoremstyle{remark}

\newtcbox{\mybox}[2][]{
  on line,
  colback=#2, 
  colframe=black, 
  boxrule=0.5pt, 
  arc=3pt, 
  boxsep=1pt, 
  left=2pt, right=2pt, 
  top=1pt, bottom=1pt, 
  #1 
}

\newcommand\copyrightnotice{%
\begin{tikzpicture}[remember picture,overlay]
  \node[draw, inner sep=4pt] at (8.1cm, -13.75cm) {\parbox{\dimexpr\textwidth-\fboxsep-\fboxrule\relax}
  {\footnotesize This work has been submitted to the IEEE for possible publication. Copyright may be transferred without notice, after which this version may no longer be accessible.}
  };
\end{tikzpicture}%
}

\begin{document}

\title{\vspace{3.6cm}}

\author{\vspace{3.6cm}}
\affil{}

\maketitle


\ifthenelse{\boolean{arxiv}}{\copyrightnotice}{}

\dois{}{}

\chapterinitial{P}ath-parametric methods have become increasingly popular in navigation algorithm design, spanning high-level planners~\cite{verschueren2016time, spedicato2017minimum, arrizabalaga2023sctomp}, reinforcement learning (RL) policies~\cite{song2021autonomous,wurman2022outracing,kaufmann2023champion}, and low-level model predictive controllers (MPC)~\cite{liniger2015optimization,oelerich2024boundmpc,arrizabalaga2022towards}. The core idea behind these methods is to either introduce the path parameter as an additional degree of freedom—allowing the system to regulate its progress along the path~\cite{lam2010model, faulwasser2015nonlinear}—or to perform a coordinate transformation that projects the \emph{Euclidean states} onto \emph{spatial states}, i.e., the progress along the path and the orthogonal distance from it~\cite{verscheure2009time, van2016path, arrizabalaga2022spatial}. These parametric formulations have proven effective for three key reasons: (1) they naturally capture the concept of advancement along the path, (2) they embed the path’s geometric properties, such as curvature and torsion, into the system dynamics, and (3) spatial constraints become convex in the orthogonal components of the spatial states.


Given the broad range of problems encompassing path-parametric approaches, existing methods remain detached from each other and are frequently presented as independent work. This has resulted in a disjointed body of literature, where these techniques are viewed as distinct methods. Consequently, the reader is left with a fragmented view of the path-parametric problem, making it difficult to understand the interplay between the different techniques. To close this gap, in this paper we show how all these approaches are interconnected by presenting a universal formulation for path-parametric planning and control.

\begin{summary}
\summaryinitial{W}e present a unified framework for path-parametric planning and control. This formulation is universal as it standardizes the entire spectrum of path-parametric techniques -- from traditional path following to more recent contouring or progress-maximizing Model Predictive Control and Reinforcement Learning -- under a single framework. The ingredients underlying this universality are twofold: First, we present a compact and efficient technique capable of computing singularity-free, smooth and differentiable moving frames. Second, we derive a spatial path parameterization of the Cartesian coordinates for any arbitrary curve without prior assumptions on its parametric speed or moving frame, and that perfectly interplays with the aforementioned path parameterization method. The combination of these two ingredients leads to a planning and control framework that unites existing path-parametric techniques in literature. 

\vspace{3.0mm}

\noindent \textbf{Website}: \hspace{1mm}{\small \texttt{\url{https://path-parametric.github.io/}}}
\end{summary}


\newpage

To address this, the path-parametric problem is analyzed from three different yet interconnected perspectives (i-iii): We study the (i) \emph{interplay of existing parametric techniques} and show how they can be unified under a single framework consisting of \mybox{myLightYellow}{two ingredients}: (ii) a \emph{path-parameterization technique} and (iii) a \emph{spatial representation of the system dynamics}. Aiming to exemplify the utility of this framework, we implement state-of-the-art path-parametric methods in a two-link \mybox{myLightBeige}{robotic manipulator} and a miniature \mybox{framc}{race-car}. Finally, we extend the framework's applicability to unstructured navigable spaces by computing \mybox{framegreen}{safety corridors}.



\section{Path-parametric planning and control}~\label{sec:path_parametric_planning_and_control}

Before presenting a universal formulation for path-parametric planning and control, we first formally define the problems we aim to address. We then establish the conceptual connections between state-of-ther-art methods and, ultimately, demonstrate how they can be unified within a single framework.

\subsection{An overview of path-parametric methods}
Path-parametric methods are those that require from a reference path parameterized either by its arc-length or an arbitrary variable. This variable is referred to as the \emph{path-parameter} and it inherently captures the notion of progress along the path. The difference on how this path parameter is leveraged in the design of the planning and control algorithms is what distinguishes the different path-parametric methods. Conceptually, these differences can be grouped according to two standards: the system states and the navigation criterion. The system states refer to the way the path parameter is introduced in the system dynamics, while the navigation criterion relates to how the path parameter is used for planning and control.

Within the different system state representations, we distinguish three main categories: (i) \emph{augmented states}, where the path parameter is introduced as an additional (virtual) degree of freedom in the system dynamics (ii) \emph{projected states}, where the system dynamics are projected onto a coordinate system that is aligned with the reference path, and (iii) \emph{transformed states}, where the system dynamics are transformed from the temporal domain to the spatial domain, causing them to evolve according to the path parameter instead of time. 

The navigation criteria can also be grouped into three categories: (i) \emph{progress or velocity profile regulation}, where the path-parameter's speed is desired to follow a predefined speed profile, (ii) \emph{time minimization}, where the aforementioned transformation from the temporal to the spatial domain is leveraged to convert the time minimization into a finite horizon problem, and (iii) \emph{progress maximization}, where the path parameter is used to maximize the progress along the path.
    
Existing path-parametric methods in the literature result from combining one of the categories corresponding to the state representation with those within the navigation criteria. This is shown in Table~\ref{tab:param_methods}, which provides an overview of the most relevant path-parametric methods in literature, alongside the system states and navigation criteria they use. The literature attached in the table is sorted by year of publication, and the specific method employed to implement the path-parametric representation: Control Law (CL), Optimization-based Planner (OP), Model Predictive Control (MPC) and Reinforcement Learning (RL). This wide spectrum demonstrates how path-parametric methods have evolved over time, becoming more sophisticated as the planning and control research fields have advanced. 
\begin{figure*}[t]
	\centering
    \begin{mdframed}[
    backgroundcolor=myLightGray,
    innertopmargin=4pt,
    linecolor=strongerGray,
    linewidth=0pt,
    innerrightmargin=8pt,
    innerleftmargin=8pt,
    roundcorner=6pt,
]
     \includegraphics[width=\linewidth]{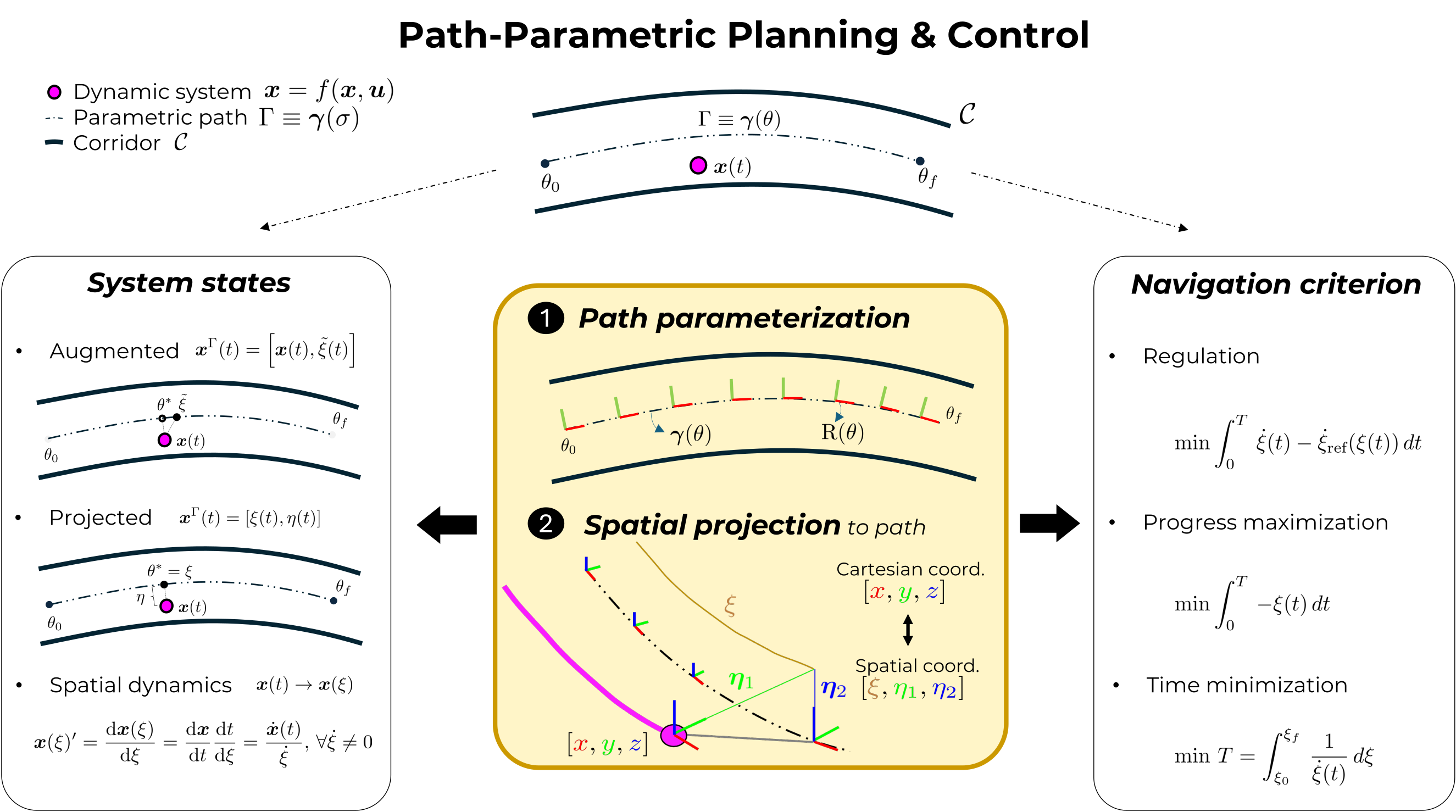}
    \end{mdframed}    
    \captionsetup{labelformat=simple, labelfont={color=strongerGray, bf}} 
    \def\figurename{\textcolor{strongerGray}{\textbf{FIGURE}}}
	\caption{Path-parametric methods rely on a reference path parameterized by an auxiliary variable, $\sigma$. These methods can be classified based on two key aspects: the system states and the navigation criterion (as indicated by the white boxes). However, the literature on parametric methods remains fragmented, with existing approaches often presented in isolation. To unify these methods, we introduce a universal formulation that highlights their underlying connections. This formulation consists of two main components (shown in the yellow box): (i) a path-parameterization technique for computing moving frames, and (ii) a spatial projection of the Cartesian system dynamics onto the parametric path, formulated without imposing prior assumptions on the moving frame.}\label{fig:uppc_overview}
    \vspace{2mm}
\end{figure*}
\subsection{A brief history: From Path Following to MPC and RL}\label{subsec:history}


The description of the complex shapes and motions encountered in the world is a fundamental pursuit in the sciences and engineering. While traditional Cartesian equations are adept at representing simple lines and circles, they often prove inadequate for capturing the intricacies of curves and surfaces observed in nature and human-made objects. This limitation necessitates the development of more versatile tools, and parametric equations emerge as a powerful solution. For example, parametric equations can be employed to express complex motions and paths, such as the trajectory of a projectile or the outline of a shape. For this reason, the concept of parameterizing a path by its arc length (or a proxy variable) has captivated the interest of humanity for centuries. Equally significant is the related idea of defining a curve by its curvature, a measure of "how much it bends." The fascination with these methods of defining curves can be traced back to ancient Greece, where philosophers such as Aristotle and mathematicians like Archimedes explored these concepts. The advent of differential calculus in the 17th century introduced new tools and rekindled interest in this problem, attracting the attention of some of history's greatest mathematicians, including Newton, Descartes, Leibniz, Euler, and Gauss. For a comprehensive account of the fascinating history of curve parameterization, please refer to \cite{thehistoryofcurvature}.



In the realm of planning and control, the inception of path-parametric methods emerged in the 1980s, notably within the domain of robotic manipulators. Early endeavors identified the advantages of incorporating the path parameter in planning and control laws, such as rescaling infeasible trajectories~\cite{hollerbach1983dynamic} or computing end-effector motions for traversing a path in minimum time~\cite{shin1985minimum, pfeiffer1987concept}. These works were the first leveraging the path parameter to transform the temporal dynamics into spatial dynamics, as explained in the previous subsection and depicted in Fig.~\ref{fig:uppc_overview}. Soon thereafter, applications in the context of navigation for mobile robots appeared in~\cite{nelson1988local,kanayama1990stable,cox1991blanche}, where the error to the path was decoupled into a tangent and an orthogonal component. This decoupling allowed for the design of control laws that regulated the progress along the path, while ensuring the robot would always converge to it. These findings sowed the seed for upcoming more sophisticated formulations, such as the works under the spatial projections category (column 7 in Table~\ref{tab:param_methods} and Fig.~\ref{fig:uppc_overview}). The subsequent years served for establishing the theoretical foundations and getting a better understanding of these formulations~\cite{hauser1995maneuver,skjetne2004robust}, resulting in the articulation of \emph{path-following} as a distinct and superior alternative to path-tracking~\cite{do2004robust,aguiar2005path,aguiar2007trajectory, faulwasser2015nonlinear}.

\begin{sidebarGray}{An overview of path-parametric formulations and their applications}
\noindent\begin{minipage}{\textwidth}
\begin{table}[H]
    \captionsetup{labelformat=simple, labelfont={color=strongerBlue, bf}} 
    \def\figurename{\textcolor{strongerGray}{\textbf{TABLE}}} 
    \captionsetup{labelformat=simple, labelfont={color=strongerGray, bf}} 
    \caption{Path-parametric approaches sorted by the year, system applicability, methodology, capacity to account for deviations from the path, system state representation and navigation criterion. The methods are "Control Law" (CL), "Optimization-based Planner" (OP), "Model Predictive Control" (MPC), "Reinforcement Learning" (RL), and for those that do not fit into any of the previous, "Other" (OT).}\label{tab:param_methods}
    \begin{adjustbox}{width=1\textwidth}
	\centering
	\begin{tabular}{|c c || c c c||c c c|c c c|}
		\hline
		\multirow{2}{*}{\textbf{Ref.}} & \multirow{2}{*}{\textbf{Year}} & \multirow{2}{*}{\textbf{System}} & \multirow{2}{*}{\textbf{Method}} & \multirow{2}{*}{\textbf{Dev.}} & \multicolumn{3}{c|}{\textbf{System states}} &  
        \multicolumn{3}{c|}{\textbf{Navigation criterion}}  \\
        & & & & &Augm. & Proj. & Transf. & Reg. & Time & Progr. \\
        \hline
        \cite{hollerbach1983dynamic} & 1984 & Rob. Mani. & OT & \xmark & & & $\bullet$  & $\bullet$ & & \\
        \cite{shin1985minimum} & 1985 & Rob. Mani. & OP & \xmark & & & $\bullet$  & & $\bullet$ & \\
        \cite{pfeiffer1987concept} & 1987 & Rob. Mani. & OT & \xmark & & & $\bullet$  & & $\bullet$ & \\ 
        \cite{nelson1988local} & 1988 & UGV & CL & \xmark & $\bullet$ & & & $\bullet$ & & \\ 
        \cite{kanayama1990stable} & 1990 & UGV & CL & \xmark & $\bullet$ & & & $\bullet$ & & \\ 
        \cite{cox1991blanche} & 1991 & UGV & CL & \xmark & $\bullet$ & & & $\bullet$ & & \\ 
        \cite{hauser1995maneuver} & 1995 & Any$^{1}$ & CL &  \xmark& $\bullet$& & & $\bullet$ & & \\
        \cite{skjetne2004robust} & 2004 & Any & CL &  \xmark& $\bullet$& & & $\bullet$ & & \\
        \cite{do2004robust} & 2004 & Ship & CL & \xmark & $\bullet$& & & $\bullet$ & & \\
        \cite{aguiar2005path} & 2005 & Any & CL & \xmark & $\bullet$& & & $\bullet$ & & \\
        \cite{aguiar2007trajectory} & 2007 & Underactuated & CL & \xmark & $\bullet$& & & $\bullet$ & & \\
        \cite{verscheure2009time} & 2009 & Rob. Mani. & OP & \xmark & & &$\bullet$ &   & $\bullet$ & \\
        \cite{faulwasser2009model} & 2009 & Any & MPC & \cmark & $\bullet$ & & & $\bullet$ & &\\
        \cite{lam2010model} & 2010 & Any & MPC & \xmark & &$\bullet$ & &  & & $\bullet$\\
        \cite{kehrle2011optimal} & 2011 & Car & MPC & \cmark & & & $\bullet$ & & $\bullet$ & \\ 
        \cite{gao2012spatial} & 2012 & Car & MPC & \cmark & & & $\bullet$ & & $\bullet$ & \\
        \cite{faulwasser2013predictive} & 2013 & Rob. Mani. & MPC & \xmark &$\bullet$  & & & $\bullet$& &  \\
        \cite{frasch2013auto} & 2013 & Car & MPC & \cmark & & & $\bullet$ & $\bullet$ & & \\
        \cite{bock2013real} & 2013 & Crane & MPC & \xmark & $\bullet$ & & & $\bullet$ & & \\
        \cite{liniger2015optimization} & 2015 & Car & MPC & \cmark &$\bullet$  & &   &  &  & $\bullet$ \\
        \cite{faulwasser2015nonlinear} & 2015 & Any & MPC & \cmark &$\bullet$  & &   &$\bullet$  &  &  \\
        \cite{van2016path} & 2016 & Rob. manip. & MPC & \cmark & & & $\bullet$ &  $\bullet$&  & \\
        \cite{verschueren2016time} & 2016 & Rob. manip. & OP & \cmark &  &  & $\bullet$  &   &$\bullet$  & \\
        \cite{kumar2017path} & 2017 & Quadrotor & CL & \xmark &  & $\bullet$ &  & $\bullet$  & &\\ 
        \cite{spedicato2017minimum} & 2017 & Quadrotor & OP & \cmark &  &  & $\bullet$  &  & $\bullet$  & \\
        \cite{brito2019model} & 2019 & Car & MPC & \cmark & $\bullet$ & & & & & $\bullet$ \\
        \cite{kloeser2020nmpc} & 2020 & Car & MPC & \cmark &  &  $\bullet$ & & $\bullet$ &  & \\
        \cite{reiter2021mixed} & 2021 & Car & MPC & \cmark &  &$\bullet$ & &  $\bullet$ &  & \\
        \cite{ji2021cmpcc} & 2021 & Quadrotor & MPC & \cmark & $\bullet$ & & & & & $\bullet$ \\
        \cite{arrizabalaga2022towards} & 2022 & Quadrotor & MPC & \cmark &   &  $\bullet$ &  &  $\bullet$ &  & \\
        \cite{romero2022model} & 2022 & Quadrotor & MPC & \xmark & $\bullet$&  & &  &  & $\bullet$ \\
        \cite{arrizabalaga2022spatial} & 2022 & Particle & MPC & \cmark &  & $\bullet$ &  &  &  & $\bullet$ \\
        \cite{wurman2022outracing} & 2022 & Quadrotor & RL & \cmark & $\bullet$ & & &  &  & $\bullet$ \\
        \cite{arrizabalaga2023pose} & 2023 & Particle & CL & \xmark & $\bullet$ &  &   &  $\bullet$& & \\
        \cite{arrizabalaga2023sctomp} & 2023 & Car, Quadrotor & OP & \cmark & &  & $\bullet$   &  &  $\bullet$ & \\
        \cite{reiter2023frenet} & 2023 & Car & MPC & \cmark & &$\bullet$ & & $\bullet$ &  & \\
        \cite{fork2023euclidean} & 2023 & Quadrotor & OP & \cmark & & & $\bullet$ &  & $\bullet$  & \\
        \cite{kaufmann2023champion} & 2023 & Quadrotor & RL & \cmark & $\bullet$ & & &  &  & $\bullet$ \\
        \cite{oelerich2024boundmpc} & 2024 & Rob. manip. & MPC & \cmark & &$\bullet$ & &  $\bullet$&  & \\
        \cite{krinner2024time} & 2024 & Quadrotor & MPC & \cmark & $\bullet$& & &  &  & $\bullet$ \\
        \hline
	\end{tabular}
\end{adjustbox}
$^1$ Any system that is feedback linearizable\\
\end{table}
\end{minipage}

\end{sidebarGray}
These findings ushered in a new era, shifting path-parametric methods from classical closed-form solutions to optimization-based formulations. This transition is constituted by two separate lines of work which laid the foundations for multiple applications and extensions, ultimately becoming the top contributors for the attention that path-parametric methods receive today. On the one hand, in the seminal work~\cite{verscheure2009time}, it was shown that traversing a path in minimum time with a robotic manipulator -- originally tackled in~\cite{shin1985minimum,pfeiffer1987concept} -- can be formulated as a convex optimization problem. On the other hand, ~\cite{faulwasser2009model,lam2010model} shed some light on the appealing attributes that result from embedding the tangent and orthogonal error components -- originally introduced in~\cite{nelson1988local,kanayama1990stable,cox1991blanche} -- into an optimization problem that is solved in a receding horizon (or MPC) fashion. For the first time, this provided the capacity to deviate from the path, allowing the user to trade-off between traversability and tracking accuracy. Despite the technical differences between~\cite{lam2010model} and~\cite{faulwasser2009model}, this line of work is commonly referred to as Contouring Control, or MPCC in the specific context of MPC.

These two lines of works laid the foundations that, combined with advances in numerical solvers and an increases in computational power, led to a boom in the applicability of path-parametric methods, especially in the context of receding horizon optimization-based control algorithms (MPC). Most of the successful applications of path-parametric methods in real-world systems can be found in this family of works. Additionally, the advent of Reinforcement Learning (RL) brought yet another use to the path-parametric problem, where the notion of progress and other geometric features inherent to the path-parametric methods became very appealing for the design of the reward function. Compared to the previous solutions, most of these methods allowed deviations from the path by relying on collision-free tunnels, tubes or corridors. This led to a paradigm shift, where instead of committing to a given path, any trajectory within the admissible space is valid, a very attractive feature for case studies with constrained task spaces. As a consequence, the resulting literature broadened across a wide range of applications, such as autonomous driving~\cite{kehrle2011optimal,gao2012spatial,frasch2013auto,liniger2015optimization,brito2019model,kloeser2020nmpc,reiter2023frenet}, autonomous agile flight~\cite{spedicato2017minimum,ji2021cmpcc,arrizabalaga2022towards, romero2022model,fork2023euclidean,kaufmann2023champion}, robotic manipulators~\cite{verschueren2016time, van2016path} or more exotic cases like cranes~\cite{bock2013real}. 


From the overview in this subsection it is apparent that path-parametric methods have progressed significantly, spanning across various applications and methodologies. To illustrate this clearly, we have categorized all the aforementioned works in Table~\ref{tab:param_methods}. From this table, we characterize path-parametric methods according to three observations: First, they are a combination of the system state representation and the navigation criteria explained in the previous subsection, and thus, can be grouped accordingly. Second, the rise of new gradient-based techniques within planning and control, such as optimization and learning, have driven the evolution of path-parametric methods. Third, these methods have become more sophisticated over time, allowing for deviations from the path, and thereby, increasing the applicability to a wider range of problems.

\subsection{A universal formulation for all path-parametric methods}
As discussed earlier, path-parametric methods have evolved to meet the diverse needs of the planning and control communities. This evolution has significantly expanded their application across various domains, resulting in a wide array of real-world implementations. However, this has resulted in a more disjointed literature, with each method often presented in isolation. This fragmented approach has created a scattered body of knowledge where these techniques are perceived as distinct entities, making it challenging for readers to grasp their interconnectedness and broader implications.

To address this gap, this paper aims to demonstrate the unified nature of path-parametric planning and control methods. We show how these approaches are inherently linked by introducing a universal formulation. Central to this formulation are two fundamental components that underpin all existing works in the literature. These components are: (i) the method of path parameterization, which defines how the desired geometric reference is articulated, and (ii) the representation of system dynamics relative to this path parameterization, crucial for understanding how the system behaves along the specified path. 

Currently, the literature lacks a generic treatment of these components, contributing to its fragmented nature. By establishing a generalized approach that transcends specific methodologies, we aim to unify path-parametric techniques into a single framework that encompasses everything from traditional path following control laws to more advanced optimization and learning-based methods. In the remainder of the manuscript, we delve into the details necessary to formulate these two foundational ingredients, while ensuring that they meet the demands outlined earlier. Through this unified framework, we seek to not only consolidate existing knowledge but also facilitate the future development and application of path-parametric methods across diverse fields.

\subsection*{Notation}
We will use $\dot{(\cdot)} = \dv{(\cdot)}{t}$ for time derivatives and $(\cdot)' = \dv{(\cdot)}{\theta}$ for differentiating over path parameter $\theta$.

\section{Path parameterizing the reference path}~\label{sec:path_parameterizing_the_reference_path}
In this section, we focus on the \mybox{myLightYellow}{\emph{first ingredient}} of the universal framework, namely the method used to transcribe the reference path into a parametric function. We do this in three separate steps: First, we formally define the geometric reference as a parametric function with a moving frame attached to it. Second, we refer to the problem of computing this frame, by considering the existing options, and third, we present an efficient, simple and compact algorithm for this task.

\subsection{Assigning a path parameter to the reference path}

Existing autonomous navigation systems define the reference path either as a set of waypoints~\cite{foehn2021time} or as a parametric function determined by a higher-level planner~\cite{tordesillas2019faster,zhou2021raptor}. In the first scenario, where the reference is characterized by a collection of waypoints $\bm{wp} = \left[\bm{wp}_1, \bm{wp}_2, \ldots, \bm{wp}_n\right]\in\mathbb{R}^{3\times n}$, the path parameterization is done by interpolating these waypoints with a smooth curve, as in~\cite{wang2002arc}. In the second scenario, this interpolation step is unnecessary because the higher-level planner directly provides a predetermined parametric function formulation, such as B-splines. In either case, the reference path is ultimately expressed as a smooth parametric function $\bm{\gamma}\,:\,\mathbb{R}\mapsto\mathbb{R}^3$, dependent on the \emph{path parameter} $\theta$. This implies that the \emph{arc-length} of the reference is given by
\begin{equation}
l(\theta) = \int_{\theta_0}^{\theta} ||{\bm{\gamma}}^{'}(\theta)||\, d\theta,
\end{equation}
highlighting a distinction often misunderstood in literature, namely, that the arc-length $l(\theta)$ is distinct from the path parameter $\theta$. To better understand this concept, we define the term inside the integral as the \emph{parametric speed}:
\begin{equation}\label{eq:parametric_speed}
\sigma(\theta) = ||\bm{\gamma}^{'}(\theta)||.
\end{equation}
Intuitively, the parametric speed captures the variation of the path parameter $\theta$ with respect to the arc-length $l(\theta)$. Hence, only when $\sigma = 1$ do the arc-length and the path parameter coincide, resulting in a reference path parameterized by its arc-length\footnote{For further details on how to path parameterize the reference path by its arc-length, please refer to~\cite{wang2002arc}}.

\subsection{Assigning a moving frame to the reference path}
The ability to decouple the system dynamics into tangential and orthogonal components relative to the reference path is fundamental to path-parametric formulations. This decoupling facilitates the design of algorithms that guide the system along the path while limiting deviations from it. To achieve this, it is necessary to define a local frame that evolves with the path. As such, we augment the position function $\bm{\gamma}(\theta)$ by attaching a moving frame whose rotation matrix is given by another parametric function $\mathrm{R}\,:\,\mathbb{R}\mapsto\mathbb{R}^{3x3}$ also parameterized by $\theta$. Combining it with the position function $\bm{\gamma}(\theta)$ introduced in the previous subsection formally defines the \emph{geometric reference} as:
\begin{equation}\label{eq:geom_ref}
    \Gamma = \{\theta \in[\theta_0,\theta_f] \subseteq\mathbb{R}\mapsto\bm{\gamma}(\theta) \in \mathbb{R}^3, \mathrm{R}(\theta) \in \mathbb{R}^{3x3}\}\,.
\end{equation}
We refer to the moving frame $\mathrm{R}(\theta) = \left[\bm{e_1}(\theta),\bm{e_2}(\theta),\bm{e_3}(\theta)\right]$ as the \emph{path-frame} $(\cdot)^\Gamma$ and is assumed to be \emph{adapted}, i.e., the first component of the frame coincides with the path's tangent $\bm{e_1}(\theta) = \frac{\bm{\gamma}'(\theta)}{||\bm{\gamma}'(\theta)||}=\frac{\bm{\gamma}'(\theta)}{\sigma(\theta)}$. The change of this frame with respect to the path parameter is determined by the angular velocity $\bm{\omega}(\theta)=\left[\omega_1(\theta),\omega_2(\theta),\omega_3(\theta)\right]$, which can also be represented in the path-frame as $\bm{\omega}^{\Gamma}(\theta)=\left[\omega^\Gamma_1(\theta), \omega^\Gamma_2(\theta), \omega^\Gamma_3(\theta)\right]$:
\begin{multline}\label{eq:ang_vel}
    \bm{\omega}(\theta) = \bm{\omega}^{\Gamma}(\theta)\,\mathrm{R}^{\intercal}(\theta)=\\ \omega^\Gamma_1(\theta)\bm{e_1}(\theta) + \omega^\Gamma_2(\theta)\bm{e_2}(\theta) + \omega^\Gamma_3(\theta)\bm{e_3}(\theta).
\end{multline}
In either case, the motion of the moving frame $\mathrm{R}(\theta)$ is given by
\begin{multline}\label{eq:R_ode}
    \mathrm{R}'(\theta) = \overbrace{\begin{bmatrix}
	0 & -\omega_3(\theta) &\omega_2(\theta) \\ \omega_3(\theta) &0 &-\omega_1(\theta)\\ -\omega_2(\theta) &\omega_1(\theta) &0
	\end{bmatrix}}^{\Omega(\theta)}\mathrm{R}(\theta)\,\equiv\,\\
    \mathrm{R}(\theta)\underbrace{\begin{bmatrix}
	0 & -\omega^\Gamma_3(\theta) &\omega^\Gamma_2(\theta) \\ \omega^\Gamma_3(\theta) &0 &-\omega^\Gamma_1(\theta)\\ -\omega^\Gamma_2(\theta) &\omega^\Gamma_1(\theta) &0
	\end{bmatrix}}_{\Omega^{\Gamma}(\theta)}
\end{multline}
where $\Omega(\theta)$ and $\Omega^{\Gamma}(\theta)$ are the skew symmetric matrices associated to the angular velocity vectors in world-frame $\bm{\omega}(\theta)$ and path-frame $\bm{\omega}^{\Gamma}(\theta)$, respectively\footnote{For a detailed proof of the equivalence in eq.~\eqref{eq:R_ode}, please see Theorem 1 in~\cite{zhao2016time}}. 
From \eqref{eq:R_ode}, it follows that the components of the angular velocity in the moving frame are given by
\begin{multline}\label{eq:angvel_comp}
    \left[\omega^\Gamma_1(\theta), \omega^\Gamma_2(\theta), \omega^\Gamma_3(\theta)\right] =\\ \left[\bm{e_2}'(\theta)\,\bm{e_3}(\theta), \bm{e_3}'(\theta)\,\bm{e_1}(\theta), \bm{e_1}'(\theta)\,\bm{e_2}(\theta) \right]
\end{multline}
As will become apparent in the upcoming section, the angular velocities of the moving frames correspond to the \emph{curvature} $\kappa$ and \emph{torsion} $\tau$ of the path, and their computation is highly dependant on the choice of the moving frame. Subsequently, we will delve into the different options for describing the moving frame, and ultimately present an efficient, simple and compact algorithm for their computation.



\begin{sidebarGray}{An overview of moving frames: Frenet-Serret, Euler Rodrigues, and Parallel Transport}
\summaryinitial{A}s highlighted in the seminal work of Bishop, \emph{there is more than one way to frame a curve}~\cite{bishop1975there}. Aiming to provide an updated perspective on this topic, we focus on the most prominent frames: the Frenet-Serret Frame (FSF), the Euler-Rodrigues Frame (ERF), and the Parallel Transport Frame (PTF). We will explore their underlying mathematical foundations and the properties they inherit as a result of their respective formulations.
\textbf{}
\begin{table}[H] 
	\centering
    \label{tab:frame_choices}
	\begin{tabular}{|c||c|c|c|}
	    \hline
	    Frame& Singularity-free & Twist-free \\
		\hline
		Frenet Serret & \xmark  & \xmark  \\
        \hline
        Euler Rodrigues &\cmark  &\xmark\\
        \hline
		Parallel Transport &\cmark   &\cmark\\
		\hline
	\end{tabular}
\end{table}
\subsubsection{Frenet Serret Frame (FSF)}
The Frenet Serret frame is defined by imposing the second component to be the normalized derivative of the tangent, and the third component othogonal to the first and second.
\begin{subequations}\label{eq:fsf_def}
\begin{equation}
    \bm{e_2}(\theta) = \frac{\bm{e_1}^{'}(\theta)}{||\bm{e_1}^{'}(\theta)||},\quad \bm{e_3}(\theta) = \bm{e_1}(\theta) \cross \bm{e_2}(\theta)\,.
\end{equation}
Combining these with the tangent component $\bm{e_1}$, we can explicitly define the FSF by the path function $\gamma(\theta)$ and its derivatives:
\begin{align}
    &\bm{e_1}(\theta) =\frac{\bm{\gamma}^{'}(\theta)}{||\bm{\gamma}^{'}(\theta)||},\quad 
    \bm{e_2}(\theta) = \frac{\bm{\gamma}^{'}(\theta)\cross\left(\bm{\gamma}^{''}(\theta)\cross\bm{\gamma}^{'}(\theta)\right)}{||\bm{\gamma}^{'}(\theta)||\cdot||\bm{\gamma}^{''}(\theta)\cross\bm{\gamma}^{'}(\theta)||}\,,\notag\\
    &\bm{e_3}(\theta) =\frac{\bm{\gamma}^{'}(\theta)\cross \bm{\gamma}^{''}(\theta)}{||\bm{\gamma}^{'}(\theta)\cross \bm{\gamma}^{''}(\theta)||}\,.
\end{align}
\end{subequations}
Given that the frame is exclusively dependant on the derivatives at the evaluating point, FSF is a \emph{local frame}, i.e., the frame can be evaluated by only relying on local derivative information. Notice that the second component is not defined when $\gamma^{'}$ and $\gamma^{''}$ are parallel. To understand the meaning of this, we need to calculate the angular velocity of the frame $\bm{\omega}_{\text{FSF}}$. To compute it, we combine the eqs. in~\eqref{eq:ang_vel} with~\eqref{eq:fsf_def}, resulting in
\begin{subequations}\label{eq:angvel_fsf}
    \begin{flalign}
    &\omega_{\text{FSF},1}^{\Gamma} = \bm{e_2}^{'}(\theta)\bm{e_3}(\theta) = \sigma(\theta)\tau(\theta)\,,\\
    &\omega_{\text{FSF},2}^{\Gamma} = \bm{e_3}^{'}(\theta)\bm{e_1}(\theta) = 0\,,\\
    &\omega_{\text{FSF},3}^{\Gamma} = \bm{e_1}^{'}(\theta)\bm{e_2}(\theta) = \sigma(\theta)\kappa(\theta)\,.
    \end{flalign}
where the first and third components relate to the \emph{torsion} $\tau(\theta)$ and \emph{curvature} $\kappa(\theta)$, respectively. Intuitively, these express how the curve twists (corridor) over the first component or bends over the third component (road). Mathematically, they are given by:
\begin{align}
    &\tau(\theta) = \frac{\omega^\Gamma_{\text{FSF},1}(\theta)}{\sigma(\theta)} =\frac{||(\bm{\gamma}^{'}(\theta)\cross\bm{\gamma}^{''}(\theta))\cdot\bm{\gamma}^{'''}(\theta)||}{||\bm{\gamma}^{'}(\theta)\cross\bm{\gamma}^{''}(\theta)||^{2}}\,,\\
    &\kappa(\theta) =  \frac{\omega^\Gamma_{\text{FSF},3}(\theta)}{\sigma(\theta)} = \frac{||\bm{\gamma}^{'}(\theta)\cross\bm{\gamma}^{''}(\theta)||}{||\bm{\gamma}^{'}(\theta)||^{3}}\,.
\end{align}
Notice that the parametric speed $\sigma(\theta)$ is decoupled from the curvature $\kappa(\theta)$ and torsion $\tau(\theta)$, proving that they are agnostic to the underlying path-parameterization. Putting all together, we can compute the angular velocity of the frame as:
\begin{align}
    \bm{\omega}_{\text{FSF}}(\theta) &= \omega^\Gamma_{\text{FSF},1}(\theta)\bm{e_1}(\theta) + \omega^\Gamma_{\text{FSF},3}(\theta)\bm{e_3}\notag\\
    &=\sigma(\theta)\left(\tau(\theta)\bm{e_1}(\theta) + \kappa(\theta)\bm{e_2}(\theta)\right)\,.
\end{align}
\end{subequations}
From~\eqref{eq:angvel_fsf}, it is apparent that the FSF is characterized by $\omega^\Gamma_{\text{FSF},2}=0$ (second component always pointing to the center of the curve). This implies that (i) it has singularities when curvature vanishes and (ii) the frame flips when the path transitions from left to right turn. Additionally, it contains a twist along the tangent component, which results in a non-realistic motion of the frame. Therefore we look into alternative moving frames.

\subsubsection*{Euler Rodrigues Frame (ERF)}
The ERF presents a potential solution to the limitations of the FSF. It is fully defined, i.e., has no singularities even when the reference path is straight, and it is a local frame, implying that it can be calculated by exclusively relying on local derivative information. However, the ERF is constructed upon a Pythagorean Hodograph (PH) curve, a subset of polynomials characterized by the property that the parametric speed $\sigma(\theta)$ is a polynomial on the path parameter $\theta$. Imposing this condition, results in a family of polynomials whose geometric features, such as the arclength, curvature, torsion, etc. can be computed in closed form, by exclusively relying on rational functions. Another benefit of these polynomials is that they inherit an adapted frame, the ERF, which is fully defined and, for a given PH curve, their computation is straightforward.

However, converting a given reference path $\bm{\gamma}(\theta)$ into a PH curve is not a trivial task, and may require numerical routines. Moreover, guaranteeing differentiability and continuity of the ERF and its angular velocity further increases the complexity of the conversion. Besides that, the algebra and calculus involved in the computation of PH curves and their corresponding ERFs are highly involved, introducing an additional layer of complexity and  making them less prone to be reproduced by practitioners within the planning and control communities. Given the amount of detail required to derive and compute these curves and frames, they will be omitted in this manuscript. For further information on PH curves and ERFs, see~\cite{farouki2008pythagorean} and \cite{choi2002euler}, respectively. A more applied perspective with navigation applications leveraging PH curves and ERFs can be found in~\cite{arrizabalaga2022spatial,arrizabalaga2023sctomp, arrizabalaga2023pose}. Lastly, if you are interested in the construction of PH curves and ERFs for path-parametric planning and control algorithms, a real-time and two times differentiable algorithm is given in our accompanying manuscript~\cite{arrizabalaga2024phodcos}.
\end{sidebarGray}

\begin{sidebarGray}{\continuesidebar}
\subsubsection*{Parallel Transform Frame (PTF)}
The PTF offers a solution to the shortcomings of the FSF, as well as the ERF. Not only it is fully defined and can be twist-free, but its implementation is also very simple, making it lightweight, efficient and easy to reproduce. These reasons make the PTFs the most suitable choice for path-parametric planning and control algorithms. Their construction is spread out across the literature~\cite{hanson1995parallel,wang2008computation,fork2023euclidean}, but none of them addresses the computation of the moving frame's differentiability and angular velocity. Therefore, in this subsection we present an efficient, simple and compact algorithm that given a parametric curve $\bm{\gamma}(\theta)$, finds the associated PTF and angular velocity $\mathrm{R}_\text{PTF}(\theta)$, $\bm{\omega}_\text{PTF}(\theta)$, while also accounting for the respective derivatives. Before presenting this algorithm, we provide an overview of the PTF.

The PTF originates from the seminal work \emph{There is more than one way to frame a curve}~\cite{bishop1975there}, which raises the realization that the second $\bm{e_2}$ and third $\bm{e_3}$ components of the moving frame can be chosen freely, as long as they remain in the orthogonal plane and form an orthonormal frame with the remaining tangent component $\bm{e_1}$. This allows for choosing the second and third components to form a parallel vector field that only rotates by the necessary amount, so that it is always perpendicular to the tangent component. This is equivalent to imposing the derivative of the second and third components to point in the direction of the tangential unit vector:
\begin{subequations}
\begin{equation}\label{eq:ptf_def}
\bm{e_2}^{'}(\theta) = k_1(\theta)\bm{e_1}(\theta),\quad \bm{e_3}^{'}(\theta) = k_2(\theta)\bm{e_1}(\theta)\,.
\end{equation}
To compute the auxiliary variables $k_1$ and $k_2$, we combine~\eqref{eq:ptf_def} with the orthonormality conditions
\begin{flalign}
    0 &= 1-\bm{e_2}^{\intercal}(\theta)\bm{e_2}(\theta),\quad 0 = 1-\bm{e_3}^{\intercal}(\theta)\bm{e_3}(\theta)\,\\
    0 &= \bm{e_1}^{\intercal}(\theta)\bm{e_2}(\theta),\quad 0 = \bm{e_1}^{\intercal}(\theta)\bm{e_3}(\theta)\,.\label{eq:ptf_def_last}
\end{flalign}
\end{subequations}
Differentiating~\eqref{eq:ptf_def_last} with path parameter $\theta$, combining it with~\eqref{eq:ptf_def} and multiplying with $\bm{e_1}(\theta)$, we get
\begin{equation*}
    k_1(\theta)=-{\bm{e}_{1}^{'}(\theta)}^{\intercal}\bm{e_2}(\theta)\,,\quad k_2(\theta)=-{\bm{e}_{1}^{'}(\theta)}^{\intercal}\bm{e_3}(\theta)\,,
\end{equation*}
which ultimately leads to
\begin{flalign}\label{eq:ptf_eqmo}    
\bm{e_2}^{'}(\theta) = \left(-{\bm{e}_{1}^{'}(\theta)}^{\intercal}\bm{e_2}(\theta)\right)\bm{e_1}(\theta),\notag\\
\bm{e_3}^{'}(\theta) = \left(-{\bm{e}_{1}^{'}(\theta)}^{\intercal}\bm{e_3}(\theta)\right)\bm{e_1}(\theta)\,.
\end{flalign}
Eqs.~\eqref{eq:ptf_eqmo} provide the key insight required to compute the angular velocity of the PTF. By merging them with the definitions of the angular velocity for a generic moving frame in~\eqref{eq:angvel_comp}, we get
\begin{subequations}\label{eq:angvel_ptf}   
\begin{flalign}
    \omega^\Gamma_{\text{PTF},1}(\theta) &= \bm{e_2}^{'}(\theta)\bm{e_3}(\theta)=\left(-{\bm{e}_{1}^{'}(\theta)}^{\intercal}\bm{e_2}(\theta)\bm{e_1}(\theta)\right)\bm{e_3}(\theta)=0\,,\notag\\
    \omega^\Gamma_{\text{PTF},2}(\theta) &= \bm{e_3}^{'}(\theta)\bm{e_1}(\theta)=\left(-{\bm{e}_{1}^{'}(\theta)}^{\intercal}\bm{e_3}(\theta)\bm{e_1}(\theta)\right)\bm{e_1}(\theta)\notag\\
    &=-{\bm{e}_{1}^{'}(\theta)}^{\intercal}\bm{e_3}(\theta)\,,\notag\\
    \omega^\Gamma_{\text{PTF},3}(\theta) &= \bm{e_1}^{'}(\theta)\bm{e_2}(\theta)\,.
\end{flalign}
where $(\cdot)^\Gamma$ indicates that the angular velocity is expressed in the path-frame $\Gamma$. When translated to the world-frame as defined in~\eqref{eq:ang_vel}, it becomes
\begin{multline}\label{eq:angvel_world_ptf}
\bm{\omega}_{\text{PTF}}(\theta) = \bm{\omega}^\Gamma_{\text{PTF}}(\theta)\mathrm{R}^{\intercal}_\text{RMF}(\theta)\\
=\omega^\Gamma_{\text{PTF},2}(\theta)\bm{e_2}(\theta) + \omega^\Gamma_{\text{PTF},3}(\theta)\bm{e_3}(\theta)\,.
\end{multline}
\end{subequations}
From eqs.~\eqref{eq:angvel_ptf}, there are two key insights to be drawn. First, in contrast to the FSF in~\eqref{eq:angvel_fsf}, the angular velocity of the PTF does not have a component along the tangent $\bm{e_1}$, i.e., $\omega^\Gamma_{\text{PTF},1}=0$, and thus, the PTF is twist-free. Second, the angular velocity $\bm{\omega}_{\text{PTF}}(\theta)$ is exclusively dependant on the second and third components of the moving frame $\mathrm{R}_{\text{PTF}}(\theta)$ and the derivative of its tangent $\bm{e_1}^{'}(\theta)$. Thus, if the initial frame is given $\mathrm{R}_{\text{PTF}}(\theta_0)=\left[\bm{e_1}(\theta_0),\bm{e_2}(\theta_0),\bm{e_3}(\theta_0)\right]$, we can compute the PTF at any point along the curve by forward integrating it as
\begin{equation}\label{eq:ptf_so3_integrate}
    \mathrm{R}_{\text{PTF}}(\theta_{i+1}) = e^{\Omega_{\text{PTF}}(\theta_i)\Delta\theta_i}\mathrm{R}_{\text{PTF}}(\theta_{i})\,,
\end{equation}
where $\Delta\theta_i = \theta_{i+1}-\theta_i$ and $\Omega_{\text{PTF}}(\theta_i)$ refers to the skew symmetric matrix associated with the angular velocity $\bm{\omega}_\text{PTF}$, as defined in~\eqref{eq:R_ode}. Notice that $\bm{e_1}^{'}(\theta)$ drives the integration forward, and thus, the reference path $\bm{\gamma}(\theta)$ needs to be at least $C^2$. Additionally, conducting the integration within $\mathrm{SO}(3)$, ensures that the frame remains orthonormal, in contrast to other standard methods, such as Euler or Runge-Kutta~\cite{wanner1996solving}.  Algorithm~\ref{algo:ptf_integration} summarizes the steps required to compute the PTF and its angular velocity.

\begin{mdframed}[backgroundcolor=myLightYellow,innertopmargin=0pt,linecolor=strongerYellow,linewidth=1pt,innerrightmargin=8pt,innerleftmargin=8pt,]
\begin{algorithm}[H]
\caption{\textit{Parallel Transport Frame Integr.} (\textbf{PTFI}):\newline 
    \textit{Input:}$\quad\theta_{0,\,\dots\,,N},\,\mathrm{R}_{\text{PTF}}(\theta_0),\,\bm{e_1}^{'}(\theta_{0,\,\dots\,,N})$\newline
    \textit{Output:}$\quad\mathrm{R}_{\text{PTF}}(\theta_{0,\,\dots\,,N}),\,\bm{\omega}_{\text{PTF}}(\theta_{0,\,\dots\,,N})$
}\label{algo:ptf_integration}

\begin{algorithmic}[1]
\Function{PTFI}{$\theta_{0,\,\dots\,,N},\,\mathrm{R}_{\text{PTF}}(\theta_0),\,\bm{e_1}^{'}(\theta_{0,\,\dots\,,N})$}
\For{$i\in\{0,\,...,N-1\}$}
\State $\bm{\omega}_{\text{PTF},i}\gets \textsc{AngVel}(\bm{e_{1,i}}^{'}\,,\mathrm{R}_{\text{PTF},i})$~\eqref{eq:angvel_ptf}
\State $\Omega\gets\textsc{SkewMatrix}(\bm{\omega}_{\text{PTF},i})$~\eqref{eq:R_ode}
\State $\Delta \theta = \theta_{i+1}-\theta_i$
\State $\mathrm{R}_{\text{PTF},i} \gets \textsc{Integr}(\mathrm{R}_{\text{{PTF}},i},\Omega, \Delta\theta)$~\eqref{eq:ptf_so3_integrate}
\EndFor
\State\Return $\mathrm{R}_{\text{PTF}}(\theta_{0,\,\dots\,,N}),\,\bm{\omega}_{\text{PTF}}(\theta_{0,\,\dots\,,N})$
\EndFunction
\end{algorithmic}
\end{algorithm}

\end{mdframed}

Having obtained the PTF components $\mathrm{R}_\text{PTF}$ and angular velocity $\bm{\omega}_\text{PTF}$, we can now proceed to computing the respective higher derivatives. Given that differentiable path-parametric methods require from first order (learning) and second order (optimization) gradients, we will focus on the derivation of the first two derivatives. However, the suggested methodology can be extended to higher order derivatives by following the same procedure.

\end{sidebarGray}

\begin{sidebarGray}{\continuesidebar}

Starting with the first order derivatives, $\mathrm{R}^{'}_\text{PTF}$ is readily avaible from~\eqref{eq:R_ode} and~\eqref{eq:angvel_ptf}, while the angular acceleration $\bm{\alpha}_{\text{PTF}}$ can be obtained by derivating~\eqref{eq:angvel_ptf}:
\begin{subequations}\label{eq:angacc_ptf}
    \begin{equation}
        \bm{\alpha}_{\text{PTF}}(\theta) = \bm{\alpha}^\Gamma_{\text{PTF}}(\theta)\mathrm{R}^{\intercal}_\text{PTF}(\theta) + \bm{\omega}^\Gamma_{\text{PTF}}(\theta)\mathrm{R}^{'\,\intercal}_\text{PTF}(\theta)
    \end{equation}
with
\begin{align}
    \begin{split}
    \bm{\alpha}^{\Gamma}_{\text{PTF}}(\theta) = \left[0,
    -\left(\bm{e_1}^{''}(\theta)\bm{e_3}(\theta) + \bm{e_1}^{'}(\theta)\bm{e_3}^{'}(\theta)\right), \right.\\
    \left.\left(\bm{e_1}^{''}(\theta)\bm{e_2}(\theta) + \bm{e_1}^{'}(\theta)\bm{e_2}^{'}(\theta)\right)\right]\,.
    \end{split}
\end{align}
\end{subequations}
To derive the second derivative of the moving frame  $\mathrm{R}^{''}_\text{PTF}$, we express the angular velocity~\eqref{eq:angvel_ptf} and acceleration~\eqref{eq:angacc_ptf} by their skew-symmetric matrices $\Omega_{\text{PTF}}$, $\Omega^{'}_{\text{PTF}}$ and combine them with the derivative of~\eqref{eq:R_ode} over $\theta$, which results in:
\begin{equation}\label{eq:R_ode2}
    \mathrm{R}^{''}_\text{PTF} = \Omega_{\text{PTF}}^{'}(\theta)\mathrm{R}_\text{PTF}(\theta) + \Omega_{\text{PTF}}(\theta)\mathrm{R}^{'}_\text{PTF}(\theta)\\
\end{equation}
Finally, to calculate the second derivative of the angular velocity --denoted as angular jerk $\bm{j}_\Gamma$-- we derivate~\eqref{eq:angacc_ptf}:
\begin{subequations}\label{eq:angjerk_ptf}
\begin{multline}
            \bm{j}_{\text{PTF}}(\theta) = \bm{j}^\Gamma_{\text{PTF}}(\theta)\mathrm{R}^{\intercal}_\text{PTF}(\theta)+\bm{\alpha}^\Gamma_{\text{PTF}}(\theta)\mathrm{R}^{'\,\intercal}_\text{PTF}(\theta) +\\ \bm{\alpha}^\Gamma_{\text{PTF}}(\theta)\mathrm{R}^{'\,\intercal}_\text{PTF}(\theta) + \bm{\omega}^\Gamma_{\text{PTF}}(\theta)\mathrm{R}^{''\,\intercal}_\text{PTF}(\theta)\,,
\end{multline}
where
\begin{align}
    \begin{split}
        &\bm{j}^{\Gamma}_{\text{PTF}}(\theta) = \left[ 0, \right. \\
        &-\left(\bm{e_1}^{'''}(\theta)\bm{e_3}(\theta) + \bm{e_1}^{''}(\theta)\bm{e_3}^{'}(\theta) + \bm{e_1}^{'}(\theta)\bm{e_3}^{''}(\theta) + \bm{e_1}^{'}(\theta)\bm{e_3}^{''}(\theta)\right), \\
        &\left. \bm{e_1}^{'''}(\theta)\bm{e_2}(\theta) + \bm{e_1}^{''}(\theta)\bm{e_2}^{'}(\theta) + \bm{e_1}^{'}(\theta)\bm{e_2}^{''}(\theta) \right]\,.
    \end{split}
\end{align}

\end{subequations}
Eqs.~\eqref{eq:angvel_ptf},~\eqref{eq:angacc_ptf},~\eqref{eq:angjerk_ptf} show how the continuity of the reference path $\bm{\gamma}(\theta)$ and angular frame $\bm{\omega}(\xi)$ are related:
\begin{subequations}\label{eq:angvel_cont}
\begin{gather}
    \bm{e_1}^{'}(\theta),\,\mathrm{R}_{\text{PTF}}(\theta)\rightarrow\bm{\omega}_{\text{PTF}}(\theta)\,,\\
    \bm{e_1}^{''}(\theta),\,\mathrm{R}_{\text{PTF}}(\theta),\, \mathrm{R}^{'}_{\text{PTF}}(\theta)\rightarrow\bm{\alpha}_{\text{PTF}}(\theta)\,,\\
    \bm{e_1}^{'''}(\theta),\,\mathrm{R}_{\text{PTF}}(\theta),\, \mathrm{R}^{'}_{\text{PTF}}(\theta),\mathrm{R}^{''}_{\text{PTF}}(\theta)\rightarrow\bm{j}_{\text{PTF}}(\theta)\,,
\end{gather}
\end{subequations}
i.e., a path with continuity degree $\bm{\gamma}\in C^n$ relates to $\mathrm{R}_\text{PTF}\in C^{n-1}$ and $\bm{\omega}_{\text{PTF}}\in C^{n-2}$. For example, if we desire $\bm{\omega}_{\text{PTF}}$ to be $C^2$, $\bm{\gamma}$ needs to be (at least) $C^4$. The specific steps necessary for computing the first and second order derivatives of the moving frame components and angular velocity are detailed in Algorithm~\ref{algo:ptf_derivatives}.
\begin{mdframed}[backgroundcolor=myLightYellow,innertopmargin=0pt,linecolor=strongerYellow,linewidth=1pt,innerrightmargin=8pt,innerleftmargin=8pt,]
\begin{algorithm}[H]
    \caption{\textit{Parallel Transport Frame Deriv.} (\textbf{PTFD}):\newline 
    \textit{Input:}$\quad\mathrm{R}_{\text{PTF}},\,\bm{\omega}_{\text{PTF}},\,\,\bm{e_1}^{'},\,\bm{e_1}^{''},\,\bm{e_1}^{'''}$ eval. at $\theta_{0,\,\dots\,,N}$\newline
    \textit{Output:}$\quad\mathrm{R}^{'}_{\text{PTF}},\,,\mathrm{R}^{''}_{\text{PTF}},\,\bm{\alpha}_{\text{PTF}},\,\bm{j}_{\text{PTF}}$ eval. at $\theta_{0,\,\dots\,,N}$
}\label{algo:ptf_derivatives}
\begin{algorithmic}[1]
\Function{PTFD}{$\mathrm{R}_{\text{PTF}}(\theta_{0,\,\dots\,,N}),\,\bm{\omega}_{\text{PTF}}(\theta_{0,\,\dots\,,N})$}
\For{$i\in\{0,\,...,N\}$}
\State $\mathrm{R}_\text{PTF,i}^{'} \gets \textsc{FrameVel}(\mathrm{R}_\text{PTF,i},\bm{\omega}_{\text{PTF,i}})${\small~\eqref{eq:R_ode}}
\State $\bm{\alpha}_{\text{PTF,i}} \gets \textsc{AngAcc}(\bm{e_{1,i}}^{''},\mathrm{R}_{\text{PTF},i},\mathrm{R}^{'}_{\text{PTF,i}})${\small~\eqref{eq:angacc_ptf}}
\State $\mathrm{R}_\text{PTF,i}^{''} \gets \textsc{FrameAcc}(\mathrm{R}_\text{PTF,i},\bm{\omega}_{\text{PTF},i}$
\Statex\hspace{13.5em}$\bm{\alpha}_{\text{PTF},i})${\small~\eqref{eq:R_ode2}}
\State $\bm{j}_{\text{PTF},i} \gets \textsc{AngJerk}(\bm{e_{1,i}}^{'''},\mathrm{R}_{\text{PTF,i}},$\Statex\hspace{12em} $\mathrm{R}^{'}_{\text{PTF},i},\mathrm{R}^{''}_{\text{PTF},i})${\small~\eqref{eq:angjerk_ptf}}
\EndFor
\State\Return\hspace{0.2em}$\mathrm{R}^{'}_{\text{PTF}}(\theta_{0,\,\dots\,,N}),\,\mathrm{R}^{''}_{\text{PTF}}(\theta_{0,\,\dots\,,N}),$\Statex\hspace{4.7em}$\,\bm{\alpha}_{\text{PTF}}(\theta_{0,\,\dots\,,N}),\,\bm{j}_{\text{PTF}}(\theta_{0,\,\dots\,,N})$
\EndFunction
\end{algorithmic}
\end{algorithm}
\end{mdframed}
In summary, Algorithms~\ref{algo:ptf_integration} and~\ref{algo:ptf_derivatives} facilitate the computation of the moving frame $\mathrm{R}_\text{RMF}$, angular velocity $\bm{\omega}_\text{RMF}$ and its derivatives $\{\mathrm{R}^{'}_\text{RMF},\,\mathrm{R}^{''}_\text{RMF},\,\bm{\alpha}_\text{RMF},\,\bm{j}_\text{RMF}\}$. This combination of accessibility and efficiency, along with the inherent advantages of the PTFs — namely, being singularity-free and twist-free — renders the proposed moving frame computation method highly suitable for path-parametric planning and control algorithms.
\end{sidebarGray}

\subsection{Choosing a moving frame}
The path-frame associated to the geometric reference $\Gamma$ in~\eqref{eq:geom_ref} plays a crucial role in the formulation of path-parametric methods, since it allows for decoupling the system dynamics into components tangential and orthogonal to the path. This raises the need for a technique to augment the position function $\gamma(\theta)$ into a moving frame $\mathrm{R}(\theta)$, i.e., a method that solely requires a parametric position reference to compute a frame that evolves with it.

This is a well-studied problem with multiple solutions available in the existing literature, each differing based on the choice of the underlying frame. The most common options are the Frenet Serret Frame (FSF)~\cite{struik1961lectures,abbena2017modern}, the Euler Rodrigues Frame (ERF)~\cite{choi2002euler,vsir2007} and the Parallel Transport Frame (PTF)~\cite{bishop1975there,hanson1995parallel,wang2008computation}. From a computation perspective, the first two frames are analytical, as they are given in closed form by local derivative information, while the latter requires from global information, and thus, its computation relies on a numerical routine.

The most common choice in the literature is the FSF due to its analytical simplicity. However, it is undefined when the reference path is a straight line (i.e., when the curvature vanishes) and introduces an unnecessary twist over its first component, causing undesired nonlinearities when projecting system dynamics to the path frame or representing the collision-free space around the path. Within analytical frames, an alternative is the ERF, which remains defined even if the reference path is straight. However, the ERF cannot be guaranteed to be twist-free and relies on a Pythagorean Hodograph curve, whose computation is complex and may require numerical routines. Consequently, the most common alternative to the FSF is the PTF, which is both singularity-free and twist-free but requires numerical computation.

Existing methods to compute PTFs are discrete~\cite{wang2008computation}, and thus do not allow for the computation of higher derivatives. Additionally, they focus exclusively on the computation of the rotation matrix and neglect its angular velocity. However, as explained in Section~\ref{subsec:history}, current path-parametric methods heavily rely on learning and optimization algorithms, which require first and second order derivatives. To address this, in this manuscript we present an efficient, simple, and compact method based on PTFs that computes the components and angular velocity of a singularity-free and twist-free path frame, while also accounting for the computation of the respective derivatives. Aiming to make this manuscript self-contained, before presenting this algorithm, we also provide an overview of the alternative methods.

\subsection{Numerical tests}
To showcase the concepts presented in this section, we provide two numerical tests: First, we compare the FSF to the PTF for a two-dimensional planar curve and a three-dimensional spatial curve. Second, we observe how the continuity of the reference path and the angular velocity of the moving frames are related. 

\begin{figure*}[t]
 \begin{tcolorbox}[colback=myLightYellow, colframe=white, boxrule=0pt, width=\textwidth, enlarge left by=0mm, enlarge right by=0mm, arc=6pt]
	\centering
	\includegraphics[width=\linewidth]{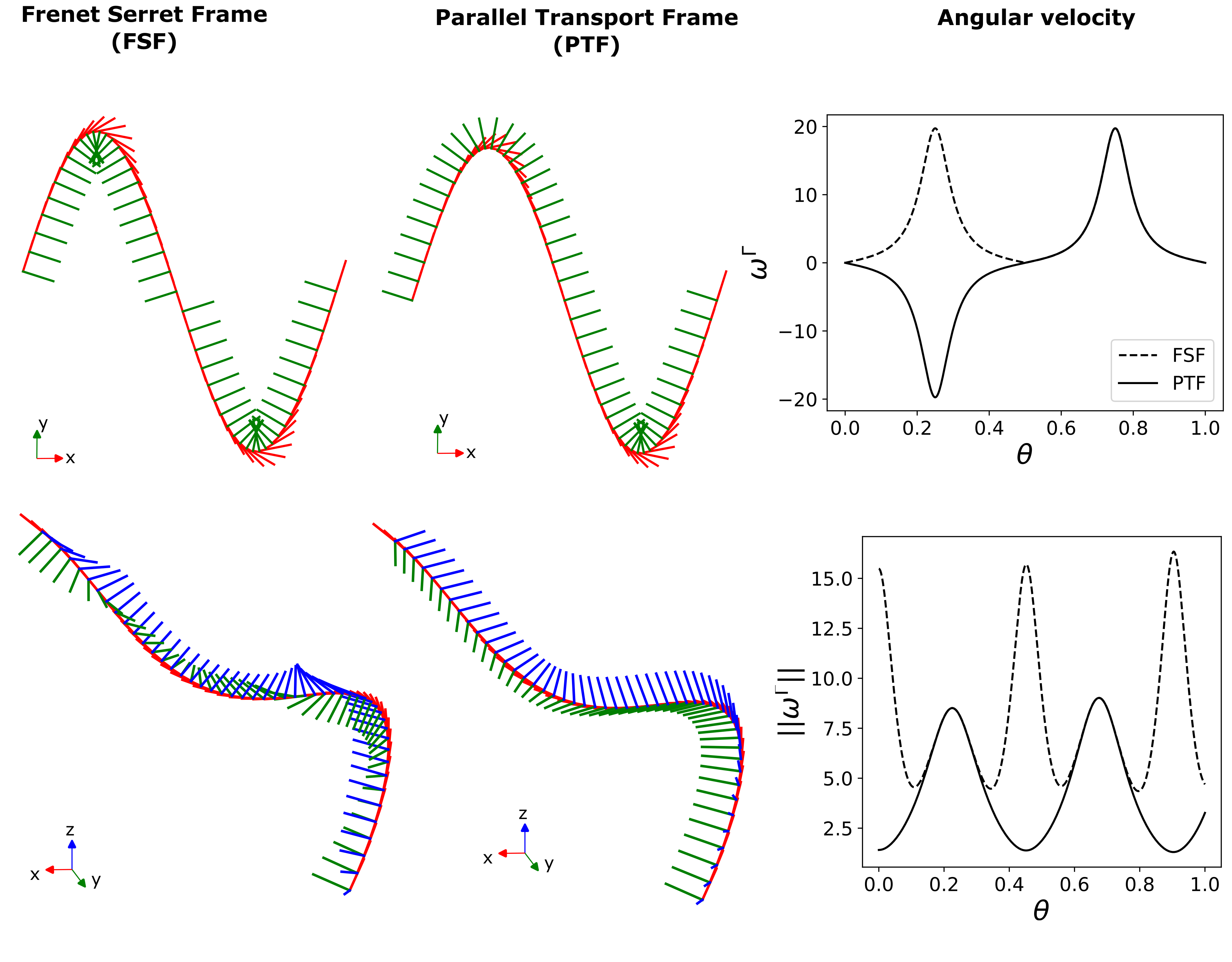} 
\end{tcolorbox}
    \def\figurename{\textcolor{strongerYellow}{\textbf{FIGURE}}} 
    \captionsetup{labelformat=simple, labelfont={color=strongerYellow, bf}} 
	\caption{Comparison of Frenet Serret Frame (FSF, first column) and Parallel Transport Frame (PTF, second column) for a two-dimensional planar curve (first row) and three-dimensional spatial curve (second row). The first, second and third components of the moving frame are shown in red, green and blue, respectively. The third colum shows the angular velocity of the moving frames.}\label{fig:sing_twist}
\end{figure*}

In Fig.~\ref{fig:sing_twist} we show a 2D (top row) and 3D (bottom row) comparison between the FSF (first column) and the PTF (second column). The planar comparison is based on the curve $\bm{\gamma}(\theta)=\left[\gamma, \sin(2\pi\theta)\right]$ and clearly depicts how the FSF presents a singularity in the inflection point, where the frame is not defined and its normal component abruptly flips. The spatial curve is given by
\begin{flalign*}
    \bm{\gamma}(\theta)=[(0.6 + 0.3\cos(\theta))\cos(2\theta), 
    \\(0.6 + 0.3\cos(\theta))\sin(2\theta),
    \\ 0.3\sin(7\theta)]\,,
\end{flalign*}
and showcases how the FSF presents an undesired rotation over its first component. In contrast, the PTF does not suffer from this twist, ultimately leading to a moving frame with a lower angular velocity magnitude (third column).

\begin{figure*}[!ht]
	\centering
 \begin{tcolorbox}[colback=myLightYellow, colframe=white, boxrule=0pt, width=\textwidth, enlarge left by=0mm, enlarge right by=0mm,arc=6pt]
	\includegraphics[width=\linewidth]{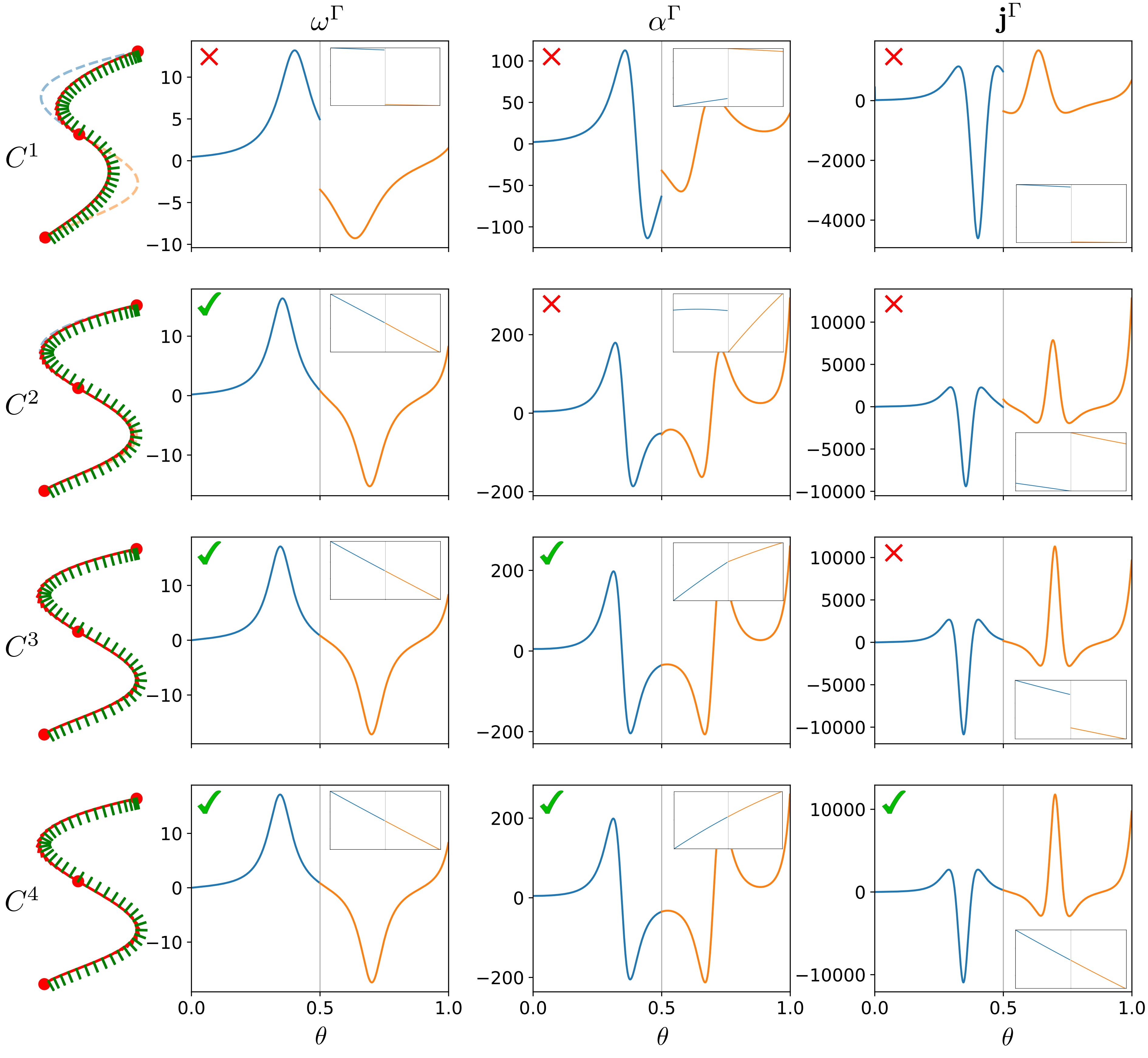}
\end{tcolorbox}
    \def\figurename{\textcolor{strongerYellow}{\textbf{FIGURE}}} 
    \captionsetup{labelformat=simple, labelfont={color=strongerYellow, bf}} 
 \caption{Numerical validation of the continuity analysis conducted in eqs.~\eqref{eq:angvel_cont}, namely that a reference curve $\bm{\gamma}$ that is $C^n$ relates to an angular velocity $\bm{\omega}_\text{PTF}$ that is $C^{n-2}$. For this purpose we divide an exemplary curve into two sections and interpolate with various continuity degree ($C^0$ to $C^4$ from top to bottom). The left column shows the exemplary curve, while the remaining columns depict the angular velocity $\bm{\omega}^\Gamma$, acceleration $\bm{\omega}^\Gamma$ and jerk $\bm{\omega}^\Gamma$. The intersection is given by the red dot located in the middle of the curve at $\theta=0.5$. The evaluations associated to the first and second sections are depicted in blue and orange. The boxes in the upper right side of each plot provide a mode detailed look into the intersection, allowing us to differ the continuous and discontinuous cases. Additionally, the continuous cases have been labelled by a green tick, while the discontinuous ones are marked by a red cross.}\label{fig:angvel_cont}
\vspace{-2mm}
\end{figure*}

Regarding the continuity analysis, in eqs.~\eqref{eq:angvel_cont} it was concluded that a reference path $\bm{\gamma}$ that is $C^n$ relates to an angular velocity $\bm{\omega}_\text{PTF}$ that is $C^{n-2}$. To numerically validate this statement, we divide an illustrative curve $\bm{\gamma}(\theta)=\left[0.5\cos(9\theta),e^{\cos(1.8\theta)}\right]$ into two sections, conduct interpolations of different degrees and observe the continuity of the angular velocity. The obtained results are shown in Fig.~\ref{fig:angvel_cont} and confirm the theoretical analysis: When the interpolation is $C^2$, the angular velocity is $C^0$ (only $\bm{\omega}_\text{PTF}$ is continuous); when the interpolation is $C^3$, the angular velocity is $C^1$ ($\bm{\omega}_\text{PTF}, \bm{\alpha}_\text{PTF}$ are continuous); and when the interpolation is $C^4$, the angular velocity is $C^2$ ($\bm{\omega}_\text{PTF}, \bm{\alpha}_\text{PTF}, \bm{j}_\text{PTF}$ are continuous).

\section{Path parameterizing the Cartesian coordinates}~\label{sec:path_parameterizing_the_cartesian_coordinates}
In this section, we concentrate on the \mybox{myLightYellow}{\emph{second ingredient}} of the universal framework, specifically a parametric reformulation that projects the Cartesian system dynamics into the spatial states associated with a parametric path. 

This is addressed in three distinct steps: First, we formally define the spatial states, as an alternative representation of the Cartesian coordinates. Second, we derive the equations of motion for this representation without making any assumptions regarding the underlying path parameterization. This ensures full compliance with the methods outlined in the previous section and demonstrates that existing formulations in the literature are specific cases of the presented derivation. Lastly, we discuss how the parametric terms required by the derived equations of motion seamlessly integrate with the algorithms introduced in the previous section.

\subsection{Spatial states: An alternative to Cartesian coordinates}
We consider continuous time, (non)linear dynamic systems of the form
\begin{subequations}\label{eq:dynamical_system}
\begin{equation}\label{eq:dynamical_system_f}
    \dot{\bm{x}}(t) = f(\bm{x}(t),\bm{u}(t)),
\end{equation}
where $\bm{x}\in\mathbb{R}^{n_x}$ and $\bm{u}\in\mathbb{R}^{n_u}$ define the system states and inputs, respectively. We assume that the Cartesian coordinates $\bm{p}^W(t) \in \mathbb{R}^3$ associated to the system's longitudinal location with respect to a world-frame $(\cdot)_{W}$ in the Euclidean space are given by a (non)linear mapping $h$, such that
\begin{equation}\label{eq:dynamical_system_cartesian}
    \bm{p}^{\text{W}}(t) = h(\bm{x}(t))\,.
\end{equation}
\end{subequations}
\noindent To project the system dynamics~\eqref{eq:dynamical_system_f} onto the reference $\Gamma$ in~\eqref{eq:geom_ref}, we introduce the \emph{spatial coordinates} as an alternative to the Cartesian representation in~\eqref{eq:dynamical_system_cartesian}. For this purpose, we decouple the system's translational motion into two terms: a \emph{tangent} element -- describing the progress along the path -- and a \emph{transverse} component -- representing the distance perpendicular to the path --.

\begin{figure}
	\centering
    \begin{tcolorbox}[colback=myLightGray, colframe=white, boxrule=0pt, enlarge left by=0mm, enlarge right by=0mm, arc=6pt]
	\includegraphics[width=1.15\linewidth]{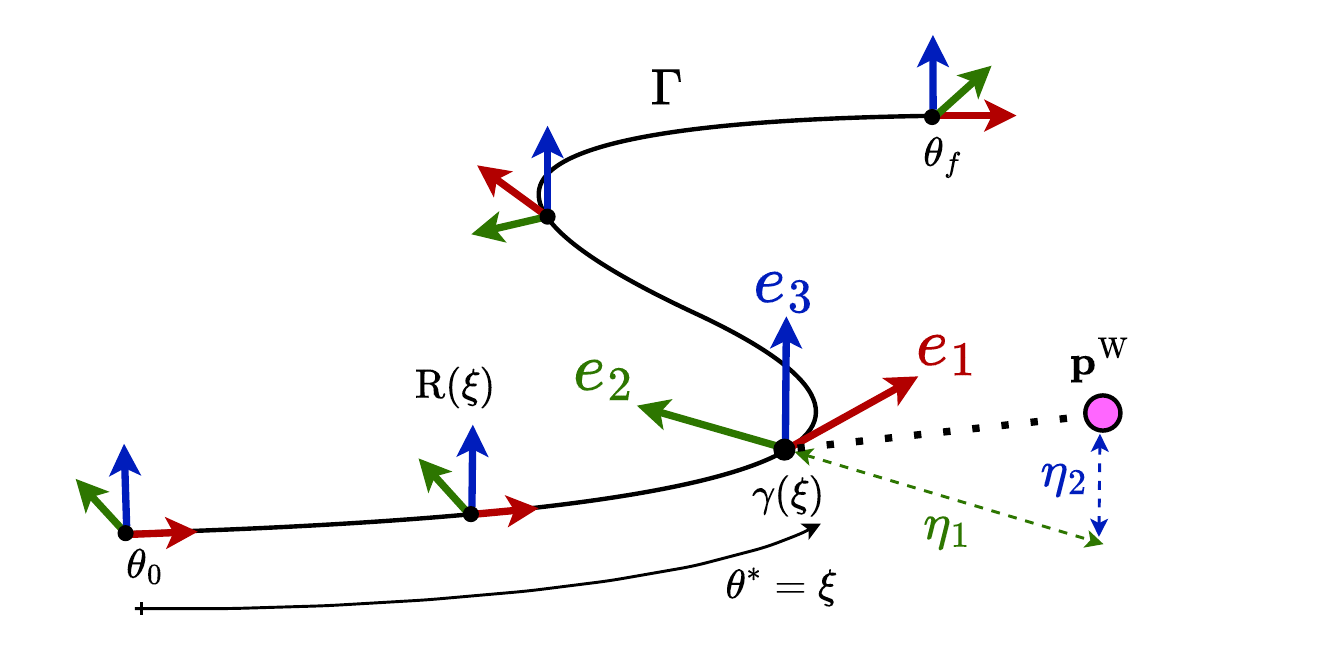}
	\end{tcolorbox}
    \def\figurename{\textcolor{strongerGray}{\textbf{FIGURE}}} 
    \captionsetup{labelformat=simple, labelfont={color=strongerGray, bf}} 
 \caption{Spatial projection of the three-dimensional Cartesian coordinates $\bm{p}^W$, represented by the pink dot, onto a geometric path $\Gamma$ with an associated adapted-frame $\text{R}(\xi) = \{\bm{e_1}(\xi),\bm{e_2}(\xi),\bm{e_3}(\xi)\}$. The distance to the closest point on the path $\bm{\gamma}(\xi)$ is decomposed into the transverse coordinates $\bm{\eta} = \left[\eta_1,\eta_2\right]$.}\label{fig:spatial_coordinates}
\end{figure}

Given the dynamical system's Cartesian location $\bm{p}^W(t)$ in~\eqref{eq:dynamical_system_cartesian}, we define the \emph{progress variable} $\xi(t)$ as the path parameter $\theta$ of the closest point in the reference $\Gamma$, i.e., 
\begin{equation}\label{eq:xi}
\xi(t) =\theta^*(t)= \text{arg}\min_{\theta}\frac{1}{2}||\bm{p}^W(t)-\bm{\gamma}(\theta)||^2\,.
\end{equation}
An explicit representation of the progress variable $\xi(t)$ as in~\eqref{eq:xi} allows for expressing the distance between the dynamical system and the reference path, in both the world-frame and the path-frame:
\begin{subequations}\label{eq:dist}
\begin{equation}
    \bm{d}^{\Gamma}(t) = \mathrm{R}(\txi)^\intercal\bm{d}^{\text{W}}(t)\,,
\end{equation}
where
\begin{equation}
    \bm{d}^{\text{W}}(t)=\bm{p}^W(t)-\bm{\gamma}(\txi)\,.
\end{equation}
\end{subequations}
Since we have assumed the path-frame $\mathrm{R}(\txi)$ to be adapted, the first element of $\bm{d}^{\Gamma}(t)$, is zero, while the remaining two are the perpendicular projections, which refer to the aforementioned transverse component $\bm{\eta}(t)=\left[\eta_1(t),\eta_2(t)\right]$. Consequently, eq.~\eqref{eq:dist} simplifies into
\begin{equation}\label{eq:transverse}
\bm{d}^{\Gamma}(t) = \left[0,\bm{\eta}(t)\right] = \left[0, \bm{e_2}(\txi)\bm{d}^{\text{W}}(t), \bm{e_3}(\txi)\bm{d}^{\text{W}}(t)\right]^\intercal\,.
\end{equation}
Finally, combining the progress variable $\xi(t)$ in~\eqref{eq:xi} with the transverse coordinates $\bm{\eta}$(t) in~\eqref{eq:transverse}, we define the \emph{spatial coordinates} as an alternative representation for the location of the dynamical system~\eqref{eq:dynamical_system} in the Euclidean space:
\begin{equation}\label{eq:spatial_coord}
    \bm{p}^\Gamma(t) = \left[\xi(t),\bm{\eta}(t)\right]\in\mathbb{R}^3
\end{equation}
For an illustrative visualization of the spatial coordinates associated to the reference path $\Gamma$, please refer to Fig.~\ref{fig:spatial_coordinates}.

\subsection{Derivation of equations of motion}\label{subsec:deriv_ode}
Given the dynamical system in~\eqref{eq:dynamical_system} and the parametric reference path $\Gamma$ in~\eqref{eq:geom_ref}, we derive the equations of motion for the spatial coordinates in~\eqref{eq:spatial_coord}. To conduct this derivation, we offer two distinct approaches: one based on kinematics and another rooted in optimality principles.

\subsubsection*{A kinematic derivation}
From the distances in~\eqref{eq:dist} the Cartesian coordinates of the dynamical system~\eqref{eq:dynamical_system} can be expressed as
\begin{equation}\label{eq:pos_wf}
    \bm{p}^W(t) = \bm{\gamma}(\txi) + \mathrm{R}(\txi)\bm{d}^\Gamma(t)
\end{equation}
Differentiating \eqref{eq:pos_wf} with respect to time results in
\begin{equation}\label{eq:vel_wf}
    \bm{v}^\text{W}(t) = \dot{\xi}(t)\left(\bm{\gamma}'(\txi)+\mathrm{R}'(\txi)\bm{d}^{\Gamma}(t)\right)+\mathrm{R}(\txi)\dot{\bm{d}}^{\Gamma}(t)\,.
\end{equation}
Denoting $\bi^\text{W} = \left[1,0,0\right]^\intercal$ as the first component of the world-frame and, recalling from~\eqref{eq:parametric_speed} that the curve's parametric speed is $\sigma(\txi) = ||\bm{\gamma}'(\txi)||$, we can derive:
\begin{gather}\label{eq:funcpos_der}
    \bm{\gamma}'(\txi)\equiv\sigma(\txi)\bm{e_1}(\txi)\,\equiv\,\mathrm{R}(\txi) \bi^\text{W}\sigma(\txi)\,.
\end{gather}
Introducing \eqref{eq:funcpos_der} in \eqref{eq:vel_wf} and multiplying it with $\mathrm{R}^\intercal(\txi)$ leads to
\begin{multline*}
    0 = \dot{\xi}(t)\left(
    \sigma(\txi)\bi^\text{W}+
    \mathrm{R}(\txi)^\intercal \, \mathrm{R}'(\txi)\, \bm{d}^{\Gamma}(\txi)
    \right)+\\ \dot{\bm{d}}^{\Gamma}(t) - \mathrm{R}^\intercal(\txi)\bm{v}^{\text{W}}(t)\,.
\end{multline*}
Leveraging the skew symmetric matrix in~\eqref{eq:R_ode}, the latter equation can be simplified to
\begin{multline*}
    0 = \dot{\xi}(t)\left(
    \sigma(\txi)\bi^\text{W}+
    \Omega^{\Gamma}(\txi)\bm{d}^{\Gamma}(\txi)
    \right)
    +\\
    \dot{\bm{d}}^{\Gamma}(t) - \mathrm{R}^\intercal(\txi)\bm{v}^{\text{W}}(t)\; ,
\end{multline*}
which combined with~\eqref{eq:R_ode} and~\eqref{eq:dist}, finally yields the equations of motion for the  \emph{spatial coordinates}:

\begin{framedsidebarYellow}{Path-parameterized Cartesian motion}
    

\begin{minipage}{15.5cm}
\sdbaryellowinitial{T}he motion of a three-dimensional point moving at speed $\bm{v}^\text{W}(t)$ with respect to a reference path parameterized by $\txi$ with parametric speed $\sigma(\txi)$ and an adapted path-frame $\mathrm{R}(\xi)=\left[\bm{e_1}(\txi),\bm{e_2}(\txi),\bm{e_3}(\txi)\right]$ whose angular velocity is $\bm{\omega}(\txi) = \omega^\Gamma_1(\txi)\bm{e_1}(\txi) + \omega^\Gamma_2(\txi)\bm{e_2}(\txi) + \omega^\Gamma_3(\txi)\bm{e_3}(\txi)$ is given by the following equations:
\begin{subequations}\label{eq:spatial_coord_eq_motion}
\begin{gather}
    \dot{\xi}(t) = \frac{\bm{e_1}(\txi)^\intercal \bm{v}^\text{W}(t)}{\sigma(\txi) - \omega^\Gamma_3(\txi) \eta_1(t) + \omega^\Gamma_2(\txi) \eta_2(t)}\,,\label{eq:xi_dot}\\
    \dot{\eta}_1(t) = \bm{e_2}(\txi)^\intercal \bm{v}^\text{W}(t) + \dot{\xi}(t) \omega^\Gamma_1(\txi) \eta_2(t)\,,\label{eq:w1_dot}\\
    \dot{\eta}_2(t) = \bm{e_3}(\txi)^\intercal \bm{v}^\text{W}(t) - \dot{\xi}(t) \omega^\Gamma_1(\txi) \eta_1(t)\,.\label{eq:w2_dot}
\end{gather}
\end{subequations}
\end{minipage}

\end{framedsidebarYellow}

\subsubsection*{An optimality derivation}
In here we show that an alternative approach also allows for the derivation of the equations of motion in~\eqref{eq:spatial_coord_eq_motion}. More specifically, we focus on the original definition of the progress variable $\xi(t)$ in~\eqref{eq:xi}. Given that this is an unconstrained optimization, we leverage the first order optimality condition, so that
\begin{equation}\label{eq:first_order_opt}
    0 = \dv{}{\theta}\left(\frac{1}{2}||\bm{p}^W(t) - \bm{\gamma}(\theta)||^2\right)\,,
\end{equation}
which for the optimal path parameter $\theta^*(t)=\txi$ results in $0 = \left(\bm{p}^W(t) - \bm{\gamma}(\txi)\right)\bm{\gamma}^{'}(\txi)$. Similarly, we enforce the first optimality condition with respect to time, i.e., $0 = \dv{}{t}\left(\left(\bm{p}^W(t) - \bm{\gamma}(\txi)\right)\bm{\gamma}^{'}(\txi)\right)$, whose expansion is
\begin{multline*}
    0 = \bm{v}^\text{W}(t)\bm{\gamma}^{'}(\txi)-\dot{\xi}(t)\bm{\gamma}^{'}(\txi)\bm{\gamma}^{'}(\txi)+\\ \dot{\xi}(t)\left(\bm{p}^W(t)-\bm{\gamma}(\txi)\right)\bm{\gamma}^{''}(\txi)\,.
\end{multline*}
Noticing that $\sigma^2(\txi)=\bm{\gamma}^{'}(\txi)\bm{\gamma}^{'}(\txi)$ the equation above simplifies to
\begin{equation}\label{eq:xidot_opt}
    \dot{\xi}(t) = \frac{\bm{v}^\text{W}(t)\bm{\gamma}^{'}(\txi)}{\sigma^2(\txi) -\bm{d}^{\text{W}}(t)\bm{\gamma}^{''}(\txi)}\,.
\end{equation}
Recalling that $\bm{e_1}(\txi)=\frac{\bm{\gamma}(\txi)}{\sigma(\txi)}$, follows that $\bm{\gamma}^{''}(\txi)=\bm{e_1}^{'}(\txi)\sigma(\txi) + \bm{e_1}(\txi)\sigma^{'}(\txi)$, whose derivative in the first term is known from~\eqref{eq:R_ode}, i.e., $\bm{e_1}^{'}(\txi) = \bm{e_2}(\txi)\omega^\Gamma_3(\txi) - \bm{e_3}(\txi)\omega^\Gamma_2(\txi)$ and second term gets cancelled because $\bm{d}^{\text{W}}(t)$ is perpendicular to $\bm{e_1}(\txi)$. Incorporating this information into~\eqref{eq:xidot_opt} results in
\begin{align}
    \dot{\xi}(t) = \frac{\bm{v}^\text{W}(t)\bm{\gamma}^{'}(\txi)}{\sigma^2(\txi) -\sigma(\txi)\bm{d}^{\text{W}}(t)\beta}\,,\quad\text{where}\notag\\
    \beta = (\bm{e_2}(\txi)\omega^\Gamma_3(\txi) - \bm{e_3}(\txi)\omega^\Gamma_2(\txi))\,.
\end{align}
Dividing both the numerator and denominator by $\sigma(\txi)$ and combining it with~\eqref{eq:transverse}, coincides with ~\eqref{eq:xi_dot}, the equation of motion for the progress variable $\txi$ derived by the previous kinematic approach. The remaining equations of motions for the transverse coordinates $\bm{\eta}(t)$ ~\eqref{eq:w1_dot} and ~\eqref{eq:w2_dot}, can easily be obtained from derivating~\eqref{eq:transverse} on time and following similar simplifications as above.
\subsection{A universal path-parameterization}
The equations of motion for the spatial states derived in~\eqref{eq:spatial_coord_eq_motion} are universal, as they do not rely on any assumptions regarding the underlying path parameterization. In other words, they are applicable to any parametric path, regardless of its parametric speed and moving frame. To showcase this universality, we demonstrate how the well-known Frenet-Serret based parametric model is a particular instance of the equations of motion in~\eqref{eq:spatial_coord_eq_motion}. Additionally, we show how the two-dimensional case, relevant for planar application such as autonomous driving, is a trivial simplification of the three-dimensional one.

\subsection{A particular case: The Frenet Serret based models}
\noindent The analytical simplicity and ease of implementation make the FSF the most widely applied moving frame in literature~\cite{verschueren2016time, van2016path, kumar2017path}. This spread is rooted in the autonomous driving community~\cite{gao2012spatial,reiter2021mixed,reiter2023frenet}, where the planar application of the FSF allows for numerical tricks to dodge its fundamental limitations. This influence, combined with the aforementioned simplicity, have made the FSF the de facto standard for path-parametric planning and control, even for three-dimensional applications~\cite{spedicato2017minimum,arrizabalaga2022towards,krinner2024time}, where the FSF suffers from all the limitations discussed in Fig.~\ref{fig:sing_twist}. As a consequence, the parametric formulations available in the literature are specific to the FSF. To show this, it is sufficient to tailor eqs.~\eqref{eq:spatial_coord_eq_motion} by specifying the angular velocity as in~\eqref{eq:angvel_fsf} i.e., $\left[\omega^\Gamma_1,\omega^\Gamma_2,\omega^\Gamma_3\right] = \sigma\left[\tau,0,\kappa\right]$. Furthermore, if the curve is assumed to be parameterized directly by its arc-length $L=\xi$, the parametric speed reduces to a unit magnitude $\sigma(\xi) = \dv{L}{\xi}=1$: 
\begin{subequations}\label{eq:point_mass_eq_fs}
\begin{gather}
    \dot{\xi}(t) = \frac{\bm{e_1}(\txi)^\intercal \bm{v}^\text{W}(t)}{1 - \kappa(\txi) \eta_1(t)}\,,\\
    \dot{\eta}_1(t) = \bm{e_2}(\txi)^\intercal \bm{v}^\text{W}(t) + \dot{\xi}(t) \tau(\txi) \eta_2(t)\,,\\
    \dot{\eta}_2(t) = \bm{e_3}(\txi)^\intercal \bm{v}^\text{W}(t) - \dot{\xi}(t) \tau(\txi) \eta_1(t)\, .
\end{gather}
The resultant equations of motion are specific to the FSF and match the ones available in literature, e.g.~\cite{verschueren2016time, van2016path}, showcasing the generality of our equations in~\eqref{eq:spatial_coord_eq_motion}.
\end{subequations}
\subsection{The planar 2D case}
In the case of planar motions, such as ground vehicles, the parameterization in~\eqref{eq:spatial_coord_eq_motion} smplifies to two states, i.e., the second component of the transverse components gets cancelled $\eta_2(t)=0$. Consequently, the equations of motion in~\eqref{eq:spatial_coord_eq_motion} reduce to the following planar model:
\begin{subequations}
\begin{gather}
    \dot{\xi}(t) = \frac{\bm{e_1}(\txi)^\intercal \bm{v}^\text{W}(t)}{\sigma(\txi) -\omega^\Gamma_3(\txi) \eta_1(t)}\,,\\
    \dot{\eta}_1(t) = \bm{e_2}(\txi)^\intercal \bm{v}^\text{W}(t)\, .
\end{gather}
\end{subequations}
In a similar way as for the three-dimensional case, the planar model can be tailored to the FSF by specifying the angular velocity as in~\eqref{eq:angvel_fsf} and the parametric speed as $\sigma(\xi) = 1$. This results in the following planar model:
\begin{subequations}\label{eq:fs_planar}
\begin{gather}
    \dot{\xi}(t) = \frac{\bm{e_1}(\txi)^\intercal \bm{v}^\text{W}(t)}{1 - \kappa(\txi) \eta_1(t)}\,,\\
    \dot{\eta}_1(t) = \bm{e_2}(\txi)^\intercal \bm{v}^\text{W}(t)\,,
\end{gather}
\end{subequations}
which is commonly employed in the autonomous driving community.

\subsection{Modularity: Frames and Equations}
Before concluding this section, there are two points that we would like to emphasize: First, the equations derived in~\eqref{eq:spatial_coord_eq_motion} are universal, in the sense that they can be used alongside any moving frame and path-parameterization technique. For example, in the previous subsection we have tailored them for the FSF case. To shed some light on the choice of the moving frame, we suggested the PTF as the most suitable candidate. It goes without saying that Algorithms~\ref{algo:ptf_integration} and~\ref{algo:ptf_derivatives} interplay perfectly with the equations of motion in~\eqref{eq:spatial_coord_eq_motion}. The fusion of the PTF and the universal equations of motion is a powerful formulation, that enjoys the benefits of both ingredients.

The second point is to recognize that, despite our best efforts, we -- the authors -- might have failed to identify the most appropriate frame and parameterization technique. It is very likely that future researchers will come up with more appropriate methods to define and compute moving frames. Despite this, it is important to insist that the equations of motion in~\eqref{eq:spatial_coord_eq_motion} still remain relevant. The presented universal framework is modular in the sense that the underlying ingredients are completely decoupled, i.e., the path parameterization technique is agnostic to the equations of motion of the spatial states. Therefore, even if future research leads to the development of more suitable moving frames, they can still be used alongside the equations of motion in~\eqref{eq:spatial_coord_eq_motion}.

\vspace{1cm}
\section{Why path-parametric?}
To highlight the appeal, practicality and universality of the presented path-parametric framework in designing planning and control algorithms, our analysis is structured into two parts. First, we delve into low-level control, exploring the foundational reasons that led to the development of path-parametric methods. Second, we demonstrate how these core ideas extend to broader motion planning scenarios, where the desired trajectories are intended to fully exploit the available free space. 
\subsection{An illustrative case-study: A two-link robotic manipulator}
To perform these experiments, we utilize a two-link \mybox{myLightBeige}{\emph{robotic manipulator}}, which serves as a baseline system. This platform allows us to analyze the motions of a nonlinear system subject to constraints in both configuration and task spaces. The state vector, defined as $\bm{x} = \left[p_x, p_y, \theta_1, \theta_2, \dot{\theta}_1, \dot{\theta}_2\right]$, captures the end-effector position, joint angles, and their respective velocities. The control inputs, represented as $\bm{u} = \left[\ddot{\theta}_1, \ddot{\theta}_2\right]$, correspond to the joint accelerations. The equations of motion for the end-effector's position are given by
\begin{subequations}\label{eq:robot_mani_ode}
\begin{align}
\dot{p}_x &= -L_1\dot{\theta}_1\sin(\theta_1) - L_2\dot{\theta}_2 \sin(\theta_2), \\
\dot{p}_y &= L_1\dot{\theta}_1\cos(\theta_1) + L_2\dot{\theta}_2 \cos(\theta_2),
\end{align}
where $L1=1$ and $L2=1$ are the lengths of the links. Additionally, the geometric reference used along the upcoming examples is a planar sinusoidal curve parameterized by $\xi$ and given by the following equation:
\end{subequations}
\begin{equation}\label{eq:robot_mani_ref}
    \bm{\gamma}(\xi) = 1 + 0.5 \sin(2\,\pi\,\xi)
\end{equation}


\begin{figure}[t]
    \centering
    \begin{tcolorbox}[colback=myLightBeige, colframe=white, boxrule=0pt, enlarge left by=0mm, enlarge right by=0mm, arc=6pt,  boxsep=-3pt]
    \includegraphics[width=\linewidth]{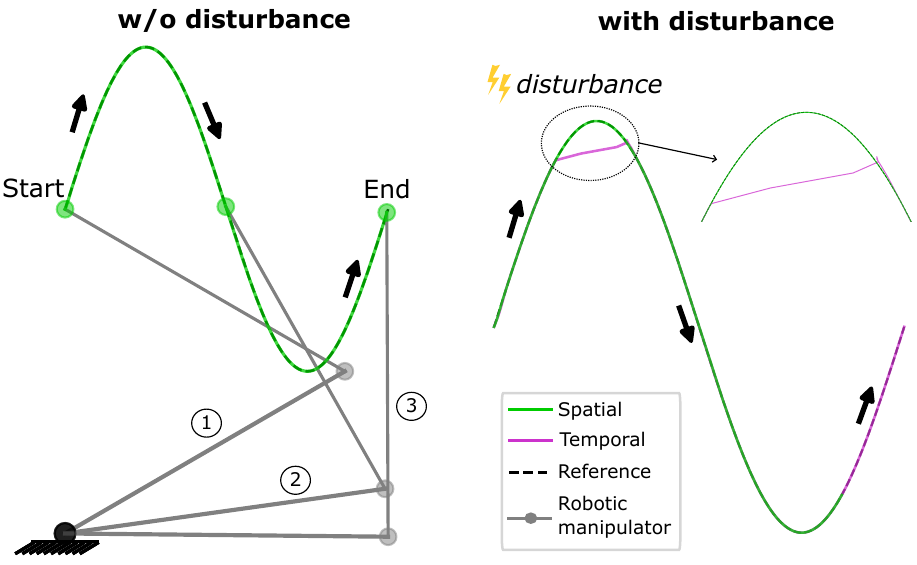}
    \end{tcolorbox}
    \caption{Comparison between a \tmp and \spa reference-tracking for a two joint robotic manipulator when traversing a sinusoidal path. The right side depicts three successive sequences for motions of the robotic manipulator in a nominal scenario, without disturbances. In the right, we show the trajectories obtained with both methods in the presence of a disturbance. In the left, the temporal reference keeps on progressing while the disturbance is happening, forcing it to \textit{catch-up}, resulting in a large deviation. In comparison, the spatial reference only depends on its location, and thus, is able to resume without generating a large error.}\label{fig:whypp_temporal_vs_spatial}
\end{figure}

\subsection{Temporal vs Spatial reference-tracking}
We begin by analyzing a simple yet illustrative example that demonstrates the benefits of path-parametric methods: a comparison between temporal and spatial references. The task involves guiding the robotic manipulator's end-effector along the geometric reference defined in~\eqref{eq:robot_mani_ref} using the dynamics described in~\eqref{eq:robot_mani_ode}. A \SI{2}{\second} disturbance at the first maximum point temporarily obstructs the end-effector's progress along the path. To ensure a fair comparison, both the \tmp and \spa formulations are calibrated to achieve the same navigation time in the absence of disturbances.

In particular, the controllers are formulated in a Nonlinear Model Predictive Control (NMPC) fashion, where an Nonlinear Program (NLP) is solved at a receding horizon fashion. Specifically, we solve the NLP in a multiple shooting fashion with $N$ nodes as:
\begin{subequations}\label{eq:tmp_vs_spa}
\begin{align}
\min_{\substack{\bm{x}_{0,...,N}\\\bm{u}_{0,...,N-1}}}&||\bm{x}_N - \bm{x}_{\text{ref}}(\cdot)||^{2}_{Q_E} + \sum_{k=0}^{N} ||\bm{x}_k - \bm{x}_{\text{ref}}(\cdot)||^{2}_{Q} +||\bm{u}_k||^{2}_{R}  \label{eq:tmp_vs_spa_cost}\\
\text{s.t.}\quad&\bm{x}_0 = \bm{x_i},\\
&\bm{x}_{k+1} = f(\bm{x}_k,\bm{u}_k, dt), \quad  k = \left[0,...\,,N-1\right]\label{eq:tmp_vs_spa_dynamics}\\
&c(\bm{x}_k,\bm{u}_k)\geq  0, \quad k = \left[0,...\,,N-1\right]\label{eq:tmp_vs_spa_ineqs}
\end{align}
\end{subequations}

\noindent where the reference in~\eqref{eq:tmp_vs_spa_cost} is time-dependent -- $ x_\text{ref}(t)$ -- for the tracking case, while it is path-parameter dependent -- $x_\text{ref}(\xi)$ -- in the spatial case. The continuity constraints in~\eqref{eq:tmp_vs_spa_dynamics} account for the equations of motion of the robotic manipulator given in~\eqref{eq:robot_mani_ode} with a horizon $T$ and a time step $dt=T/N$. The state and input constraints are expressed with~\eqref{eq:tmp_vs_spa_ineqs}. 

We solve the NLP \eqref{eq:tmp_vs_spa} using the optimal control framework ACADOS \cite{verschueren2018towards}, which employs a sequential quadratic programming (SQP) method. The underlying quadratic programs leverage their multi-stage structure and are solved using HPIPM \cite{frison2020hpipm}. For dynamics integration, an explicit 4th-order Runge-Kutta method is applied with $T=1$ and $N=20$. To navigate the reference path, the weights are designed to prioritize position tracking, with $Q=Q_E=\text{diag}\left(1,1,0,0,0,0\right)$ and $R=\text{diag}\left(1\mathrm{e}-4,1\mathrm{e}-4\right)$.

The resulting motions for the temporal and spatial references are illustrated in Fig.~\ref{fig:whypp_temporal_vs_spatial}. The comparison clearly shows that the motion associated with the temporal reference deviates from the desired path. This occurs because, while the end-effector is blocked, the temporal reference continues to advance. Consequently, once the \SI{2}{\second} disturbance ends, the end-effector must "\emph{catch up}", leading to the observed deviation. In contrast, the spatial reference remains stationary during the disturbance, allowing the motion to resume seamlessly without any deviations. This realization underpins the rise of path-following methods, a paradigm shift that overcomes the limitations of the temporal dependency. The following subsection delves deeper into this concept.

\subsection{Path-following: A superior alternative to path-tracking}
We have observed that tracking a spatial reference offers superior robustness against disturbances. This principle forms the foundation of path-following formulations, which shift the focus from a time-varying reference dictating \emph{where to be when} to minimizing deviations from the reference path. In path-following, the velocity along the path becomes a secondary concern, adjustable to improve performance. In other words, the problem is no longer about adhering to a predefined time schedule but instead treats velocity as an additional degree of freedom for traversing the reference. The versatility of path-following across a wide range of applications, combined with its independence from the inherent limitations of traditional reference tracking~\cite{aguiar2005path}, explains the substantial attention it has garnered in the literature. Comprehensive overviews of existing approaches can be found in~\cite{faulwasser2015nonlinear,hung2023review}.

\begin{figure}[t]
\centering
\begin{tcolorbox}[colback=myLightBeige, colframe=white, boxrule=0pt, enlarge left by=0mm, enlarge right by=0mm, arc=6pt,  boxsep=-3pt]
\includegraphics[width=\columnwidth]{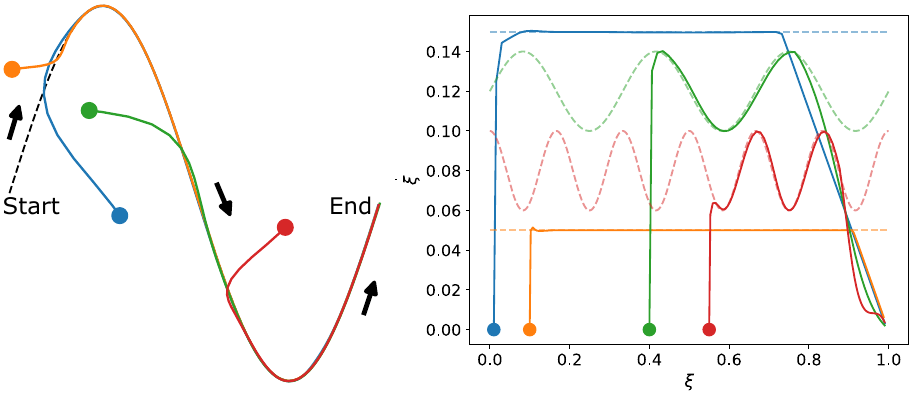}
\end{tcolorbox}
\caption{End-effector motions of a two-link robotic manipulator computed with an optimization-based path-following approach, trading off between orthogonal convergence to the path and achieving a predefined velocity profile. Trajectories are color-coded by initial conditions. The plot on the right shows traversal velocities, with desired reference velocities as dashed lines.}\label{fig:whypp_xidot}
\vspace{-5mm}
\end{figure}
\begin{figure*}[t]
    \vspace{-3mm}
    \centering
    \begin{tcolorbox}[colback=myLightBeige, colframe=white, boxrule=0pt, enlarge left by=0mm, enlarge right by=0mm, arc=6pt]
    \includegraphics[width=\linewidth]{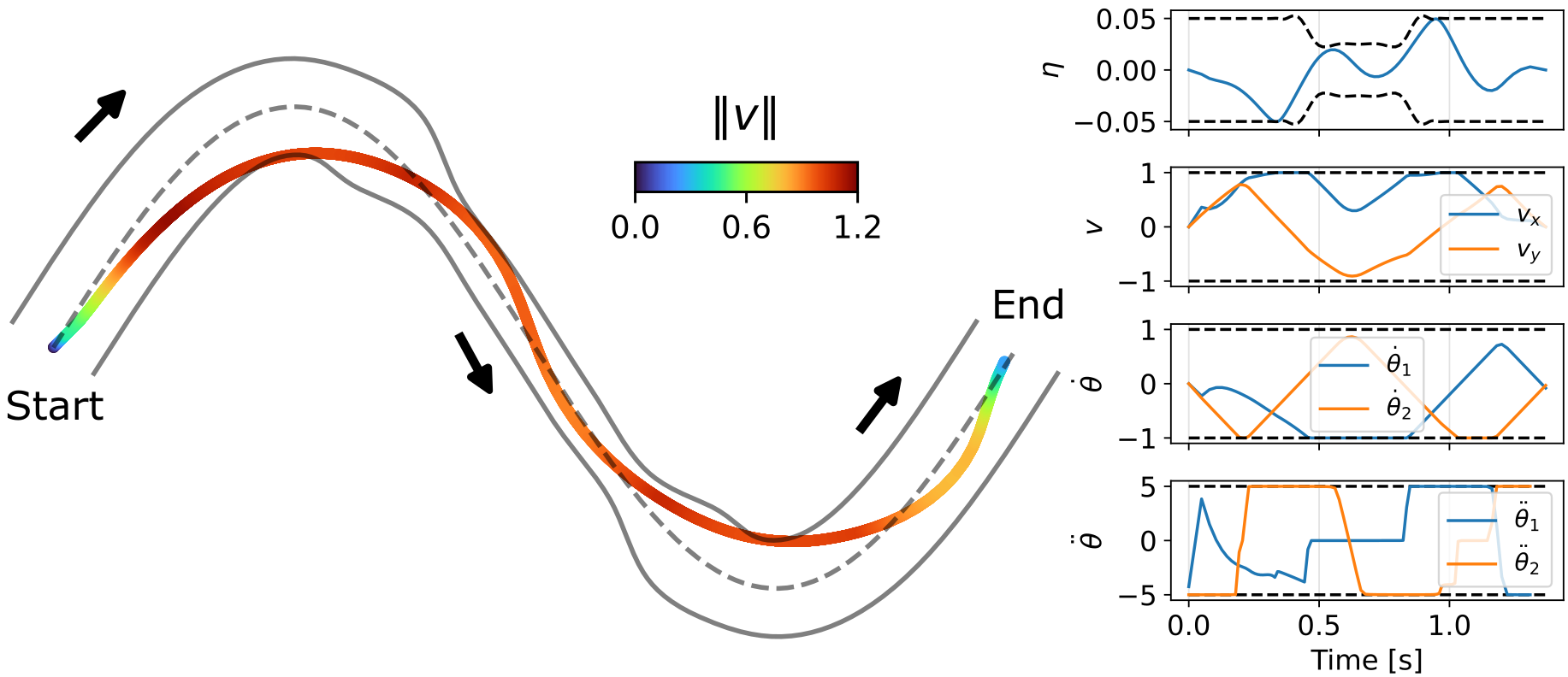}
    \end{tcolorbox}
    \caption{End-effector motions of a two-link robotic manipulator traversing a reference path within a corridor, representing the admissible deviation region, color-coded by velocity norm. On the right, from top to bottom: orthogonal spatial coordinate, end-effector velocity components, and joint angle velocities and accelerations, with bounds shown as black dashed lines.}\label{fig:whypp_corridor}
    \vspace{-4mm}
\end{figure*}
\noindent To showcase the relevance of path-following, we build upon the previous reference-tracking example by formulating a path-following method. Instead of tracking a position reference, this approach tracks a desired velocity profile $\dot{\xi}_\text{ref}$ that determines how the end-effector traverses the reference path. To achieve this, we project the Cartesian coordinates of the end effector $\left[p_x, p_y\right]$ to the spatial coordinates $\left[\xi, \eta\right]$, namely the progress along the path and the orthogonal distance to it. As a consequence the new states of the robotic manipulator are $\bm{x} = \left[\xi, \eta, \dot{\xi}, \dot{\eta}, \theta_1, \theta_2, \dot{\theta}_1, \dot{\theta}_2\right]$. The corresponding equations of motion for the spatially projected robotic manipulator are derived by taking the first derivative and combining eqs.~\eqref{eq:robot_mani_ode} with the path-parameterized Cartesian motions in~\eqref{eq:spatial_coord_eq_motion}.

For this case-study, we reuse the same NMPC formulation as before. However, the weights associated to the cost function of the NLP~\eqref{eq:tmp_vs_spa_cost} are chosen to trade-off between converging to the reference path $\lim_{t\to\infty}||\eta(t)||_2=0$ and tracking a desired velocity profile $\lim_{t\to\infty}||\dot{\xi}(t) - \dot{\xi}_{\text{ref}}(t)||_2=0$, such that $Q=Q_E=\text{diag}\left(0,1\mathrm{e}-1,1,1\mathrm{e}-4,0,0,0,0\right)$ and $R=\text{diag}\left(1\mathrm{e}-4,1\mathrm{e}-4\right)$.

To test the performance of the optimization-based path-following formulation, we initialize the end-effector at four distinct positions, each associated with a different velocity profile, characterized by different magnitudes, shapes and frequencies. The results depicted in Fig.~\ref{fig:whypp_xidot} verify that, regardless of the initial state and the desired velocity profile, the position of the robotic manipulator converges to the geometric reference, both in the task space and the traversing speed.

This example raises the question of what the optimal velocity profile $\dot{\xi}_\text{ref}$ is and how it can be computed. To answer these questions, we turn to the upcoming subsection, where we showcase the capabilities enabled by path-parametric methods in a more generic context of motion planning, ultimately expanding the range of methods that benefit from the presented framework.


\subsection{Motion planning: Spatial-awareness}
In the previous subsections, we demonstrated how the path-parametric framework enables the design of controllers capable of achieving complex navigation behaviors, such as explicitly controlling the convergence rate to a reference path or maintaining a predetermined traversal velocity profile. Here, we aim to extend these examples to a more general motion planning scenario, providing a comprehensive view of the full potential of path-parametric methods.

To this end, we extend the previous two-link robotic manipulator case study by assuming that the sinusoidal reference path~\eqref{eq:robot_mani_ref} is enclosed within a corridor $\mathcal{C}$, allowing the end-effector to navigate within this region. This scenario is commonly encountered in industrial robotics, where exact tracking of the reference is not required, and approaching the surrounding area within a desired tolerance is sufficient. Specifically, our goal is to find the time-optimal motion that takes advantage of this admissible region to move the robotic manipulator's end-effector from the start to the end of the reference path, while respecting both dynamical and spatial constraints.

We approach this as an offline motion planning problem, where — similar to~\cite{verschueren2016time,spedicato2017minimum, arrizabalaga2023sctomp} — we utilize the transformation from the temporal to the spatial domain, $\bm{x}(t)\rightarrow\bm{x}(\xi)$, to convert the time minimization problem into a finite horizon problem. In particular, the Optimal Control Problem (OCP) we solve is:

\begin{subequations}\label{eq:minT_par}
    \begin{flalign}
     &\qquad\qquad\min_{\bm{x}(\cdot),\bm{u}(\cdot)} T = \int_{\xi_0}^{\xi_f}\frac{1}{\dot{\xi}(\bm{x}(\xi))}\, d\xi&
    \end{flalign}
    \vspace{-5mm}
	\begin{alignat}{3}
	\text{s.t.}\quad& \bm{x}(\xi_0) = \bm{x_0}\,,\quad \bm{x}(\xi_f) = \bm{x_f}\,,\\
	&\bm{x}' = \frac{f(\bm{x}(\xi),\bm{u}(\xi))}{\dot{\xi}(\bm{x}(\xi))}, &\quad&\xi \in \left[\xi_0,\xi_f\right]\label{eq:dynamic_const}\\
	&\bm{x}(\xi)\in\mathcal{X}\,,\,\bm{u}(\xi)\in\mathcal{U}\,,    &\quad&\xi \in \left[\xi_0,\xi_f\right]\label{eq:other_constr}\\
	&\bm{x}(\xi)\in \mathcal{C} \label{eq:spatial_constr}\,.
	\end{alignat}
Due to the transformation from the temporal to the spatial domain $\bm{x}(t)\rightarrow\bm{x}(\xi)$, the system dynamics in~\eqref{eq:dynamic_const} evolve according to path-parameter $\xi$, and thus, the resulting OCP~\eqref{eq:minT_par} is a finite horizon problem, as opposed to the original time minimization problem. Besides that, we represent the end-effector by the same projected states as before, i.e., $\bm{x} = \left[\xi, \eta, \dot{\xi}, \dot{\eta}, \theta_1, \theta_2, \dot{\theta}_1, \dot{\theta}_2\right]$ and $\bm{u}=\left[\ddot{\theta}_1, \ddot{\theta}_2\right]$. This enables to formulate the corridor bounds in~\eqref{eq:spatial_constr} as convex constraints in the orthogonal coordinate. For the specific planar case at hand, this simplifies to
\begin{equation}\label{eq:corridor_constraint}
    \underline{\eta}(\xi)\leq\eta(\xi)\leq\overline{\eta}(\xi) \,,
\end{equation}
where $\underline{\eta}\,:\,\mathbb{R}\mapsto\mathbb{R}$ and $\overline{\eta}\,:\,\mathbb{R}\mapsto\mathbb{R}$ are $C^2$ parametric functions on $\xi$ that describe the upper and lower bounds of the corridor. Readers interested in how such a parametric corridor example extends to 3D and unknown environments should refer to the following rubric, where we provide an in-depth explanation on how such corridors can be computed. 

Beyond the spatial limitations inherent to the corridor geometry, we introduce further constraints on the configuration and task spaces, as defined in equation \eqref{eq:other_constr}, to ensure a more accurate representation of a realistic scenario. Specifically, these constraints are
\begin{flalign}
    &-1\leq\bm{\dot{\theta}}(\xi)\leq1\,,\quad-5\leq\bm{\ddot{\theta}}(\xi)\leq 5\,,\\
    &-1\leq\bm{v}(\xi)\leq1\,.
\end{flalign}
\end{subequations}
\noindent Following a similar approach to the one described in~\eqref{eq:tmp_vs_spa}, we transform the optimal control problem (OCP) defined in \eqref{eq:minT_par} into a nonlinear program (NLP) by discretizing it with the multiple shooting approach into N nodes. We then solve the resulting NLP using the IPOPT solver \cite{wachter2006implementation}.

The computed end-effector motions are illustrated in the left panel of Fig.~\ref{fig:whypp_corridor}, where the color gradient corresponds to the magnitude of the velocity. The right panel displays the evolution of the associated states and inputs. Given the behavior incentivized by the time-optimal formulation, the trajectory fully exploits the actuation by seeking the optimal trade-off between speed and spatial bounds. This phenomenon is observable from two different perspectives: Firstly, both the end-effector's trajectory and the orthogonal spatial coordinate $\eta$ dynamically adapt to the narrowing section of the corridor, showcasing the end-effector's ability to stay within the confines while leveraging the available space. Secondly, the bottom plot reveals that at least one joint acceleration remains saturated during most of the navigation, akin to the bang-bang behavior associated with time-optimal motions.
The specific time-optimal cost function, robotic manipulator dynamics, and corridor formulations chosen for this example serve as a proof of concept, illustrating how path-parametric methods offer a compelling framework for achieving agile motion while ensuring safety guarantees.


\section{Minimizing Time or Maximizing Progress?}

After exploring the fundamental advantages that have propelled the rise of the path-parametric paradigm, we shift our focus to one of its most notable applications: time-optimal navigation. Time optimal control tackles a fundamental challenge in control theory: steering a dynamic system from an initial state to a desired final state in minimal time~\cite{athans2007optimal}. Since the 1950s, this problem has attracted extensive research attention due to its wide-ranging applications—from military operations like missile interception to robotic navigation, rapid manufacturing changeovers, and optimized logistics routing~\cite{bobrow1985time}. Despite decades of theoretical advancement, a comprehensive mathematical framework for solving minimum-time control problems remains elusive.

Unlike other approaches, precomputing time-optimal trajectories offline --as done in Fig.~\ref{fig:whypp_xidot}-- for subsequent online tracking proves inadequate. Real-world implementations inevitably encounter noise and model uncertainties, causing deviations from the reference trajectory. Once the system deviates, the remaining path loses its time-optimality property. Continuing to follow such a trajectory not only compromises time efficiency but also risks system failure, as time-optimal motions operate at the physical limits of the system where any overshoot can have severe consequences~\cite{foehn2021time}.

To overcome these limitations, researchers have developed low-level controllers that directly compute approximately time-optimal motions without relying on offline references. These controllers employ progress maximization—a technique that optimizes the system's advancement along a geometric reference path~\cite{liniger2015optimization,romero2022model,kloeser2020nmpc, arrizabalaga2022towards}. As shown in \mybox{framc}{\hyperref[sidebar:tm_pm]{\emph{Time Minimization vs. Progress Maximization}}} this approach closely approximates true time-optimal motions while offering significant advantages in problem formulation and practical implementation. The theoretical foundations established in previous sections provide crucial insights into this methodology.
\begin{figure}[t]
    \centering
    \def\figurename{\textcolor{strongerGreen}{\textbf{FIGURE}}} 
    \captionsetup{labelformat=simple, labelfont={color=strongerGreen, bf}} 
    \begin{tcolorbox}[colback=framegreen, colframe=strongerGreen, boxrule=1pt, enlarge left by=0mm, enlarge right by=0mm, arc=6pt, boxsep=-5pt]
        \includegraphics[width=1\linewidth]{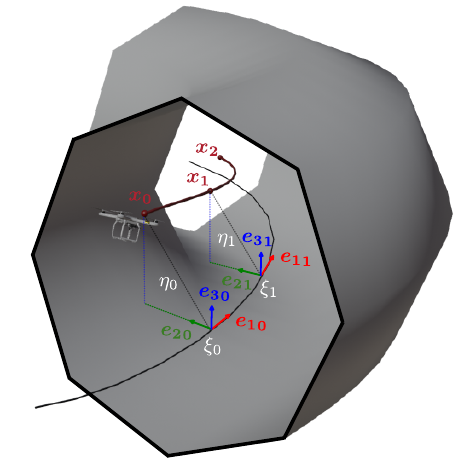}
    \end{tcolorbox}
    \caption{Path-parametric methods enable the formulation of planning and control algorithms within a moving frame relative to the reference path. The resulting spatial coordinates—path progress $\xi$ and orthogonal distance $\eta$—facilitate the development of \emph{spatially aware} algorithms that generate motions capable of better exploiting the geometric properties of the environment. In this context, safety is ensured by constraining the orthogonal coordinate $\eta$ within an admissible region, referred to as a \emph{corridor}. To demonstrate how these corridors can be efficiently generated, we focus on the method presented in~\cite{arrizabalaga2024differentiable} and highlight how these corridors seamlessly integrate with the suggested path-parametric framework.}
    \vspace{-5mm}
    \label{fig:tunnel_quad}
\end{figure}

\begin{sidebar}[sidebar:tm_pm]{Time-Minimization vs Progress-Maximization}
\sdbarinitial{P}rogress maximization has emerged as a leading approach for minimum-time motion control. Driven by advancements in numerical optimization and embedded solvers, progress maximization based prediction-based controllers have shown very promising results in real-world applications~\cite{liniger2015optimization,romero2022model,kloeser2020nmpc,arrizabalaga2022towards}. These controllers are built on path-parametric methods, allowing systems to operate near their performance limits and achieve behavior that closely approximates time-optimality. Additionally, these formulations leverage the capacity to easily impose collision-free constraints, which would otherwise be non-convex or difficult to enforce without the path-parametric structure~\cite{reiter2023frenet}. The combination of these attributes has made progress maximization the preferred approach for achieving agile performance along a designated reference path. However, while progress maximization serves as an approximation to time minimization, the precise quantification of the gap between these two methods remains unresolved. In this study, we aim to shed some light on this question by performing an experimental comparison of both approaches.


\subsubsection{An illustrative case-study: A miniature racing car}
As an exemplary system upon which we can test both control formulations, we choose a 1:43 miniature racecar. The state vector, defined as $\bm{x} = \left[X, Y, \varphi, v_x, v_y, r, d, \delta\right]$, represents the car’s position, orientation, linear and angular velocities, throttle position, and steering angle, respectively. The control inputs, throttle rate and steering rate, are denoted by $\bm{u} = \left[\dot{d}, \dot{\delta}\right]$. The equations of motion for these states are given by
\begin{subequations}\label{eq:car_dynamics}
\begin{align}
\dot{X} &= v_x \cos(\varphi) - v_y \sin(\varphi), \\
\dot{Y} &= v_x \sin(\varphi) + v_y \cos(\varphi), \\
\dot{\varphi} &= r, \\
\dot{v}_x &= \frac{1}{m} \left( F_{r,x} + F_{\text{fric}} - F_{f,y} \sin(\delta) + mv_y r \right), \\
\dot{v}_y &= \frac{1}{m} \left( F_{r,y} - F_{f,y} \cos(\delta) - m\,v_x r \right), \\
\dot{r} &= \frac{1}{I_z} \left( F_{f,y} l_f \cos(\delta) - F_{r,y} l_r \right),
\end{align}
\end{subequations}
whose tire forces are expressed as
\begin{subequations}\label{eq:tire_model}
\begin{align}
F_{f,y} &= D_f \sin(C_f \arctan(B_f \alpha_f))\,,\\
F_{r,y} &= D_r \sin(C_r \arctan(B_r \alpha_r)) \,, \\
F_{r,x} &= (C_{m1} - C_{m2}v_x)d\,, \\
F_{\text{fric}} &= -C_r - C_d v_x^2\,, 
\end{align}
where  $\alpha_f$ and $\alpha_r$ are the front and rear slip angles:
\begin{gather}
     \alpha_f = -\arctan \left(\frac{r\,l_f + v_y}{v_x}\right) + \delta, \\
     \alpha_r = \arctan \left(\frac{r\,l_r - v_y}{v_x}\right).
\end{gather}
\end{subequations}
All parameters involved in this model are summarized in the following table:
\begin{table}[H] 
	\centering
	\begin{tabular}{|c||c|c|c|}
	    \hline
	    \textit{Parameter}& \textit{Value} & \textit{Description} \\
		\hline
		$m$ & Mass  & \SI{0.041}{\kilo\gram}  \\
        \hline
        $I_z$ & Inertia  & \SI{27.8e-6}{\kilo\gram\metre\squared} \\
        \hline
		$\left[C_{m1}, C_{m2}\right]$ & Motor params.   &$\left[0.287,0.545\right]$\\
        \hline
		$\left[B_f, C_f, D_f\right]$ & Front tire params.   &$\left[2.579,1.269,0.192\right]$\\
        \hline
		$\left[B_r, C_r, D_r\right]$ & Rear tire params.   &$\left[3.385,1.269,0.174\right]$\\
        \hline
		$\left[l_r, l_f\right]$ & Dist. to front/rear axle   &$\left[0.029,0.033\right]$\SI{}{\centi\metre}\\
		\hline
	\end{tabular}
\end{table}

\subsubsection{Time-Minimization}
To compute time-optimal motions that are dynamically feasible, we formulate a predictive controller that solves an NLP with a cost function that solely minimizes time. We implement this approach using a multiple shooting method, where time is incorporated as a decision variable, as detailed below:
\begin{mdframed}[backgroundcolor=timeColor,innertopmargin=0pt,linecolor=timeColorStrong,linewidth=1pt,innerrightmargin=8pt,innerleftmargin=8pt,]
\begin{subequations}\label{eq:minTime_NLP}
\begin{align}
\min_{\substack{\bm{x}_{0,...,N}\\\bm{u}_{0,...,N-1}\\ dt_{0,...,N}}}&T = \sum_{k=0}^{N} dt_k \label{eq:minTime_cost}\\
\text{s.t.}\quad&\bm{x}_0 = \bm{x_i}, \, \bm{x}_N = \bm{x_f},\\
&\bm{x}_{k+1} = f(\bm{x}_k,\bm{u}_k, dt_k), \,  &k = \left[0,...\,,N-1\right]\label{eq:minTime_dynamics}\\
&c(\bm{x}_k,\bm{u}_k)\geq  0, \, &k = \left[0,...\,,N-1\right]\\
&dt_k\geq 0, \,  &k = \left[0,...\,,N-1\right]
\end{align}
\end{subequations}
\end{mdframed}
where $f$ in~\eqref{eq:minTime_dynamics} are the nonlinear dynamics in~\eqref{eq:car_dynamics} In contrast to standard MPC, the horizon shrinks with each step, meaning that $N$ decreases as the system progresses along the trajectory. To achieve the \emph{true} time-optimal solution, as defined in equation~\eqref{eq:minTime_cost}, we focus exclusively on minimizing time without incorporating any additional stabilizing or regularizing terms. Developing a solver capable of reliably addressing the nonlinear programming problem (NLP) defined in equation~\eqref{eq:minTime_NLP} at a receding horizon and at high rates necessitates careful attention to the underlying numerical methods and requires considerable expertise. Despite recent advancements in the field~\cite{kiessling2024almost,yangnew}, there is no reliable solver available to tackle NLP~\eqref{eq:minTime_NLP} in its most general form. For this experiment, we employ a Sequential L1 Quadratic Programming (SL1QP) approach~\cite{nocedal1999numerical}, akin to the methodology used in~\cite{kiessling2022feasible}.

\subsubsection{Progress-Maximization}
A widely used approach to approximate time-optimal trajectories is to maximize progress along the path. To achieve this, the progress needs to be represented as a system state. As illustrated in Fig.~\ref{fig:uppc_overview}, this can be done by either \emph{augmenting} the state vector -- $\bm{x}^\Gamma(t) = \left[x(t), \bar{\xi}(t)\right]$ -- or \emph{projecting} it -- $\bm{x}^\Gamma(t) = \left[\bar{\xi}(t), \eta(t)\right]$~\cite{liniger2015optimization,arrizabalaga2022towards}. Including progress as a state variable enables the design of an optimization-based predictive controller, where the cost function seeks to maximize progress along the path while satisfying dynamic and feasibility constraints. This setup encourages the system to traverse the path as quickly as possible within feasible limits, resulting in highly agile motions that closely approximate time-optimality.


\end{sidebar}

\begin{sidebar}{\continuesidebar}


Despite numerous attempts, the trajectories produced by these methods are limited in optimality due to several key issues:
\begin{enumerate} 
\item Conceptually, minimizing time and maximizing progress are not equivalent goals, and thus lead to fundamentally different trajectories. 
\item The absence of a Hessian in the cost function necessitates a stabilizing or regularizing term to ensure numerical stability. A common approach is to lightly penalize the derivative of the inputs, but this ultimately compromises the primary objective. 
\item When progress is introduced as a virtual variable to augment the state space, a synchronization term is needed to align the system dynamics with the virtual kinematic chain, thereby penalizing lag errors. This further detracts from achieving the true objective. 
\end{enumerate}
Since these limitations are inherent across this family of methods, we select contouring control as the representative approach for progress maximization~\cite{liniger2015optimization, romero2022model}. This choice is justified by its success as the first formulation to achieve near time-optimal motion control in racing contexts. Although it is based on \emph{state augmentation}, the insights from our forthcoming experiments are also applicable to other progress maximization methods, including those that employ \emph{projection} techniques~\cite{kloeser2020nmpc, arrizabalaga2022towards}. In particular, the NLP problem underlying the contouring controller is:
\begin{mdframed}[backgroundcolor=progressColor,innertopmargin=0pt,linecolor=progressColorStrong,linewidth=1pt,innerrightmargin=8pt,innerleftmargin=8pt,]
\begin{subequations}\label{eq:maxProgr_NLP}
\begin{align}
&\min_{\substack{\bm{x}_{0,...,N},\,\bm{u}_{0,...,N-1}\\ \xi_{0,...,N},\,\dot{\xi}_{0,...,N-1}}} \sum_{k=0}^{N-1} -\dot{\xi}_k + g(\bm{x}_k,\bm{u}_k,\xi_k, \dot{\xi}_k)~\label{eq:maxProgr_cost}\\
&\text{s.t.}\quad \bm{x}_0 = \bm{x_i}\,,\quad\xi_0 = \xi_i\,,\\
&\bm{x}_{k+1} = f(\bm{x}_k,\bm{u}_k, dt_k), \,\hspace{0.27cm}k = \left[0,...\,,N-1\right] \\
&\xi_{k+1} = \Xi(\xi_k,\dot{\xi}_k), \, \hspace{0.71cm}  k = \left[0,...\,,N-1\right] \\
&c(\bm{x}_k,\bm{u}_k)\geq  0, \, \hspace{1.15cm} k = \left[0,...\,,N-1\right]\\
&\xi_f \geq \xi_k \geq \xi_i, \, \hspace{1.27cm} k = \left[0,...\,,N-1\right] \\ 
&\dot{\xi}_k\geq0, \, \hspace{1.98cm} k = \left[0,...\,,N-1\right]\
\end{align}
\end{subequations}
\end{mdframed}
where $g$ in~\eqref{eq:maxProgr_cost} is a representative function for bringing numerical stability to the problem, either with some regularization or smoothening of the inputs and $\Xi$ is a kinematic integration that relates the progress variable with its velocity. 

To incite agile behavior, we formulate the cost function~\eqref{eq:maxProgr_cost} akin to the aforementioned contouring control formulation~\cite{liniger2015optimization} by maximizing the velocity of the progress variable, i.e., $\min -\dot{\xi}$. However, it is important to note that the approach may vary based on the particular formulation used. Alternatives include directly maximizing the progress variable $-\xi$~\cite{arrizabalaga2022spatial}, or regulating over a target velocity profile by minimizing $(\xi - \xi_\text{ref})^2$~\cite{kloeser2020nmpc, arrizabalaga2022towards}.

\subsubsection{Comparison without model mismatch}
We begin by comparing the \tm and \prm controllers in an ideal scenario, where the model used by the controller perfectly matches the real system. Fig.~\ref{fig:comp_nomismatch} presents the trajectories generated by both controllers. While both trajectories appear quite similar, the time-minimization formulation completes the task \SI{0.437}{s} faster.
\begin{figure}[H]
\vspace{-2mm}
\def\figurename{\textcolor{strongerBlue}{\textbf{FIGURE}}} 
\renewcommand{\thefigure}{\textcolor{strongerBlue}{\textbf{\arabic{figure}}}} 
\centering
\includegraphics[width=0.9\linewidth]{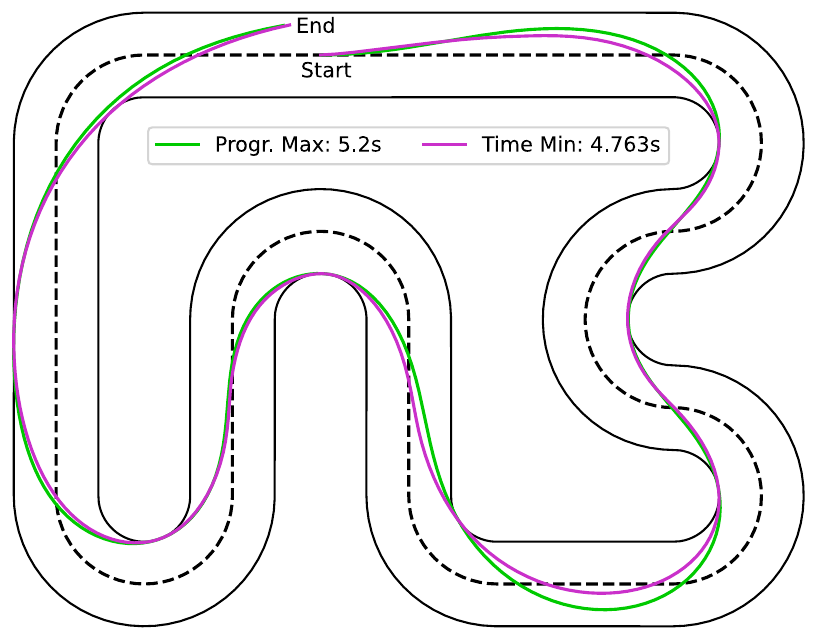}
\caption{Comparison between time-minimization and progress-maximization without model mismatch.}\label{fig:comp_nomismatch}
\end{figure}
\vspace{-4mm}
\noindent To better understand the source of this discrepancy, we refer to Fig.~\ref{fig:comp_nomismatch_tire}, which shows the rear tire forces. These forces represent the interaction between the car and the road, serving as a key indicator of how close the car is to its dynamic limits. Specifically, we examine the rear tire’s friction circle. The dashed line marks the tire's grip limit; if the applied force exceeds this limit, the car will lose traction and begin to slide. It is clear that the minimum time controller, unlike the progress maximization controller, pushes the tire to this limit, maximizing the available force without causing a slide. This aggressive utilization of grip explains why the minimum time controller achieves faster performance.
\begin{figure}[H]
\vspace{-5mm}
\bluefigure
\centering
 \includegraphics[width=0.65\linewidth]{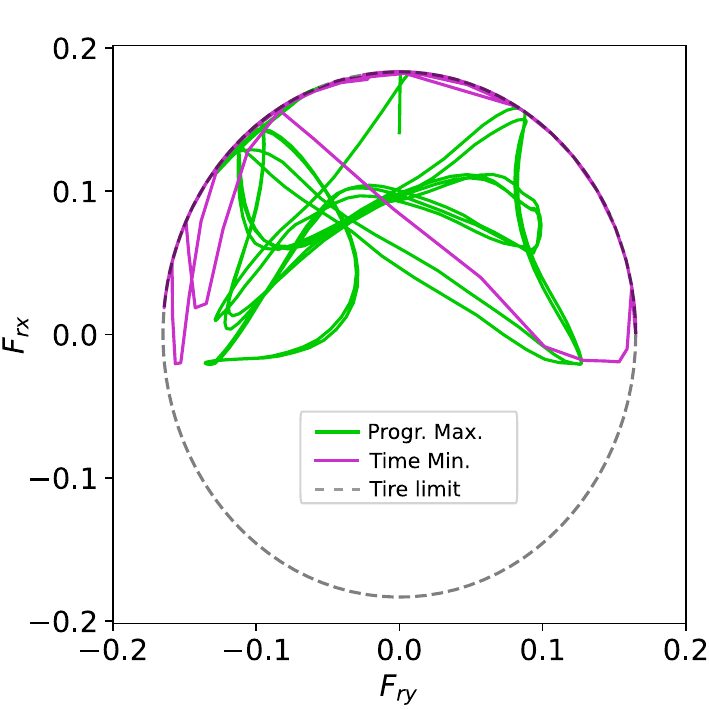}
 \caption{Friction circle for the rear tire without model mismatch. Tire limit is given by the dashed line.}\label{fig:comp_nomismatch_tire}
\end{figure}
\end{sidebar}

\begin{sidebar}{\continuesidebar}
\subsubsection{Comparison with deterministic model mismatch}
In the previous subsection, we observed that, without model mismatch, the minimum-time controller outperforms progress maximization. However, real-world systems are affected by disturbances and uncertainties. In this subsection, we explore whether this finding holds when model mismatch is introduced.

We begin by introducing a deterministic mismatch in the tire model by reducing the maximum lateral force the tire can generate. This reduction is quantified by the parameter $r$, representing the percentage decrease in the coefficient $D$ of the tire model~\eqref{eq:tire_model}, such that $\tilde{D} = (1-r)\,D$. This mismatch can be applied to either the front or the rear tire. The following figure illustrates the lateral tire forces for various values of $r$:
\begin{figure}[H]
\vspace{-2mm}
\bluefigure
\centering
 \includegraphics[width=0.95\linewidth]{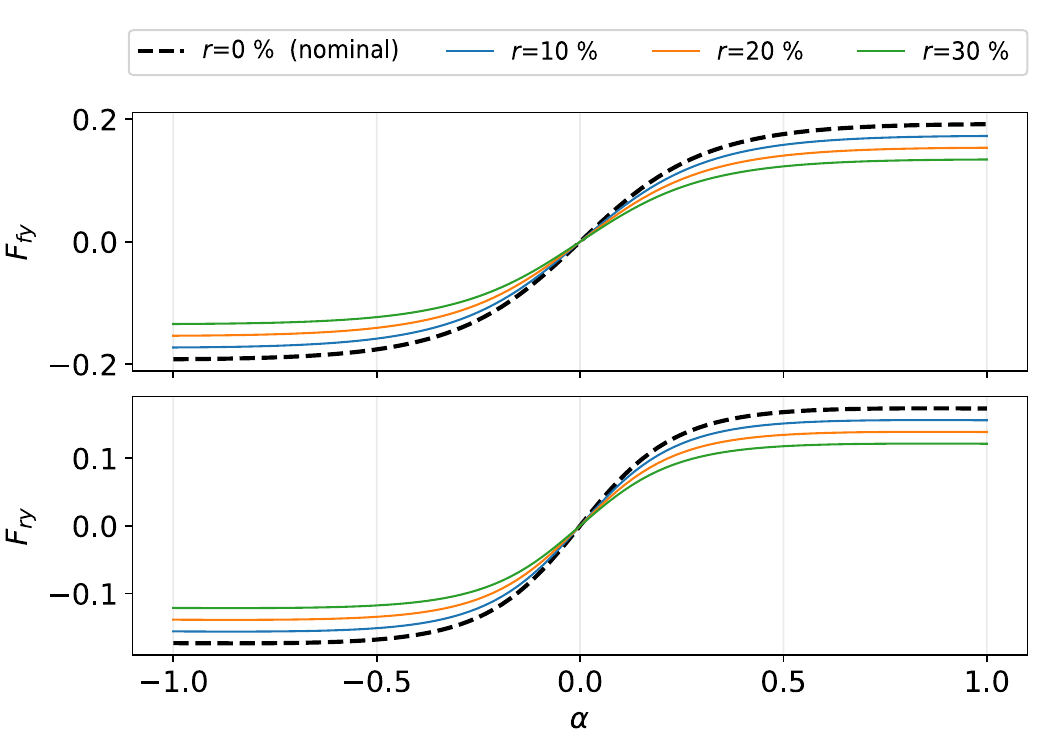}
 \caption{Tire model with reduction term $r$.}\label{fig:deterministic_mismat}
\end{figure}
\vspace{-2mm}
\noindent Given a slip angle $\alpha$, the force the tire can generate decreases as $r$ increases. Introducing this mismatch in the front tire induces understeer, while applying it to the rear tire causes oversteer. To illustrate these effects, Fig.~\ref{fig:deterministic_mismatch} shows the understeering and oversteering behavior resulting from the mismatch reduction of $20\%$:
\begin{figure}[H]
\bluefigure
\centering
 \includegraphics[width=\linewidth]{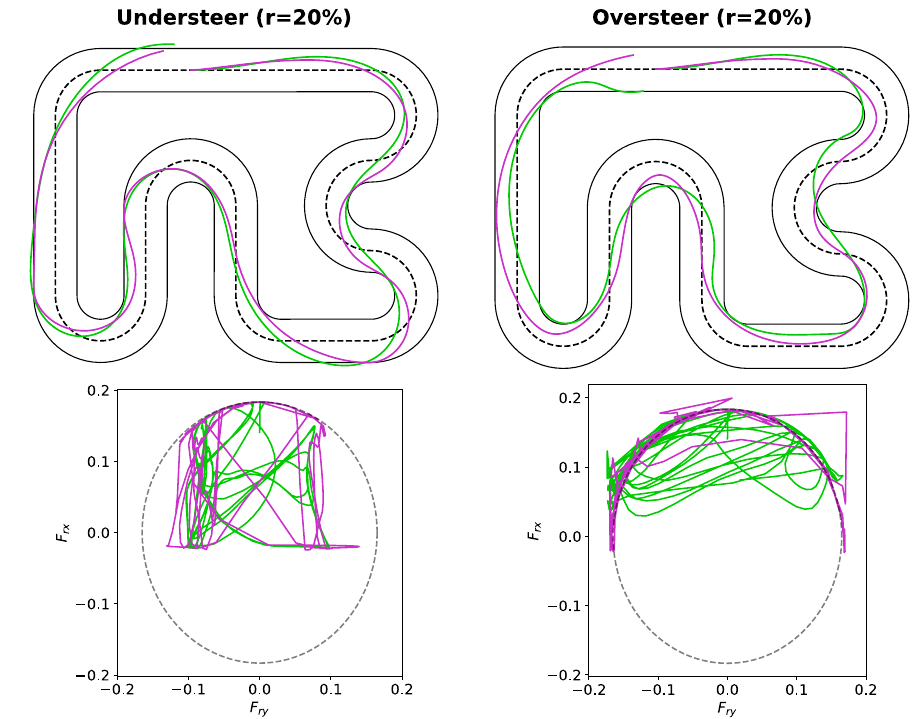}
 \vspace{-5mm}
 \caption{Understeering and oversteering trajectories and lateral rear tire forces with a mismatch of $r=20\%$ in \tm and \prm.}\label{fig:deterministic_mismatch}
\end{figure}
\noindent From Fig.~\ref{fig:deterministic_mismatch}, we observe a clear deviation from the nominal scenario without model mismatch in Fig.~\ref{fig:comp_nomismatch}. In the understeering case, the rear tires slip in every corner, causing the car to drift outward, while in the oversteering case, the front wheels slip, leading the car to drift through the corner. This behavior is also reflected in the rear tire friction circles, shown in the bottom row of Fig.~\ref{fig:deterministic_mismatch}. In the understeering scenario, the front tires reach their limit before the rear tires, preventing the rear tires from achieving maximum grip. Conversely, in the oversteering scenario, the rear tires slip at each corner, pushing forces beyond the friction circle. For a clearer understanding of these dynamics, we encourage the reader to view the accompanying video, which animates these trajectories

With the impact of mismatch on the tire model understood, we now aim to quantify its effect on the controllers. To do this, we analyze the lap times achieved by both controllers under various reductions in the front and rear tires:
\vspace{-2mm}
\begin{figure}[H]
\bluefigure
\centering
 \includegraphics[width=\linewidth]{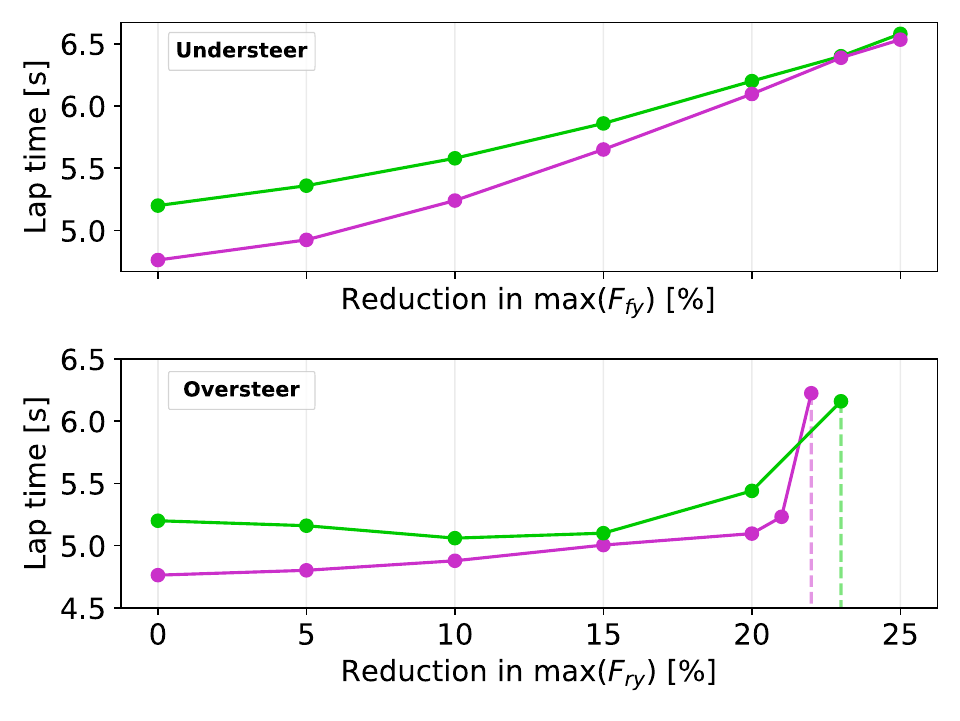}
 \caption{Laptimes for multiple tire mismatch reduction values $r$ for understeering and oversteering with \tm and \prm. The dashed lines represent the tire reduction upon which the car spins.}\label{fig:deterministic_mismatch_laptimes}
\end{figure}
\vspace{-4mm}
In both scenarios, the minimum-time controller outperforms progress maximization; however, this performance difference narrows as the mismatch increases. The minimum-time controller drives the system toward its theoretical limits, but as mismatch grows, the gap between the theoretical model and the actual system widens, causing greater discrepancies between the planned and actual trajectories. In contrast, the progress maximization controller’s cost function balances lag error, regularization, and progress maximization—all encapsulated in the $g$ term in the cost function~\eqref{eq:maxProgr_cost}—which inherently makes it more conservative. This suboptimality becomes beneficial as mismatch increases, as its commands are less aggressive. This effect is particularly evident in the oversteer scenario: while the minimum-time controller outperforms progress maximization up to a critical mismatch level ($r_r = 21.5\%$), it eventually causes the car to spin, indicated by the green dashed line. In contrast, the progress maximization controller’s less aggressive maneuvers delay the onset of spinning to a higher mismatch level ($r_r = 23\%)$, shown by the magenta dashed line.
\end{sidebar}

\begin{sidebar}{\continuesidebar}
\subsubsection{Comparison with stochastic model mismatch}
In the previous subsection, we examined each controller’s performance under a deterministic mismatch. However, in real-world scenarios, the discrepancies between theoretical and actual models are typically non-deterministic and arise from multiple sources, each with unique behaviors.

To address this, we evaluate the controllers in the presence of a non-deterministic model mismatch. For this purpose, we sample the reduction factor from a normal distribution, defined as $r = \mathcal{N}(0, \sigma)$, where $\sigma$ represents the standard deviation of the distribution. In addition, we leverage the insights from the previous subsection to prevent non-representative edge cases where the car spins. We do this by bounding the maximum reduction to $r \leq 20\%$. Ultimately, the resulting model of the tire forces is as follows:
\begin{subequations}~\label{eq:stochastic_model_mismatch}
\begin{align}
\tilde{D}_f = 1-\min(|\mathcal{N}_f(0,\sigma)|,0.2)D_f\,,\\
\tilde{D}_r = 1-\min(|\mathcal{N}_r(0,\sigma)|,0.2)D_r\,.
\end{align}
\end{subequations}
To assess the performance of the controllers under the stochastic model mismatch described in equation \eqref{eq:stochastic_model_mismatch}, we evaluated the controllers for various standard deviations, $\sigma$, ranging from 0.1 to 0.5. For each value of $\sigma$, we conducted 15 experiments, providing a thorough representation of the controllers' performance under stochastic mismatch.

The resulting lap times are depicted in Fig.~\ref{fig:stochastic_mismatch}. In both controllers, lap times increase as $\sigma$ increases. Notably, the time gap between the two methods remains constant. Furthermore, since the reduction is constrained to the admissible range—where the car is not at risk of spinning—the time-minimization controller consistently outperforms the progress-maximization controller.
\begin{figure}[H]
\bluefigure
\centering
 \includegraphics[width=\linewidth]{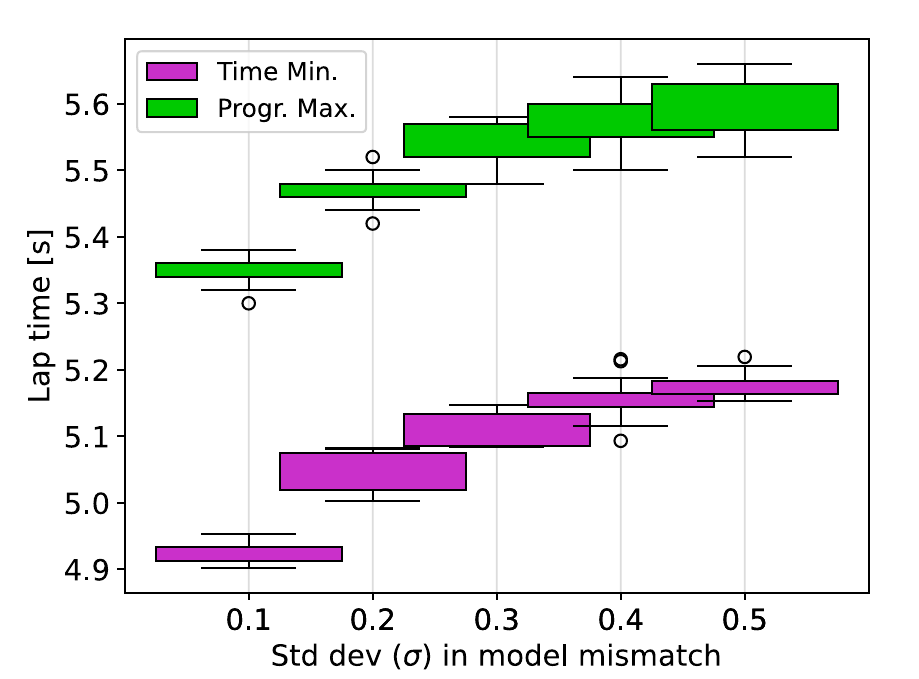}
 \vspace{-3mm}
 \caption{Lap times for a stochastic mismatch, where the reductions at the front and rear tires are obtained from a normal distributions, evaluated for multiple standard deviations. Each case study consists of $15$ runs.}\label{fig:stochastic_mismatch}
\end{figure}   

\subsubsection{Discussion}
Generally, minimizing time is preferable, particularly when the model mismatch is restricted to a permissible range. However, if the mismatch extends beyond this range, maximizing progress proves to be more reliable. Additionally, the numerical implementation of progress maximization aligns with existing numerical solvers, avoiding the complex numerics needed for solving the min time problem, which demands a specially designed solver.
\end{sidebar}

\pagebreak
\section{Path-Parametric Corridors:\\ A continuous spatial representation}
As discussed throughout this manuscript, spatial coordinates derived from path-parametric methods inherently capture the concept of advancement along a path while enabling the imposition of spatial bounds as convex constraints in the orthogonal components of the spatial states (see \textcolor{black}{Fig.~\ref{fig:tunnel_quad}}). These features make path-parametric methods a compelling toolset for planning and control algorithms in navigation. For instance, in the motion planning example of the robotic manipulator shown in Fig.~\ref{fig:whypp_corridor}, explicitly representing the admissible region allows the system to efficiently utilize the available space, guiding the end effector along the reference path. Similarly, in the racing scenario, where minimizing time and maximizing progress were compared, an explicit representation of the racetrack enabled controllers to fully exploit the road's width for optimal performance.

These examples highlight the potential of parameterizing a system’s motion using spatial coordinates. However, they also raise a critical question: how can one formulate and compute a spatial representation that effectively describes the admissible region around an arbitrary reference path as a function of the orthogonal spatial component, $\eta$?

Existing methods based on convex decomposition~\cite{deits2015computing, liu2017planning, zhong2020generating, toumieh2022voxel}, which partition free space into convex polyhedra, are incompatible with such formulations. These approaches adhere to a \emph{Euclidean} perspective and fail to leverage the capability of imposing convex constraints specifically in the orthogonal spatial coordinate $\eta$. Moreover, discretizing free space into multiple convex sets introduces the \emph{polyhedron allocation problem}, where each state must be preassigned to a specific polyhedron. This discretization disrupts the differentiability of the corridors with respect to the progress variable, making it challenging to incorporate collision-free corridors into optimization and learning frameworks.

To overcome these limitations, \cite{arrizabalaga2024differentiable} presented a method for generating \mybox{framegreen}{\hyperref[sidebar:corridor]{\emph{differentiable, continuous and collision-free}}} \mybox{framegreen}{\hyperref[sidebar:corridor]{\emph{corridors}}}. We proceed to detail its key components.


\begin{greensidebar}[sidebar:corridor]{Differentiable Parametric Collision-Free Corridors}
\begin{multicols}{2}

\sdbargreeninitial{M}otived by the incompatibility of convex decomposition based corridors for parametric formulations, ~\cite{arrizabalaga2024differentiable} proposed a method for computing corridors that are differentiable, continuous and collision-free. We now outline its main ingredients and crucial role in the path-parametric framework.

\subsection{Choosing an off-centered ellipse as the cross-section}
A parametric corridor is defined as a predefined cross-section that sweeps the parametric path given by the user. For this reason, choosing a cross section that offers maximum adaptability to the obstacle-free space and ensures differentiability and computational feasibility is of upmost importance.

The simplest option is a circle, which offers a single degree of freedom (the radius). An ellipse increases this number to three, and allowing it to have an offset from the path further raises the total degrees of freedom to 5. Beyond the circular or elliptical choices, there are more elaborate options, such as polyhedrons or semialgebraic sets. However, these cross sections cannot guarantee to remain differentiable or closed throughout the entire corridor. 

For this reason, the most favorable option within the feasible ones is a rotating off-centered ellipse. Mathematically, this ellipse is defined by the matrix $\mathrm{E}$ and the offset $\bm{p_E}$, and its equation is given by
\begin{equation}\label{eq:ellipse}
    (\bm{x}_\perp - \bm{p_E})^\intercal \mathrm{E} (\bm{x}_\perp - \bm{p_E})\leq{1}\,.
\end{equation}

\subsection{Parameterizing the ellipse with polynomials}
Having defined the cross section as a rotating off-centered ellipse, the next step is to parameterize it, so that it sweeps the reference path from the beginning to the end. This implies that the ellipse, and thereby the matrix $\mathrm{E}$ and the offset $\bm{p_E}$, evolve according to path-parameter $\xi$.  For this purpose, these variables are chosen to be parameterized by Chebyshev polynomials:

\begin{equation}\label{eq:polynomial_paramaterization}
    \{\mathrm{E}(\xi), \bm{p_E}(\xi)\} \rightarrow \{\mathrm{E}(\xi, \bm{c}_E), \bm{p}(\xi,\bm{p}_E)\}
\end{equation}
where $\bm{c}_E$ and $\bm{p}_E$ are the coefficients of the polynomials. There are two reasons underpinning the choice of Chebyshev polynomials: Firstly, utilizing polynomials ensures that the resulting corridors exhibit inherent smoothness and differentiability.  Secondly, the Chebyshev basis guarantees that our method is numerically stable, even for high polynomial degrees. Intuitively, increasing the degree of the polynomial enhances the corridor’s ability to adapt to variations along the reference path. 

Introducing the polynomial parameterization in~\eqref{eq:polynomial_paramaterization} into the off-centered ellipse cross section in~\eqref{eq:ellipse}
\begin{equation*}
    (\bm{x}_\perp - \bm{p_E}(\xi,\bm{c_P}))^\intercal \mathrm{E}(\xi,\bm{c_E}) (\bm{x}_\perp - \bm{p_E}(\xi,\bm{c_P}))\leq{1}\,.
\end{equation*}
and removing the nonlinearities leads to
\begin{equation}\label{eq:corridor}
    \bm{x}_\perp^\intercal \mathrm{E}(\xi,\bm{c_E}) \bm{x}_\perp -  \bm{d}(\xi,\bm{d_E})\leq{1}\,,
\end{equation}
where $\bm{d}(\xi,\bm{c_D})$ maintains the offset of the ellipse, while removing the nonlinearities of~\eqref{eq:polynomial_paramaterization}. From these decisions, it becomes evident that the shape and size of the corridor are entirely determined by the polynomial coefficients $\bm{c}_E$ and $\bm{d}_E$. This naturally raises the question: how can these coefficients be determined? To address this, we turn to the third and final component of the methodology.
\end{multicols}
\begin{figure}[H]
\centering
\def\figurename{\textcolor{strongerGreen}{\textbf{FIGURE}}} 
\captionsetup{labelformat=simple, labelfont={color=strongerGreen, bf}} 
\includegraphics[width=\columnwidth]{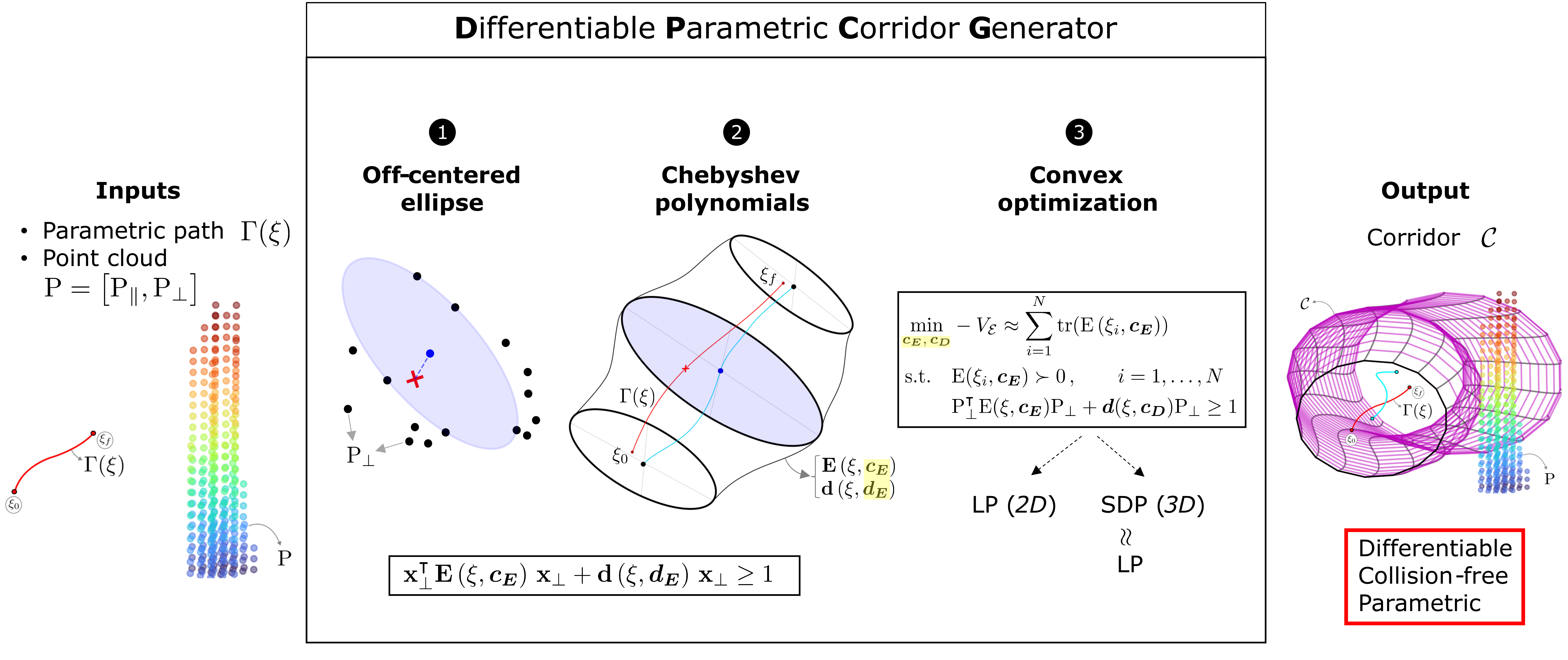}
\caption{An overview of the method introduced in~\cite{arrizabalaga2024differentiable} for computing differentiable, continuous, and parametric corridors fully compliant with the presented path-parametric framework. The method requires only a parametric path and a point cloud to generate the corridors, offering a system- and environment-agnostic tool that extends the applicability of path-parametric methods to real-world scenarios beyond well-defined and controlled environments. It achieves this through three key components: (1) selecting an off-centered ellipse as the corridor's cross-section, (2) parameterizing the cross-section using Chebyshev polynomials, and (3) formulating the corridor volume maximization problem as a lightweight convex optimization task solvable in real-time.}\label{fig:corrgen_overview}
\vspace{-2mm}
\end{figure}
\end{greensidebar}

\begin{greensidebar}{\continuesidebar}
\begin{figure}[H]
    \centering
    \def\figurename{\textcolor{strongerGreen}{\textbf{FIGURE}}} 
    \captionsetup{labelformat=simple, labelfont={color=strongerGreen, bf}} 
    \includegraphics[width=\columnwidth]{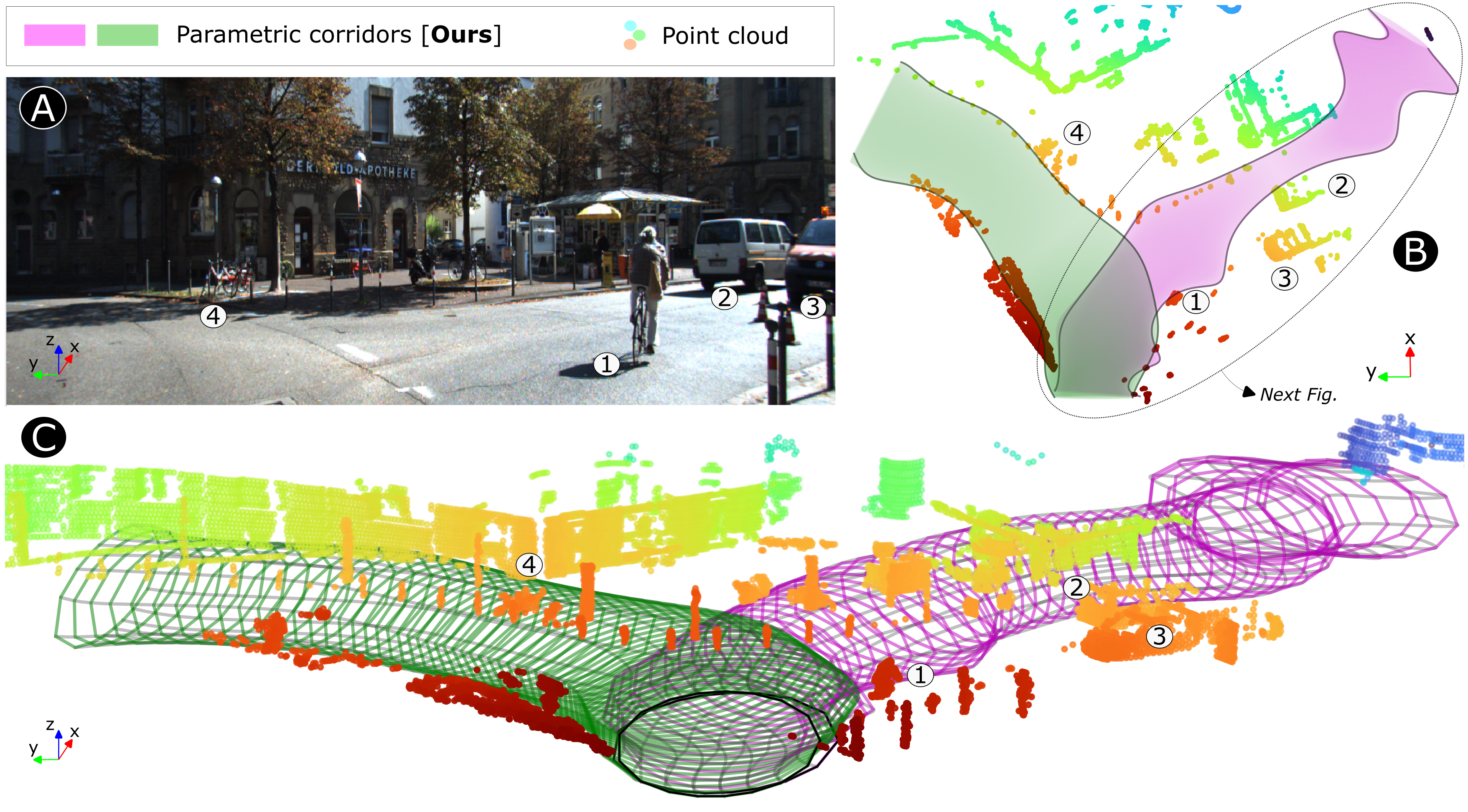}
    \caption{Visualization of two 3D path-parametric corridors (polynomial degree 9) generated using the method in~\cite{arrizabalaga2024differentiable} for a challenging KITTI Vision Benchmark~\cite{Geiger2013IJRR} scenario featuring narrow, bifurcating lanes, vehicles, and cyclists. Panel (A) presents an RGB camera view, while Panel (B) shows a top-down perspective, and Panel (C) provides an isometric view. The point cloud and corridors are color-coded by proximity to the camera. White markers (1–4) indicate key scene features. For details on the dashed corridor in Panel B, refer to Fig.~\ref{fig:kitti_second}.}\label{fig:kitti_first}
\end{figure}
\begin{multicols}{2}
\subsection{Corridor volume maximization as convex optimization}
From eq.~\eqref{eq:corridor}, it is apparent that the corridor's shape and size is exclusively dependent on the polynomial coefficients $\bm{c_E}$ and $\bm{d_E}$. From the available collision-free corridors, we aim to identify the one with the largest volume. To determine the coefficients of this corridor, we formulate an optimization problem that maximizes the volume, ensuring that the ellipse matrix $\mathrm{E}$ remains positive definite throughout the corridor, and that all points in the point cloud $\mathrm{P}_\perp$ lie outside of it. Given that the area of the ellipse in~\eqref{eq:ellipse} is given by $A_\mathcal{E}=\pi/\sqrt{\det \mathrm{E}}\,$, the optimization problem that we solve is:

\begin{subequations}\label{eq:ocp}
    \begin{alignat}{3}
    \max_{\bm{c_E},\bm{d_E}} V_\mathcal{E} &= \int_{\xi_0}^{\xi_f} -\det \mathrm{E}(\xi, \bm{c_E})\,d\xi\label{eq:ocp_cost}\\
    \text{s.t.}\quad &\mathrm{E}(\xi,\bm{c_E})\succ 0\,,\nonumber\\
&\mathrm{P}_\perp^\intercal \mathrm{E}(\xi,\bm{c_E}) \mathrm{P}_\perp -  \bm{d}(\xi,\bm{d_E})\leq{1}\,.\nonumber
    \end{alignat}
\end{subequations}

\noindent To make this problem computationally tractable, we conduct the following approximations. Firstly, we discretize the continuous parts of the optimization problem into N evaluations. Secondly, we avoid the nonlinearities associated with the determinant of the ellipse by approximating it with the trace. Thirdly, we introduce a wrapper around the reference path, ensuring that the problem always remains bounded. By incorporating all three modifications, the original problem becomes a convex optimization problem:
\begin{subequations}\label{eq:sdp}
 \begin{alignat}{3}
    \min_{\bm{c_E},\,\bm{d_E}} -V_\mathcal{E} &\approx \sum_{i=1}^{N} \text{tr}(\mathrm{E}\left(\xi_i,\bm{c_E})\right)\label{eq:sdp_cost}\\
  \text{s.t.}\quad &\mathrm{E}(\xi_i,\bm{c_E})\succ 0\,,\label{eq:pd_constraint}\\
&\mathrm{P}_\perp^\intercal \mathrm{E}(\xi_i,\bm{c_E}) \mathrm{P}_\perp -  \bm{d}(\xi_i,\bm{d_E})\leq{1}
 \end{alignat}
\end{subequations}

\noindent More specifically, the optimization problem in~\eqref{eq:sdp} is a Semidefinite-Program (SDP). Despite being a convex problem, SDPs are the most difficult convex problems to solve. To overcome this burden, the linear matrix inequality in~\eqref{eq:pd_constraint} is replaced with diagonal dominance, i.e.,
\begin{equation}\label{eq:diagdomin_approx}
    \mathrm{E} = 
    \begin{bmatrix}
\mathrm{E}_{11} & \mathrm{E}_{12} \\
\mathrm{E}_{12} & \mathrm{E}_{22}
    \end{bmatrix}\succ0\quad\rightarrow\quad\mathrm{E}_{11},\mathrm{E}_{22} >> \mathrm{E}_{12}\,.
\end{equation}
This approximation reduces the SDP into a Linear Program (LP). LPs are very well understood and extremely lightweight, and thus, can be very efficiently solved with off-the shelf solvers. As a final remark, notice that, in the planar case there is no need for the diagonal dominance approximation in~\eqref{eq:diagdomin_approx}, since the original problem is already an LP.
\end{multicols}
\end{greensidebar}
\begin{greensidebar}{\continuesidebar}
\begin{figure}[H]
\centering
\def\figurename{\textcolor{strongerGreen}{\textbf{FIGURE}}} 
\captionsetup{labelformat=simple, labelfont={color=strongerGreen, bf}} 
\includegraphics[width=\columnwidth]{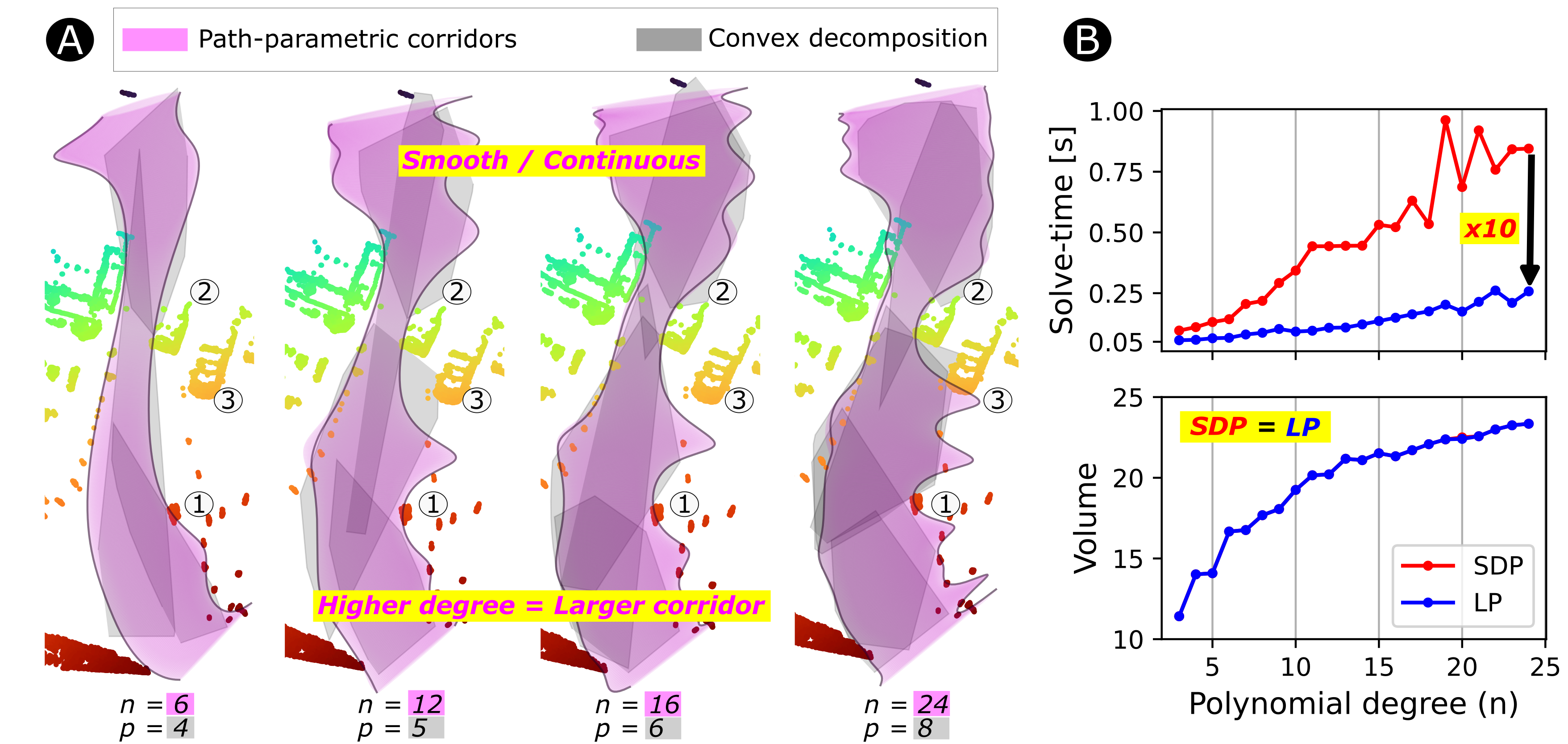}
\caption{A detailed analysis of the real-world scenario shown in Fig.~\ref{fig:kitti_first}. Panel (A) presents a top view of the corridors, generated by incrementally increasing the polynomial degree $n$, alongside the corridors obtained by decomposing the free space into $p$ convex sets, represented in gray. Panel (B) illustrates the evolution of the corridor volume and computation times as a function of the polynomial degree, for both the SDP and LP problems.}\label{fig:kitti_second}
\vspace{-3mm}
\end{figure}
\begin{multicols}{2}
\subsection{Experiment: An autonomous driving case-study}
The method's real-world applicability is demonstrated through an autonomous driving case study of a roundabout, where two corridors are computed. Figure~\ref{fig:kitti_first} shows these corridors alongside a lidar-generated point cloud and a raw RGB camera image. The corridors effectively capture the free space, spanning the entire road width while accommodating vehicles, bicycles, and road boundaries. The analysis examines three key aspects of the method.


First, to better understand the influence of polynomial degree on the shape and size of the corridors, we conduct evaluations across various degrees. The resulting solutions are presented in Fig.~\ref{fig:kitti_second}-A, where top-down views of the corridors, obtained by sequentially increasing the polynomial degree $n$, are shown. It is clear from these images that higher polynomial degrees enhance the expressiveness of the corridors. This trend is further illustrated in the plot at the bottom of panel (B), which demonstrates that an increase in polynomial degree corresponds to a growth in the corridor's volume. 
Second, to illustrate the comparison between the path-parametric corridor generation method and the well-established convex decomposition, Fig.~\ref{fig:kitti_second}-A is considered once more. The path-parametric corridors are depicted alongside the corridors resulting from decomposing the free space into $p$ convex sets. The results show that both methods encompass very similar spaces in all four cases. However, the hyperparameters differ significantly. In convex decomposition, the quantity and distribution of polyhedra are determined based on the number of segments within a precomputed linear path, which may compromise the corridor's volume if the reference path has few segments. In contrast, this method employs the polynomial degree $n$ as its unique hyperparameter, while remaining agnostic to the underlying reference. Furthermore, Figure~\ref{fig:kitti_second}-A also illustrates how a parametric cross section with a polynomial basis results in a continuous and smooth space representation, contrasting with the discreteness intrinsic to convex decomposition.

Third, to assess the trade-off between volume gains and computational cost associated with increasing the polynomial degree ($n$), Figure~\ref{fig:kitti_second}-B is examined. This figure illustrates the volumes and solve times for the pink corridor depicted in Figure~\ref{fig:kitti_first}, across polynomial degrees ranging from $n=3$ to $n=25$. It compares both the original SDP formulation and the approximated LP relaxation. Notably, the results demonstrate that the LP relaxation achieves corridor volumes identical to the original SDP while yielding computational speedups of up to a factor of 10. This enhanced performance stems from two key factors. First, the cost function~\eqref{eq:sdp_cost}, which maximizes the trace, inherently promotes diagonally dominant matrices in the resulting corridors. Second, optimizing the offset relative to a reference overcomes the constraint of requiring a diagonally dominant matrix $\mathrm{E}$. This allows the resulting corridors to fully encompass the available space while preserving the diagonally dominant structure within their underlying matrices.

These results demonstrate that this corridor generation method can compute differentiable, continuous, and smooth parametric corridors at 5–20 Hz in unstructured arbitrary environments, making it suitable for real-time deployment alongside the various planning and control techniques of the presented path-parametric framework.
\end{multicols}
\end{greensidebar}

\newpage
\section{Conclusions}

Path-parametric methods have emerged as a cornerstone in egocentric decision-making, owing to three key advantages: they inherently model progress along a path, incorporate geometric features such as curvature and torsion into system dynamics, and enable spatial bounds to be expressed as convex constraints on orthogonal spatial states. These attributes have made path-parametric formulations highly effective in planar scenarios, such as autonomous driving. However, extending these methods to real-world three-dimensional cases---where curves are defined by both curvature and torsion---has proven challenging. This subtle yet critical difference has limited their applicability to complex 3D problems, including aerial navigation and robotic manipulator control.

Existing approaches to path-parametric problems are often presented as isolated works, resulting in a fragmented literature where techniques appear disconnected. This disjointed presentation obscures the underlying relationships between methods, leaving readers with an incomplete understanding of the field. To address this, we proposed a universal formulation for path-parametric planning and control, demonstrating how these approaches are fundamentally interconnected.

For this purpose, we analyzed the path-parametric problem from multiple yet interrelated perspectives. First, we examined the \emph{interplay of existing parametric techniques} and showed how they can be unified under a single framework composed of two key components: (i) a \emph{path-parameterization technique} and (ii) a \emph{spatial representation of system dynamics}. 

To illustrate how these components enable the formulation of planning and control strategies, we applied them to a two-link robotic manipulator in two distinct contexts. First, we focused on \emph{low-level control}, exploring the foundational motivations behind the development of path-parametric methods. Second, we demonstrated how this framework extends to more versatile optimization-based \emph{motion planning} scenarios.

Next, we delved into one of the most popular instantiations of this framework: the time-minimization approximation through \emph{progress maximization}. Using a miniature racing car example, we highlighted the differences between these two formulations. Although progress maximization serves as a proxy for the original time-minimization problem, experimental results showed that it achieves comparable lap times while offering additional advantages. Notably, it simplifies implementation and facilitates the incorporation of spatial constraints by imposing convex bounds on the orthogonal distance to the road's centerline.

Finally, to generalize this approach for navigating safe corridors in diverse environments---beyond cases where a road is explicitly defined---we focused on representing the admissible environment around the parametric path. Specifically, we introduced a method for generating differentiable, continuous, and collision-free \emph{corridors}. These corridors, thanks to their properties, are fully compatible with any gradient-based path-parametric method, whether used as constraints or as environmental information to guide the optimization algorithm.

\section{Acknowledgments}

The authors would like to express their gratitude to Rudolf Reiter, Ángel Romero, and Jelena Trisovic for their invaluable insights and engaging discussions on path-parametric methods and their diverse applications in robotics and autonomous systems. Special thanks also go to Jon Santamaría for his assistance in creating the graphics for the cover figure.

\section{Author Information}

\begin{IEEEbiography}{{J}on Arrizabalaga}{\,}(jon.arrizabalaga@tum.de) is pursuing his PhD at the Munich Institute of Robotics and Machine Intelligence, Technical University of Munich, and is a visiting researcher at the Robotics Institute of Carnegie Mellon University. He received a BSc. in mechanical engineering (Spain, 2018) and an MSc. in mechatronics and robotics at KTH Royal Institute of Technology (Sweden, 2020). His research focuses on the convergence of optimization, control, and geometry, particularly in their applications to robotics and autonomous systems.
\end{IEEEbiography}

\begin{IEEEbiography}{Zbyněk Šír}{\,}(zbynek.sir@mff.cuni.cz) is an associate professor at the Mathematical Institute in the Faculty of Mathematics and Physics of Charles University in Prague. He received a DEA (Diplome d’études approfondies) from University Pierre et Marie Curie, a Master with specelization in Riemannian Geometry and Theory of Representations, and a PhD in mathematics with specialization in history of French geometry, both from Charles University of Prague. His research interestes include CAGD, theoretical differential geometry, history of geometry and other applied geometric fields.
\end{IEEEbiography}

\begin{IEEEbiography}{{Z}achary Manchester}{\,}(zmanches@andrew.cmu.edu)
is an assistant professor in the Robotics Institute at Carnegie Mellon University and founder of the Robotic Exploration Lab. He received a PhD in aerospace engineering in 2015 and a BS in applied physics in 2009, both from Cornell University. His research interests include control and optimization with applications to aerospace and robotic systems.
\end{IEEEbiography}

\begin{IEEEbiography}{{M}arkus Ryll}{\,}(markus.ryll@tum.de) is an assistant professor in the Department of Aerospace and Geodesy at the Technical University of Munich and head of the Autonomous Aerial Systems Lab. He received a PhD in 2014 from Max Planck Institute for Biological Cybernetics. Between 2014 and 2017, he was a postdoctoral researcher at the Laboratory for Analysis and Architecture of Systems (LAAS-CNRS). From 2018 to 2020, he continued his postdoctoral work in the Robust Robotics Group at the Massachusetts Institute of Technology (MIT, CSAIL). His research focuses on enabling autonomous aerial robots to interact with real-world environments.
\end{IEEEbiography}

\bibliographystyle{ieeetr}
\bibliography{lib}

\begin{thebibliography}{10}

\bibitem{verschueren2016time}
R.~Verschueren, N.~van Duijkeren, J.~Swevers, and M.~Diehl, ``Time-optimal motion planning for n-dof robot manipulators using a path-parametric system reformulation,'' in {\em 2016 American Control Conference (ACC)}, pp.~2092--2097, IEEE, 2016.

\bibitem{spedicato2017minimum}
S.~Spedicato and G.~Notarstefano, ``Minimum-time trajectory generation for quadrotors in constrained environments,'' {\em IEEE Transactions on Control Systems Technology}, vol.~26, no.~4, pp.~1335--1344, 2017.

\bibitem{arrizabalaga2023sctomp}
J.~Arrizabalaga and M.~Ryll, ``Sctomp: Spatially constrained time-optimal motion planning,'' in {\em 2023 IEEE/RSJ International Conference on Intelligent Robots and Systems (IROS)}, pp.~4827--4834, IEEE, 2023.

\bibitem{song2021autonomous}
Y.~Song, M.~Steinweg, E.~Kaufmann, and D.~Scaramuzza, ``Autonomous drone racing with deep reinforcement learning,'' in {\em 2021 IEEE/RSJ International Conference on Intelligent Robots and Systems (IROS)}, pp.~1205--1212, IEEE, 2021.

\bibitem{wurman2022outracing}
P.~R. Wurman, S.~Barrett, K.~Kawamoto, J.~MacGlashan, K.~Subramanian, T.~J. Walsh, R.~Capobianco, A.~Devlic, F.~Eckert, F.~Fuchs, {\em et~al.}, ``Outracing champion gran turismo drivers with deep reinforcement learning,'' {\em Nature}, vol.~602, no.~7896, pp.~223--228, 2022.

\bibitem{kaufmann2023champion}
E.~Kaufmann, L.~Bauersfeld, A.~Loquercio, M.~M{\"u}ller, V.~Koltun, and D.~Scaramuzza, ``Champion-level drone racing using deep reinforcement learning,'' {\em Nature}, vol.~620, no.~7976, pp.~982--987, 2023.

\bibitem{liniger2015optimization}
A.~Liniger, A.~Domahidi, and M.~Morari, ``Optimization-based autonomous racing of 1: 43 scale rc cars,'' {\em Optimal Control Applications and Methods}, vol.~36, no.~5, pp.~628--647, 2015.

\bibitem{oelerich2024boundmpc}
T.~Oelerich, F.~Beck, C.~Hartl-Nesic, and A.~Kugi, ``Boundmpc: Cartesian trajectory planning with error bounds based on model predictive control in the joint space,'' {\em arXiv preprint arXiv:2401.05057}, 2024.

\bibitem{arrizabalaga2022towards}
J.~Arrizabalaga and M.~Ryll, ``Towards time-optimal tunnel-following for quadrotors,'' in {\em 2022 International Conference on Robotics and Automation (ICRA)}, pp.~4044--4050, IEEE, 2022.

\bibitem{lam2010model}
D.~Lam, C.~Manzie, and M.~Good, ``Model predictive contouring control,'' in {\em 49th IEEE Conference on Decision and Control (CDC)}, pp.~6137--6142, IEEE, 2010.

\bibitem{faulwasser2015nonlinear}
T.~Faulwasser and R.~Findeisen, ``Nonlinear model predictive control for constrained output path following,'' {\em IEEE Transactions on Automatic Control}, vol.~61, no.~4, pp.~1026--1039, 2015.

\bibitem{verscheure2009time}
D.~Verscheure, B.~Demeulenaere, J.~Swevers, J.~De~Schutter, and M.~Diehl, ``Time-optimal path tracking for robots: A convex optimization approach,'' {\em IEEE Transactions on Automatic Control}, vol.~54, no.~10, pp.~2318--2327, 2009.

\bibitem{van2016path}
N.~van Duijkeren, R.~Verschueren, G.~Pipeleers, M.~Diehl, and J.~Swevers, ``Path-following nmpc for serial-link robot manipulators using a path-parametric system reformulation,'' in {\em 2016 European Control Conference (ECC)}, pp.~477--482, IEEE, 2016.

\bibitem{arrizabalaga2022spatial}
J.~Arrizabalaga and M.~Ryll, ``Spatial motion planning with pythagorean hodograph curves,'' in {\em 2022 IEEE 61st Conference on Decision and Control (CDC)}, pp.~2047--2053, IEEE, 2022.

\bibitem{thehistoryofcurvature}
``The history of curvature.'' \url{https://web.archive.org/web/20071106083431/http://www3.villanova.edu/maple/misc/history_of_curvature/k.htm}.
\newblock Accessed: 2024-05-27.

\bibitem{hollerbach1983dynamic}
J.~M. Hollerbach, ``Dynamic scaling of manipulator trajectories,'' in {\em 1983 American Control Conference}, pp.~752--756, IEEE, 1983.

\bibitem{shin1985minimum}
K.~Shin and N.~McKay, ``Minimum-time control of robotic manipulators with geometric path constraints,'' {\em IEEE Transactions on Automatic Control}, vol.~30, no.~6, pp.~531--541, 1985.

\bibitem{pfeiffer1987concept}
F.~Pfeiffer and R.~Johanni, ``A concept for manipulator trajectory planning,'' {\em IEEE Journal on Robotics and Automation}, vol.~3, no.~2, pp.~115--123, 1987.

\bibitem{nelson1988local}
W.~L. Nelson and I.~J. Cox, ``Local path control for an autonomous vehicle,'' in {\em Proceedings. 1988 IEEE International Conference on Robotics and Automation}, pp.~1504--1510, IEEE, 1988.

\bibitem{kanayama1990stable}
Y.~Kanayama, Y.~Kimura, F.~Miyazaki, and T.~Noguchi, ``A stable tracking control method for an autonomous mobile robot,'' in {\em Proceedings., IEEE International Conference on Robotics and Automation}, pp.~384--389, IEEE, 1990.

\bibitem{cox1991blanche}
I.~J. Cox, ``Blanche-an experiment in guidance and navigation of an autonomous robot vehicle,'' {\em IEEE Transactions on robotics and automation}, vol.~7, no.~2, pp.~193--204, 1991.

\bibitem{hauser1995maneuver}
J.~Hauser and R.~Hindman, ``Maneuver regulation from trajectory tracking: Feedback linearizable systems,'' {\em IFAC Proceedings Volumes}, vol.~28, no.~14, pp.~595--600, 1995.

\bibitem{skjetne2004robust}
R.~Skjetne, T.~I. Fossen, and P.~V. Kokotovi{\'c}, ``Robust output maneuvering for a class of nonlinear systems,'' {\em Automatica}, vol.~40, no.~3, pp.~373--383, 2004.

\bibitem{do2004robust}
K.~D. Do, Z.-P. Jiang, and J.~Pan, ``Robust adaptive path following of underactuated ships,'' {\em Automatica}, vol.~40, no.~6, pp.~929--944, 2004.

\bibitem{aguiar2005path}
A.~P. Aguiar, J.~P. Hespanha, and P.~V. Kokotovic, ``Path-following for nonminimum phase systems removes performance limitations,'' {\em IEEE Transactions on Automatic Control}, vol.~50, no.~2, pp.~234--239, 2005.

\bibitem{aguiar2007trajectory}
A.~P. Aguiar and J.~P. Hespanha, ``Trajectory-tracking and path-following of underactuated autonomous vehicles with parametric modeling uncertainty,'' {\em IEEE transactions on automatic control}, vol.~52, no.~8, pp.~1362--1379, 2007.

\bibitem{faulwasser2009model}
T.~Faulwasser, B.~Kern, and R.~Findeisen, ``Model predictive path-following for constrained nonlinear systems,'' in {\em Proceedings of the 48h IEEE Conference on Decision and Control (CDC) held jointly with 2009 28th Chinese Control Conference}, pp.~8642--8647, IEEE, 2009.

\bibitem{kehrle2011optimal}
F.~Kehrle, J.~V. Frasch, C.~Kirches, and S.~Sager, ``Optimal control of formula 1 race cars in a vdrift based virtual environment,'' {\em IFAC Proceedings Volumes}, vol.~44, no.~1, pp.~11907--11912, 2011.

\bibitem{gao2012spatial}
Y.~Gao, A.~Gray, J.~V. Frasch, T.~Lin, E.~Tseng, J.~K. Hedrick, and F.~Borrelli, ``Spatial predictive control for agile semi-autonomous ground vehicles,'' in {\em Proceedings of the 11th international symposium on advanced vehicle control}, no.~2, pp.~1--6, 2012.

\bibitem{faulwasser2013predictive}
T.~Faulwasser, J.~Matschek, P.~Zometa, and R.~Findeisen, ``Predictive path-following control: Concept and implementation for an industrial robot,'' in {\em 2013 IEEE International Conference on Control Applications (CCA)}, pp.~128--133, IEEE, 2013.

\bibitem{frasch2013auto}
J.~V. Frasch, A.~Gray, M.~Zanon, H.~J. Ferreau, S.~Sager, F.~Borrelli, and M.~Diehl, ``An auto-generated nonlinear mpc algorithm for real-time obstacle avoidance of ground vehicles,'' in {\em 2013 European Control Conference (ECC)}, pp.~4136--4141, IEEE, 2013.

\bibitem{bock2013real}
M.~B{\"o}ck and A.~Kugi, ``Real-time nonlinear model predictive path-following control of a laboratory tower crane,'' {\em IEEE Transactions on Control Systems Technology}, vol.~22, no.~4, pp.~1461--1473, 2013.

\bibitem{kumar2017path}
S.~Kumar and R.~Gill, ``Path following for quadrotors,'' in {\em 2017 IEEE Conference on Control Technology and Applications (CCTA)}, pp.~2075--2081, IEEE, 2017.

\bibitem{brito2019model}
B.~Brito, B.~Floor, L.~Ferranti, and J.~Alonso-Mora, ``Model predictive contouring control for collision avoidance in unstructured dynamic environments,'' {\em IEEE Robotics and Automation Letters}, vol.~4, no.~4, pp.~4459--4466, 2019.

\bibitem{kloeser2020nmpc}
D.~Kloeser, T.~Schoels, T.~Sartor, A.~Zanelli, G.~Prison, and M.~Diehl, ``Nmpc for racing using a singularity-free path-parametric model with obstacle avoidance,'' {\em IFAC-PapersOnLine}, vol.~53, no.~2, pp.~14324--14329, 2020.

\bibitem{reiter2021mixed}
R.~Reiter, M.~Kirchengast, D.~Watzenig, and M.~Diehl, ``Mixed-integer optimization-based planning for autonomous racing with obstacles and rewards,'' {\em IFAC-PapersOnLine}, vol.~54, no.~6, pp.~99--106, 2021.

\bibitem{ji2021cmpcc}
J.~Ji, X.~Zhou, C.~Xu, and F.~Gao, ``Cmpcc: Corridor-based model predictive contouring control for aggressive drone flight,'' in {\em Experimental Robotics: The 17th International Symposium}, pp.~37--46, Springer, 2021.

\bibitem{romero2022model}
A.~Romero, S.~Sun, P.~Foehn, and D.~Scaramuzza, ``Model predictive contouring control for time-optimal quadrotor flight,'' {\em IEEE Transactions on Robotics}, vol.~38, no.~6, pp.~3340--3356, 2022.

\bibitem{arrizabalaga2023pose}
J.~Arrizabalaga and M.~Ryll, ``Pose-following with dual quaternions,'' in {\em 2023 62nd IEEE Conference on Decision and Control (CDC)}, pp.~5959--5966, IEEE, 2023.

\bibitem{reiter2023frenet}
R.~Reiter, A.~Nurkanovi{\'c}, J.~Frey, and M.~Diehl, ``Frenet-cartesian model representations for automotive obstacle avoidance within nonlinear mpc,'' {\em European Journal of Control}, vol.~74, p.~100847, 2023.

\bibitem{fork2023euclidean}
T.~Fork and F.~Borrelli, ``Euclidean and non-euclidean trajectory optimization approaches for quadrotor racing,'' {\em arXiv preprint arXiv:2309.07262}, 2023.

\bibitem{krinner2024time}
M.~Krinner, A.~Romero, L.~Bauersfeld, M.~Zeilinger, A.~Carron, and D.~Scaramuzza, ``Time-optimal flight with safety constraints and data-driven dynamics,'' {\em arXiv preprint arXiv:2403.17551}, 2024.

\bibitem{foehn2021time}
P.~Foehn, A.~Romero, and D.~Scaramuzza, ``Time-optimal planning for quadrotor waypoint flight,'' {\em Science robotics}, vol.~6, no.~56, p.~eabh1221, 2021.

\bibitem{tordesillas2019faster}
J.~Tordesillas, B.~T. Lopez, and J.~P. How, ``Faster: Fast and safe trajectory planner for flights in unknown environments,'' in {\em 2019 IEEE/RSJ international conference on intelligent robots and systems (IROS)}, pp.~1934--1940, IEEE, 2019.

\bibitem{zhou2021raptor}
B.~Zhou, J.~Pan, F.~Gao, and S.~Shen, ``Raptor: Robust and perception-aware trajectory replanning for quadrotor fast flight,'' {\em IEEE Transactions on Robotics}, vol.~37, no.~6, pp.~1992--2009, 2021.

\bibitem{wang2002arc}
H.~Wang, J.~Kearney, and K.~Atkinson, ``Arc-length parameterized spline curves for real-time simulation,'' in {\em Proc. 5th International Conference on Curves and Surfaces}, vol.~387396, 2002.

\bibitem{zhao2016time}
S.~Zhao, ``Time derivative of rotation matrices: A tutorial,'' {\em arXiv preprint arXiv:1609.06088}, 2016.

\bibitem{bishop1975there}
R.~L. Bishop, ``There is more than one way to frame a curve,'' {\em The American Mathematical Monthly}, vol.~82, no.~3, pp.~246--251, 1975.

\bibitem{farouki2008pythagorean}
R.~T. Farouki, {\em Pythagorean—Hodograph Curves}.
\newblock Springer, 2008.

\bibitem{choi2002euler}
H.~I. Choi and C.~Y. Han, ``Euler--rodrigues frames on spatial pythagorean-hodograph curves,'' {\em Computer Aided Geometric Design}, vol.~19, no.~8, pp.~603--620, 2002.

\bibitem{arrizabalaga2024phodcos}
J.~Arrizabalaga, F.~Vega, Z.~{\v{S}}{\'I}R, Z.~Manchester, and M.~Ryll, ``Phodcos: Pythagorean hodograph-based differentiable coordinate system,'' {\em arXiv preprint arXiv:2410.07750}, 2024.

\bibitem{hanson1995parallel}
A.~J. Hanson and H.~Ma, ``Parallel transport approach to curve framing,'' {\em Indiana University, Techreports-TR425}, vol.~11, pp.~3--7, 1995.

\bibitem{wang2008computation}
W.~Wang, B.~J{\"u}ttler, D.~Zheng, and Y.~Liu, ``Computation of rotation minimizing frames,'' {\em ACM Transactions on Graphics (TOG)}, vol.~27, no.~1, pp.~1--18, 2008.

\bibitem{wanner1996solving}
G.~Wanner and E.~Hairer, {\em Solving ordinary differential equations II}, vol.~375.
\newblock Springer Berlin Heidelberg New York, 1996.

\bibitem{struik1961lectures}
D.~J. Struik, {\em Lectures on classical differential geometry}.
\newblock Courier Corporation, 1961.

\bibitem{abbena2017modern}
E.~Abbena, S.~Salamon, and A.~Gray, {\em Modern differential geometry of curves and surfaces with Mathematica}.
\newblock Chapman and Hall/CRC, 2017.

\bibitem{vsir2007}
Z.~{\v{S}}{\'\i}r and B.~J{\"u}ttler, ``$c^2$ hermite interpolation by pythagorean hodograph space curves,'' {\em Mathematics of Computation}, vol.~76, no.~259, pp.~1373--1391, 2007.

\bibitem{verschueren2018towards}
R.~Verschueren, G.~Frison, D.~Kouzoupis, N.~van Duijkeren, A.~Zanelli, R.~Quirynen, and M.~Diehl, ``Towards a modular software package for embedded optimization,'' {\em IFAC-PapersOnLine}, vol.~51, no.~20, pp.~374--380, 2018.

\bibitem{frison2020hpipm}
G.~Frison and M.~Diehl, ``Hpipm: a high-performance quadratic programming framework for model predictive control,'' {\em IFAC-PapersOnLine}, vol.~53, no.~2, pp.~6563--6569, 2020.

\bibitem{hung2023review}
N.~Hung, F.~Rego, J.~Quintas, J.~Cruz, M.~Jacinto, D.~Souto, A.~Potes, L.~Sebastiao, and A.~Pascoal, ``A review of path following control strategies for autonomous robotic vehicles: Theory, simulations, and experiments,'' {\em Journal of Field Robotics}, vol.~40, no.~3, pp.~747--779, 2023.

\bibitem{wachter2006implementation}
A.~W{\"a}chter and L.~T. Biegler, ``On the implementation of an interior-point filter line-search algorithm for large-scale nonlinear programming,'' {\em Mathematical programming}, vol.~106, pp.~25--57, 2006.

\bibitem{athans2007optimal}
M.~Athans and P.~L. Falb, {\em Optimal control: an introduction to the theory and its applications}.
\newblock Courier Corporation, 2007.

\bibitem{bobrow1985time}
J.~E. Bobrow, S.~Dubowsky, and J.~S. Gibson, ``Time-optimal control of robotic manipulators along specified paths,'' {\em The international journal of robotics research}, vol.~4, no.~3, pp.~3--17, 1985.

\bibitem{arrizabalaga2024differentiable}
J.~Arrizabalaga, Z.~Manchester, and M.~Ryll, ``Differentiable collision-free parametric corridors,'' in {\em 2024 IEEE/RSJ International Conference on Intelligent Robots and Systems (IROS)}, pp.~1839--1846, IEEE, 2024.

\bibitem{kiessling2024almost}
D.~Kiessling, C.~Vanaret, A.~Astudillo, W.~Decre, and J.~Swevers, ``An almost feasible sequential linear programming algorithm,'' {\em arXiv preprint arXiv:2401.13840}, 2024.

\bibitem{yangnew}
L.~Yang, T.~Marcucci, P.~A. Parrilo, and R.~Tedrake, ``A new semidefinite relaxation for linear and piecewise-affine optimal control with time scaling,''

\bibitem{nocedal1999numerical}
J.~Nocedal and S.~J. Wright, {\em Numerical optimization}.
\newblock Springer, 1999.

\bibitem{kiessling2022feasible}
D.~Kiessling, A.~Zanelli, A.~Nurkanovi{\'c}, J.~Gillis, M.~Diehl, M.~Zeilinger, G.~Pipeleers, and J.~Swevers, ``A feasible sequential linear programming algorithm with application to time-optimal path planning problems,'' in {\em 2022 IEEE 61st Conference on Decision and Control (CDC)}, pp.~1196--1203, IEEE, 2022.

\bibitem{deits2015computing}
R.~Deits and R.~Tedrake, ``Computing large convex regions of obstacle-free space through semidefinite programming,'' in {\em Algorithmic Foundations of Robotics XI: Selected Contributions of the Eleventh International Workshop on the Algorithmic Foundations of Robotics}, pp.~109--124, Springer, 2015.

\bibitem{liu2017planning}
S.~Liu, M.~Watterson, K.~Mohta, K.~Sun, S.~Bhattacharya, C.~J. Taylor, and V.~Kumar, ``Planning dynamically feasible trajectories for quadrotors using safe flight corridors in 3-d complex environments,'' {\em IEEE Robotics and Automation Letters}, vol.~2, no.~3, pp.~1688--1695, 2017.

\bibitem{zhong2020generating}
X.~Zhong, Y.~Wu, D.~Wang, Q.~Wang, C.~Xu, and F.~Gao, ``Generating large convex polytopes directly on point clouds,'' {\em arXiv preprint arXiv:2010.08744}, 2020.

\bibitem{toumieh2022voxel}
C.~Toumieh and A.~Lambert, ``Voxel-grid based convex decomposition of 3d space for safe corridor generation,'' {\em Journal of Intelligent \& Robotic Systems}, vol.~105, no.~4, p.~87, 2022.

\bibitem{Geiger2013IJRR}
A.~Geiger, P.~Lenz, C.~Stiller, and R.~Urtasun, ``Vision meets robotics: The kitti dataset,'' {\em International Journal of Robotics Research (IJRR)}, 2013.

\end{thebibliography}

\ifthenelse{\boolean{arxiv}}{}{\endarticle}

\end{document}